\documentclass[letterpaper]{article} % DO NOT CHANGE THIS
\usepackage[preprint]{arxiv2col}
\usepackage[hyphens]{url}            % DO NOT CHANGE THIS
\usepackage{graphicx}                % DO NOT CHANGE THIS
\usepackage{natbib}                  % DO NOT CHANGE THIS AND DO NOT ADD ANY OPTIONS TO IT
\usepackage{caption}                 % DO NOT CHANGE THIS AND DO NOT ADD ANY OPTIONS TO IT
\usepackage[utf8]{inputenc}
\usepackage{amsmath}
\usepackage{amssymb}
\usepackage{booktabs}
\usepackage{multirow}
\usepackage{subcaption}
\usepackage[table]{xcolor}
\usepackage{array}

\usepackage{tablefootnote} % real page-bottom footnotes from inside table* floats

\title{Signal-Routed Temperature Scaling: Low-Capacity Risk-Conditioned
Calibration for Small Validation Budgets}

\author{
Wenhao Liang,
Liangwei Nathan Zheng,
Lin Yue,
Wei Emma Zhang,\\
Mingyu Guo,
Olaf Maennel,
Weitong Chen
}
\affiliations{Adelaide University, Adelaide, Australia}

\begin{document}

\maketitle

%==========================================================================
% 1. Abstract
%==========================================================================
\begin{abstract}
When a classifier is recalibrated from only a few thousand held-out
examples, the capacity of the calibration map becomes a statistical design
choice rather than a purely architectural one: a scalar map can underfit
structured residual miscalibration, while a highly adaptive map can be
hard to estimate reliably from so small a split. We disentangle the
calibration objective from adaptive capacity and propose signal-routed
temperature scaling (\textbf{SRTS-BCE}), a $10$-parameter,
argmax-preserving calibrator that cross-fits a correctness-risk score over
six logit statistics and fits one top-label-BCE temperature per
$K{=}3$ risk groups, recovering TvA-TS as its $K{=}1$ limit. On
fine-tuned CIFAR-100 / ViT-B/16, SRTS-BCE reduces $\mathrm{ECE}_{15}$
from $1.65$ (scalar TvA-TS) to $0.96$, matching the higher-capacity
SMART+BCE head ($0.95$) at the full calibration budget. The two regimes
separate as the budget shrinks: at $n{=}250$ SRTS-BCE beats SMART+BCE on
all three CIFAR-100 backbones (the seed$\to$draw hierarchical interval
excludes zero), whereas the flagship comparison against the scalar remains
directional. A protocol-frozen Tiny-ImageNet follow-up reproduces the
small-budget separation and exhibits a budget-dependent ranking reversal
on Swin-T; matched routing and map controls show that the effect is tied
neither to the learned router nor to discrete grouping. Together the
results identify post-hoc calibrator capacity as a finite-sample design
choice whose preferred level shifts with the amount of available
calibration data.
\end{abstract}

%==========================================================================
% 2. Introduction
%==========================================================================
\section{Introduction}

\noindent\textbf{A held-out budget makes capacity a statistical choice.}
Post-hoc calibration is often estimated from only a few thousand
held-out examples after fine-tuning, which makes calibrator capacity a
statistical choice, not merely an architectural one: a map with too few
degrees of freedom can leave structured miscalibration uncorrected, and
one with too many can fit noise in that same small split. Fine-tuned
vision transformers \citep{dosovitskiy2021vit} make this concrete. Under
modern recipes (AdamW~\citep{loshchilov2019adamw},
MixUp~\citep{zhang2018mixup}/CutMix~\citep{yun2019cutmix},
RandAugment~\citep{cubuk2020randaugment}, label
smoothing~\citep{szegedy2016rethinking}) our ViT-B/16 on CIFAR-100 reaches
mean top-1 $91.00$ yet retains $\mathrm{ECE}_{15} = 9.69$ (five seeds),
so post-hoc recalibration can be an important deployment-time layer when
fine-tuning leaves substantial residual miscalibration.

\begin{figure*}[t]
    \centering
    \includegraphics[width=0.80\textwidth]{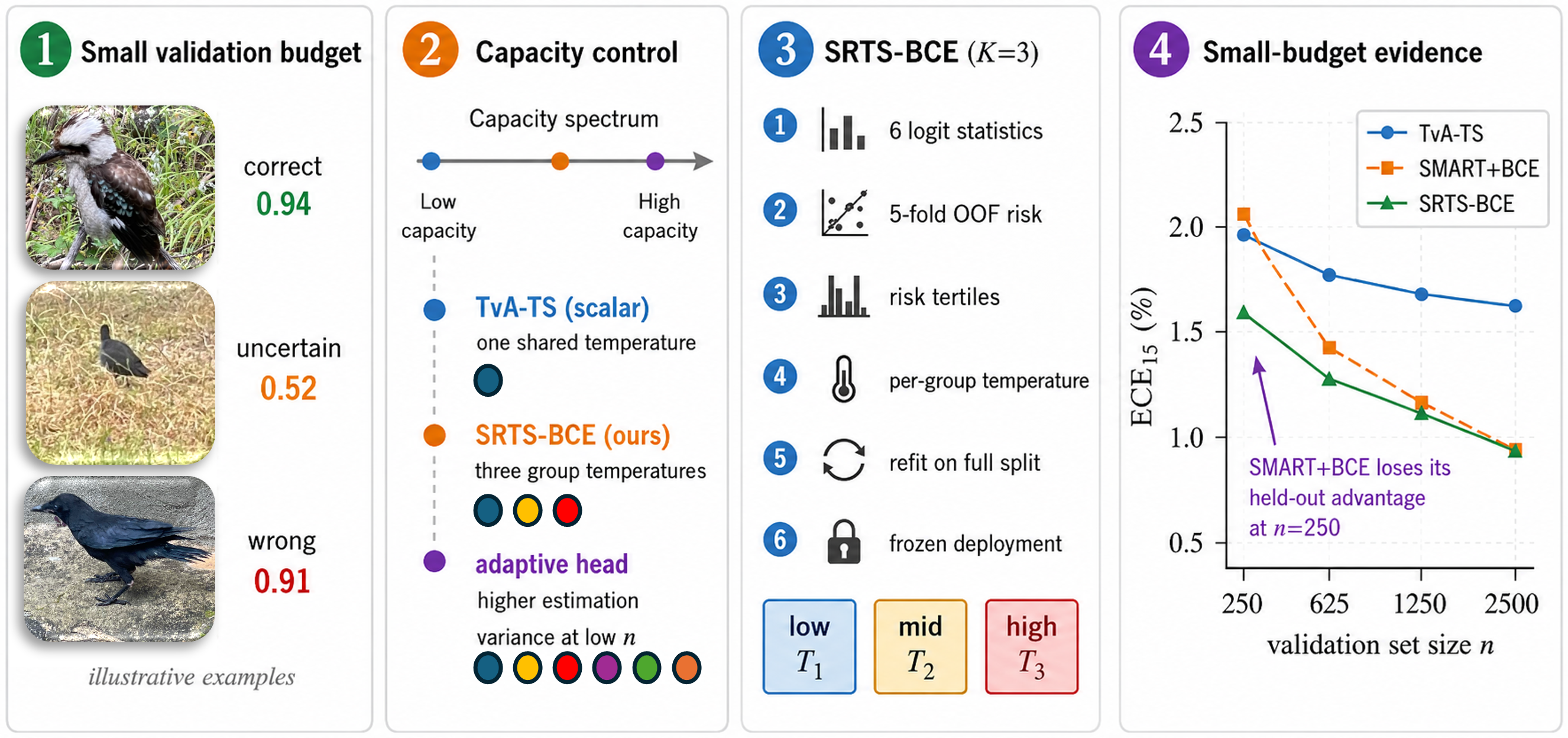}
    \caption{
    \textbf{Problem, capacity spectrum, method, and small-budget evidence.}
    (1)~A small validation set can leave a fine-tuned classifier
    miscalibrated (illustrative predictions, not benchmark samples).
    (2)~Calibration maps span a capacity spectrum: a single shared
    temperature (TvA-TS) can underfit and a sample-adaptive head carries
    higher estimation variance at small $n$; SRTS-BCE takes a low-capacity
    middle ground with three risk-conditioned group temperatures.
    (3)~Six logit statistics yield five-fold cross-fitted out-of-fold risk
    scores that define three equal-frequency risk tertiles
    ($T_1$/$T_2$/$T_3$), each fit with one top-label-BCE temperature; the
    router is refit on the full split for deployment, and each test sample
    is rerouted from its own logits with thresholds and temperatures fixed.
    (4)~Measured data (unlike the schematic panels 1--3): mean
    $\mathrm{ECE}_{15}$ vs.\ validation set size $n$ on CIFAR-100 /
    ViT-B/16 for TvA-TS, SMART+BCE, and SRTS-BCE (five seeds $\times$ $20$
    draws; protocol in Results); SMART+BCE loses its held-out advantage at
    the smallest budget $n{=}250$.
    }
    \label{fig:srts-methodology-overview}
\end{figure*}

\noindent\textbf{Existing temperature-scaling methods conflate two independent design
choices.} Post-hoc calibration is usually posed as one choice: a scalar
temperature, a group-wise variant, or a sample-adaptive head. That framing
hides a more basic issue. On the small validation splits available after
fine-tuning, calibration exhibits a \emph{capacity-control} trade-off: the calibrator
must be expressive enough to correct structured residual miscalibration, yet
constrained enough not to learn idiosyncrasies of a few thousand examples. We
therefore separate two axes. The \emph{calibration objective} distinguishes
classical TS, which minimises validation NLL, from top-label BCE, which
targets the probability assigned to the predicted class
(\citealp{lecoz2024tva}). The \emph{adaptive capacity} is the granularity at
which temperature may vary with the input: one global $T$, a low-capacity
reliability-conditioned map, or an unrestricted per-sample $T(x)$. Two
further axes recur throughout: the \emph{conditioning signal} (entropy,
margin, or learned correctness risk) and the \emph{fit/deploy discipline}
(cross-fitted and frozen vs.\ a directly trained head). Comparisons among
these methods often vary both the fitting objective and the amount of
adaptivity, making their individual effects difficult to separate.

\noindent\textbf{Unrestricted heads did not benefit from the extra capacity
at this budget.} On
${\approx}2{,}500$-sample validation splits, every evaluated unrestricted
logit- and feature-input temperature head
was less stable than scalar or constrained calibration, staying at
$\mathrm{ECE}_{15} \approx 5.7$--$11$ under both an in-sample NLL sweep and a
nested BCE-selection protocol (Results) --- more degrees of freedom than the
evaluated validation protocol can estimate reliably.

\noindent\textbf{A low-capacity middle ground between scalar and
sample-adaptive scaling.} We close this gap by conditioning a scalar
temperature map on a cross-fitted, low-dimensional logistic risk score.
Our deployed instance, \textbf{SRTS-BCE}, routes each validation sample
into one of $K{=}3$ equal-frequency risk groups and fits one
top-label-BCE temperature per group --- $10$ fitted scalars, two
orders of magnitude fewer than an unrestricted head, with TvA-TS as its
exact $K{=}1$ limit. The Method section gives a bias--variance account of
\emph{when} this helps: pooling bias grows with how strongly the residual
tracks the routing signal, variance grows with $K/n$, and grouping pays
off only when the former dominates.

\noindent\textbf{Contributions.} Our contributions are threefold.
\begin{itemize}
\item \textbf{A cross-fitted risk-conditioned calibrator.} SRTS-BCE routes
each sample through an out-of-fold correctness-risk router into one of
$K{=}3$ groups, each with its own top-label-BCE temperature
($10$ coefficients, argmax-preserving, fit once, reducing to
TvA-TS at $K{=}1$).

\item \textbf{Small-budget stability across architectures and datasets.}
Across three CIFAR-100 backbones and two protocol-frozen Tiny-ImageNet
replication cells, the relative benefit of low-capacity conditioning
increases as the calibration budget shrinks, with hierarchical inference
over seeds and calibration draws (Results, RQ1/RQ1b).

\item \textbf{Controls on objective, routing signal, and map family.} We vary the
objective, the routing signal, and the map family one at a time, and
characterise where the improvement holds across backbones, datasets, and
clean-to-shift transfer: it concentrates where the clean calibration
residual is structured and remains informative after shift (Discussion;
Tab.~\ref{tab:regime_winloss}).
\end{itemize}

%==========================================================================
% 3. Related work
%==========================================================================
\section{Related Work}
\label{sec:related}

\noindent\textbf{Scalar and top-label calibration.}
Reliability diagrams and ECE \citep{naeini2015bbq,guo2017calibration}
quantify the confidence--accuracy gap, and smoothECE
\citep{blasiok2023smooth} reduces its binning dependence; large-scale
comparisons also evaluate calibrators on proper scoring rules
\citep{berta2026calarena}. Temperature scaling (TS)
\citep{guo2017calibration} fits one scalar temperature on a held-out split
and preserves the top-1 prediction, and TvA-TS \citep{lecoz2024tva} keeps
that scalar capacity while fitting a top-label BCE objective that targets
predicted-class confidence in many-class settings. A single temperature is
economical to estimate from a small split, but applies the same correction
to every example and cannot address miscalibration that varies with the
input.

\noindent\textbf{Adaptive temperature scaling.}
Sample-dependent heads predict an input-conditional temperature $T(x)$ from
logits or representations. Adaptive temperature scaling (SATS;
\citealp{joy2023sample}) and LTS-style local temperature heads
\citep{ding2021local} place relatively few restrictions on $T(x)$;
low-parameter maps instead condition it on a single reliability statistic
--- HTS \citep{balanya2024adaptive} on prediction entropy, proposed for
data-scarce calibration, SMART \citep{guo2026smart} on the top-logit gap,
and QaTS \citep{chakraborty2026qats} on the confidence quantile through a
two-parameter monotone map. These methods trace a range of conditional
capacity, from a nearly unrestricted per-sample temperature down to a two-
or three-parameter reliability map.

\noindent\textbf{Grouped and locally conditioned calibration.}
A separate line partitions the inputs and calibrates within regions:
piecewise-constant maps on confidence \citep{kumar2019verified}, locally
conditioned maps \citep{ding2021local}, ensembles of scalers
\citep{zhang2020mixnmatch}, and learned groupings such as Semantic-Aware
Grouping \citep{yang2023semantic}, which forms a soft feature-and-logit
grouping. Meta-Cal \citep{ma2021metacal} pairs a base calibrator with a
correctness-ranking model under coverage constraints, and Simi-Mailbox
\citep{seo2025simimailbox} groups graph nodes by confidence and topology
for group-specific temperatures. These methods share calibration
parameters within regions rather than learning an unrestricted sample-wise
temperature function.

\noindent\textbf{Relation to this work.}
SRTS-BCE lies between a global temperature and a per-sample head: it groups
examples by a cross-fitted correctness-risk score and fits one top-label
BCE temperature per group, with TvA-TS as its $K{=}1$ limit. It conditions
on a learned ordering over several logit statistics and applies one
temperature per $K{=}3$ risk group, retaining far less adaptive capacity
than a sample-wise head. We therefore study how this restricted conditional
capacity behaves as the calibration budget shrinks.

%==========================================================================
% 4. Background and problem setup
%==========================================================================
\section{Background and Problem Setup}

\noindent\textbf{Problem statement.} We study fit-once post-hoc calibration
for fine-tuned image classifiers --- primarily ViT-style backbones ---
under a small clean calibration split: given a trained classifier and a
stratified clean validation set of size $n{\approx}2{,}500$, fit an
argmax-preserving calibrator once from validation logits and labels; the
calibrator is then frozen and evaluated on held-out clean test data and,
for shift experiments, on corrupted test data without refitting,
corruption labels, or test-set model selection.

\noindent\textbf{Notation.} Let $f$ be a classifier mapping inputs $x \in \mathcal{X}$ to
logits $z(x) \in \mathbb{R}^C$ over $C$ classes. The softmax confidence is
$\hat{p}(x) = \mathrm{softmax}(z(x))$ and the predicted label is
$\hat{y}(x) = \arg\max_c z(x)_c$. The classifier is \emph{calibrated} if
$\mathbb{P}(\hat{y} = y \mid \max_c \hat{p}_c = p) = p$ for all $p \in [0,1]$.

\noindent\textbf{Post-hoc temperature scaling.} The standard post-hoc remedy
\citep{guo2017calibration} divides the logits by a learned scalar $T > 0$ before
the softmax: $\hat{p}_T(x) = \mathrm{softmax}(z(x)/T)$. Global TS fits a single
$T^*$ on a held-out validation split by minimising val NLL. Positive
scalar temperature maps preserve the predicted class by construction;
this covers TS, TvA-TS, SRTS-BCE, and every evaluated adaptive head
(softplus-positive outputs). We also verify empirically that no reported
post-hoc method changes top-1 accuracy
(Tab.~\ref{tab:phaseb_top1_audit}).

\noindent\textbf{Sample-adaptive temperature heads.} These replace the scalar $T$
with a predictor $T_{\phi}(x)$ trained on validation logits (SATS) or
frozen features (LTS-feature; \textbf{Unrestricted TempHead}, a trainable
$D_{\text{feat}} \to H \to 1$ MLP) --- expressive, but prone to validation
overfit at this scale (Results).

\noindent\textbf{Calibration metrics.} We report 15-bin expected calibration error
($\mathrm{ECE}_{15}$; \citealp{naeini2015bbq}), adaptive-binning ECE
($\mathrm{AdaECE}_{15}$; \citealp{nixon2019measuring}),
classwise ECE ($\mathrm{CW\text{-}ECE}_{15}$;
\citealp{kull2019dirichlet}), smECE~\citep{blasiok2023smooth},
negative log-likelihood (NLL),
and Brier score~\citep{brier1950}. Lower is better for all
(formal definitions: App.~\ref{app:metric-definitions}). We use
$\mathrm{ECE}_{15}$ as the primary deployment-facing summary because it is
the standard post-hoc calibration reporting axis; AdaECE, smECE, and
classwise ECE are reported to expose metric-dependent trade-offs.

\noindent\textbf{Risk-coverage metrics.} We additionally report AUROC of correctness as a
selective-classification quality measure \citep{geifman2017selective}, area
under the risk-coverage curve (AURC), and excess AURC over the optimal
selector (eAURC) \citep{geifman2019biasreduced}.

App.~\ref{app:motivating-diagnostics} visualises the structured residual
left by scalar TvA-TS.

%==========================================================================
% 4. Method
%==========================================================================
\section{Method: SRTS-BCE}
\label{sec:method}

Figure~\ref{fig:srts-methodology-overview} gives a schematic overview.
SRTS-BCE is the deployed piecewise-constant instance of low-capacity
risk-conditioned temperature scaling: a top-label BCE objective, a
correctness-risk router trained with cross-fitting (yielding out-of-fold,
OOF, risk scores on the calibration split), and one scalar temperature per
risk group. The router selects a temperature \emph{group}; it does not
directly output the calibrated probability. The deployed
configuration is fixed throughout the paper: logits-only signals,
six distinct logit statistics, $K{=}3$ equal-frequency risk groups,
and one scalar temperature per group. Unless noted (SRTS-NLL or
$K$-ablation rows), ``SRTS-BCE'' means this deployed $K{=}3$
configuration everywhere, tables included.

\noindent\textbf{Algorithm 1: SRTS-BCE.}
On clean $\{(z_i,y_i)\}_{i=1}^n$ let $\hat y_i{=}\arg\max_c z_{ic}$,
$b_i{=}\mathbf 1[\hat y_i{=}y_i]$, $r_i{=}1{-}b_i$, and let $\bar s_i$ be the
six standardised logit statistics (max prob, logit margin, prob margin,
entropy, logit norm, top logit) plus an intercept. Over $F{=}5$ stratified
folds, an $L_2$ logistic router (intercept unpenalised) minimising
$\ell(a,u){=}{-}a\log u{-}(1{-}a)\log(1{-}u)$ gives out-of-fold
$q^{\mathrm{OOF}}_i=\sigma((\hat\theta^{(-f(i))})^{\!\top}\bar s_i)$; with
$\tau_k=Q_{k/K}(\{q^{\mathrm{OOF}}_i\})$,
\begin{equation}
g_i=1+\sum_{k=1}^{K-1}\mathbf 1[q^{\mathrm{OOF}}_i\ge\tau_k],\quad K{=}3 .
\label{eq:risk-group}
\end{equation}
With $c_i(T)=\mathrm{softmax}(z_i/T)_{\hat y_i}$ and $\mathcal{G}_g=\{i:g_i=g\}$,
\begin{equation}
\hat T_g=\arg\min_{T\in[0.05,20]}\;\frac{1}{|\mathcal{G}_g|}
\sum_{i\in\mathcal{G}_g}\ell\bigl(b_i,c_i(T)\bigr),
\label{eq:group-temperature}
\end{equation}
using the pooled $T$ if $|\mathcal{G}_g|<50$. The router is refit on the full
clean split, giving $g(x)$ from $q_{\mathrm{full}}(x)$ and the frozen $\tau_k$;
the deployed map is
\begin{equation}
\widetilde p_c(x)=\exp\bigl(z_c(x)/\hat T_{g(x)}\bigr)\big/
\textstyle\sum_{j}\exp\bigl(z_j(x)/\hat T_{g(x)}\bigr).
\label{eq:deployed-map}
\end{equation}
Thresholds and temperatures are frozen after this clean fit; each test or
corrupted sample is rerouted from its own logits; no corrupted labels,
clean-test pairings, or test-set selection are used. Since positive scalar
rescaling preserves logit order,
$\arg\max_c\widetilde p_c(x)=\arg\max_c z_c(x)$.

\noindent\textbf{Capacity and limits.} The logits-only SRTS-BCE calibrator has
exactly $10$ fitted scalar parameters: three group temperatures and a
linear router over six distinct logit statistics plus an intercept
(deployment also stores two quantile thresholds and six standardisation
means/scales --- data moments, not fitted parameters). This sits
deliberately between scalar TvA-TS and unrestricted neural heads,
and at $K{=}1$ the same BCE objective recovers TvA-TS
\citep{lecoz2024tva} (NLL recovers TS-NLL). Fitting is neural-training-free
(three one-dimensional temperature fits plus one low-dimensional logistic
regression): ${\approx}31\times$ faster than SMART+BCE, with microsecond
per-sample inference (Tab.~\ref{tab:efficiency}).

\noindent\textbf{Why routing can help.}
SRTS-BCE inherits the TvA-BCE objective \citep{lecoz2024tva} unchanged and
adds only conditional pooling over low-capacity OOF risk groups. The BCE
gradient matches NLL on correct predictions and is amplified by
$\hat p_i/(1-\hat p_i)$ on confident errors. Top-label BCE is a proper binary
log-loss for the correctness of the predicted label, not a multiclass
distribution-calibration objective, so it does not directly minimise
$\mathrm{ECE}$; a top-label-Brier check shows the benefit is not specific to
the BCE loss (App.~\ref{app:router-causality}).

\noindent\textbf{Fixed-partition second-order BCE approximation.}%
\label{prop:routing-gain}
For a partition fixed independently of the fitted temperatures, with one
top-label-BCE temperature per equal-frequency group, let $B$ be the
between-group variance of the calibration gradient at the pooled optimum
(so $B{=}0$ when the partition is independent of that gradient) and
$\bar\sigma^2$ its within-group variance. Expanding the population BCE to
second order around the pooled optimum and neglecting higher-order terms in
the per-group temperature deviations, the expected grouped-over-pooled
test-BCE gain is
$\Delta\approx(2L'')^{-1}(B-\tfrac{K-1}{n}\bar\sigma^2)$; to this order,
grouping helps when
\begin{equation}
\frac{B}{\bar\sigma^2}\;>\;\frac{K-1}{n}.
\label{eq:routing-gain}
\end{equation}
The relation is local and approximate: it predicts a \emph{top-label-BCE}
change (not $\mathrm{ECE}$), conditions on the given partition, excludes
router- and threshold-estimation error, and shows neither router optimality
nor equal-capacity-map superiority (RQ2); full derivation:
App.~\ref{app:routing-theory}.
\noindent Compared with the generic in-sample optimization bound (which
holds even for a random partition), the population gain vanishes for a
gradient-independent partition, so random routing stays near the baseline and
larger $K$ raises the bar. The approximation motivates three empirical checks:
(P1)~random or shuffled routing is inert; (P2)~larger $K$ eventually hurts;
(P3)~grouped fitting can improve the fitted BCE objective only when the
partition separates calibration gradients --- whether that gain transfers to
$\mathrm{ECE}$ is empirical. The finite-sample threshold and the
between-group signal (BCE group optima differ, NLL ones do not) are in
App.~\ref{app:bsigma-empirical}.

\noindent\textbf{Fixed $K{=}3$.} We fix $K{=}3$ before evaluation as a stable
fixed low-capacity setting, not a per-dataset or test-selected optimum
(the $K$-ablation is in RQ2; hold-out $K$-selection is unreliable at this
budget, Discussion). Each component addresses one failure mode: BCE the
confidence--objective mismatch, router cross-fitting the in-sample routing
bias, $K{=}3$ the $O(K/n)$ estimation variance, logits-only signals
feature-head overfitting, and fixed thresholds shift deployability; the
piecewise-constant form is chosen because it is simpler to inspect and fit
(a matched continuous map is tested in RQ2).

%==========================================================================
% 5. Experimental setup
%==========================================================================
\section{Experimental Setup}

\noindent\textbf{ID clean evaluation --- modern fine-tuning recipe} (five seeds): ViT-B/16
on CIFAR-100, $100$ epochs AdamW \citep{loshchilov2019adamw} with cosine
warmup, RandAugment \citep{cubuk2020randaugment} $+$ RandomErasing
\citep{zhong2020random} $+$ MixUp \citep{zhang2018mixup} / CutMix
\citep{yun2019cutmix} $+$ label smoothing \citep{szegedy2016rethinking}
$0.1$. Mean top-1 $91.00 \pm 0.46$, pre-TS
$\mathrm{ECE}_{15} = 9.69 \pm 1.10$ --- the recipe itself does not remove
the calibration error.

\noindent\textbf{Validation split.} We use a $5\%$ stratified calibration
split (${\approx}2{,}500$ samples; ${\approx}25$ per class on CIFAR-100), fit
once, with no test-set selection.

\noindent\textbf{Baselines.} We compare against scalar TS-NLL and TvA-TS;
the unrestricted neural heads LTS-feature, SATS-Logit, Feature-SATS, and
Frozen-feature UTempHead under the same post-hoc protocol; and the
low-parameter maps HTS and SMART. To separate capacity from the fitting
objective, HTS and SMART are also fit with top-label BCE (HTS-NLL/HTS-BCE
and SMART+BCE), matching the objective of SRTS-BCE and TvA-TS. HTS and QaTS
use equation-faithful reimplementations.

\noindent\textbf{Fixed-$K$ cross-regime evaluation.} The flagship
CIFAR-100 / ViT-B/16 evaluation uses five fine-tuning seeds; unless
marked otherwise, cross-backbone, mechanism-ablation, and shift
diagnostics use three seeds. Beyond the flagship cell we evaluate
CIFAR-100, CIFAR-10, and Tiny-ImageNet with ViT-B/16, DeiT-S
\citep{touvron2021deit}, and Swin-T \citep{liu2021swin}, plus two
ConvNets (ResNet-50 \citep{he2016resnet}, RegNetY-1.6GF
\citep{radosavovic2020regnet}) on CIFAR-100/10 under the identical
unretuned recipe. All SRTS-BCE entries in the main cross-regime
table use the same fixed $K{=}3$; no per-cell
$K$-selection or test-set selection is used.

\noindent\textbf{Clean-to-corruption shift evaluation --- CIFAR-100-C.} For the
main shift benchmark, calibrators are fit only on clean CIFAR-100
validation logits from the ID clean evaluation and applied unchanged to
CIFAR-100-C
\citep{hendrycks2019benchmarking}
(19 corruptions $\times$ 5 severities $\times$ 3 seeds). Metrics are
computed per corruption/severity cell, averaged over 95 cells per seed,
and then averaged over seeds. SRTS-BCE uses deployable rerouting: each
corrupted sample is routed using its own corrupted logits through the
clean-fitted risk router and clean-validation thresholds.

\noindent\textbf{Matched-parameterisation protocol (RQ2).} We compare
temperature maps sharing the OOF risk score, the top-label BCE
objective, and the fitting discipline; PWLinear-3 additionally matches
SRTS-BCE's fitted router and temperature-map parameter count --- three
fitted anchor temperatures vs.\ three fitted group temperatures
(stored, non-fitted objects differ: three anchor locations vs.\ two
group thresholds). Margin-K3 uses raw logit-margin tertiles.
All maps are frozen from clean validation. Uncertainty: paired
bootstraps on clean data; on CIFAR-100-C the three-tier cell /
corruption-type / hierarchical bootstrap with the hierarchical interval
primary (App.~\ref{app:matched-capacity}).

%==========================================================================
% 6. Results
%==========================================================================
\section{Results}
\label{sec:results}

\subsection{RQ1: Does SRTS-BCE improve over scalar and unrestricted calibration?}
\label{sec:results-protocol-b}

% RQ1 main clean table. Deployable-calibrator rows are the FIVE-seed
% flagship (seeds 0-4; provenance results/matched_capacity/20260724_150636/
% fiveseed_c100vit.json). The SATS-Logit reference row is the archived
% three-seed unrestricted-head result (separate experiment; no seed 3/4
% dump). Full metric set (Delta_TvA, CW-ECE) and further adaptive-head
% rows are in the supplementary full-metrics table.
% Params column = fitted scalar parameters (router + temperature map);
% stored thresholds/standardisation moments are data moments (Method).
\begin{table*}[t]
\centering
\small
\setlength{\tabcolsep}{3.5pt}
\begin{tabular}{lrrrrrrr}
\toprule
Method & Params & NLL & TopBCE & $\mathrm{ECE}_{15}$ & AdaECE$_{15}$ & smECE & eAURC \\
\midrule
CE / NoTS & 0
  & $0.4142 \pm 0.0204$ & $0.2633 \pm 0.0078$ & $9.69 \pm 1.10$
  & $9.65 \pm 1.15$ & $9.63 \pm 1.14$ & $0.0165 \pm 0.0041$ \\
TS-NLL & 1
  & $0.3309 \pm 0.0308$ & $0.2050 \pm 0.0143$ & $1.94 \pm 0.65$
  & $2.10 \pm 0.62$ & $1.97 \pm 0.52$ & $0.0139 \pm 0.0032$ \\
HTS-NLL & 2
  & $0.3311 \pm 0.0310$ & $0.2054 \pm 0.0142$ & $2.07 \pm 0.51$
  & $2.26 \pm 0.34$ & $2.13 \pm 0.38$ & $0.0140 \pm 0.0038$ \\
SRTS-NLL & 10
  & $\mathbf{0.3271 \pm 0.0275}$ & $0.1994 \pm 0.0102$ & $1.79 \pm 0.62$
  & $1.77 \pm 0.61$ & $1.81 \pm 0.55$ & $\mathbf{0.0113 \pm 0.0011}$ \\
TvA-TS ($K{=}1$) & 1
  & $0.3321 \pm 0.0296$ & $0.2041 \pm 0.0135$ & $1.65 \pm 0.44$
  & $1.77 \pm 0.69$ & $1.71 \pm 0.45$ & $0.0142 \pm 0.0033$ \\
HTS-BCE & 2
  & $0.3325 \pm 0.0304$ & $0.2041 \pm 0.0135$ & $1.63 \pm 0.25$
  & $1.51 \pm 0.31$ & $1.58 \pm 0.22$ & $0.0146 \pm 0.0039$ \\
SMART+BCE & 49
  & $0.3312 \pm 0.0290$ & $0.1979 \pm 0.0092$ & $\mathbf{0.95 \pm 0.29}$
  & $1.02 \pm 0.29$ & $1.16 \pm 0.22$ & $0.0123 \pm 0.0015$ \\
\textbf{SRTS-BCE} & 10
  & $0.3297 \pm 0.0257$ & $\mathbf{0.1967 \pm 0.0083}$ & $0.96 \pm 0.32$
  & $\mathbf{0.96 \pm 0.33}$ & $\mathbf{1.13 \pm 0.22}$ & $\mathbf{0.0113 \pm 0.0011}$ \\
\midrule
SATS-Logit & 13{,}057
  & $0.8128 \pm 0.0391$ & $0.4385 \pm 0.0337$ & $6.04 \pm 0.23$
  & $5.39 \pm 0.23$ & $7.47 \pm 0.16$ & $0.0281 \pm 0.0035$ \\
\bottomrule
\end{tabular}
\caption{\textbf{RQ1: objective and capacity on ViT-B/16 / CIFAR-100}
(\%, mean$\pm$std over \emph{five} seeds; top-1 $91.00 \pm 0.46$; fit on
the validation split, evaluated on held-out test; metric definitions,
including TopBCE, in App.~\ref{app:metric-definitions}). Params:
fitted scalars. TvA-TS $\equiv$ TS-BCE ($K{=}1$); HTS
\citep{balanya2024adaptive} is fitted with its original NLL objective and
with this paper's top-label BCE (Setup, Baselines). \textbf{Bold}: best per
column among deployable calibrators. SMART+BCE has the lowest clean
$\mathrm{ECE}_{15}$ point estimate ($0.9483$ vs.\ $0.9559$); the paired
difference is unresolved (hierarchical CI $[-0.19,+0.12]$). Full metric set and
further adaptive-head rows: Tab.~\ref{tab:sota_3seed_full_metrics}.}
\label{tab:sota_3seed_summary}
\end{table*}

\noindent\textbf{Objective alignment enables useful routing.} TS-NLL
($1.94 \pm 0.65$) $\to$ TvA-TS ($1.65 \pm 0.44$): a $0.29$ pp gain from
the BCE objective alone at unchanged capacity ($K{=}1$).

\noindent\textbf{SRTS-BCE:} $\mathrm{ECE}_{15} = 0.96 \pm 0.32$ ---
$0.70$ pp below TvA-TS on every seed and $0.98$ pp below TS-NLL
(seed$\times$sample hierarchical CI vs.\ TvA-TS $[-0.88,-0.36]$;
sample-only sensitivity views: App.~\ref{app:best-per-criterion});
the $K{=}1 \to K{=}3$ gain is over $2\times$ the objective gain.

\noindent\textbf{The clean cell ties the strongest peer.} SMART+BCE (the
SMART logit-gap head \citep{guo2026smart}, BCE-trained) also reaches
$\mathrm{ECE}_{15}=0.95$ (SoftECE-trained original $1.21$;
Tab.~\ref{tab:sota_3seed_full_metrics}). The clean cell therefore
does not distinguish SRTS-BCE from the strongest objective-matched
peer: the hierarchical CI is $[-0.19,+0.12]$, and across the nine
objective-matched transformer cells the gap is at most $0.23$ pp
(Tab.~\ref{tab:cross_dataset_full_metrics_fixedk3}).

\subsection{RQ1b: Does capacity become decisive as the calibration budget shrinks?}
\label{sec:results-budget}

\begin{figure*}[t]
\centering
\includegraphics[width=\textwidth]{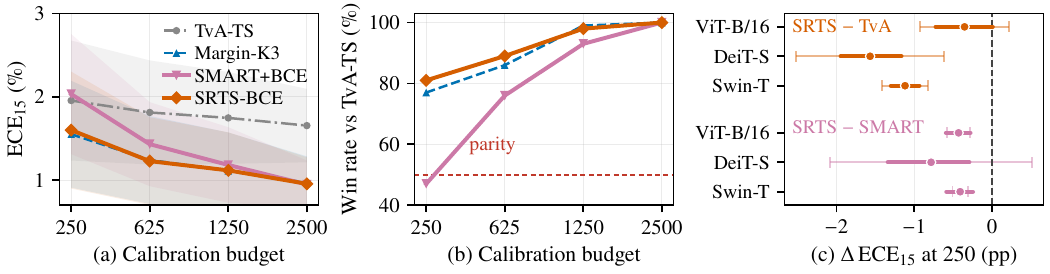}
\caption{\textbf{Small-budget stability (RQ1b).}
Calibrators are fit on $20$ stratified calibration resamples per
training seed and evaluated on the fixed test set. (a)~CIFAR-100 / ViT-B/16 mean
$\mathrm{ECE}_{15}$ vs.\ budget ($\pm$std bands): the low-capacity maps
(SRTS-BCE, Margin-K3) degrade gracefully; SMART+BCE rises above the
scalar at $250$. (b)~win rate against
TvA-TS: SRTS-BCE stays high ($81\%$ at $250$) while SMART+BCE falls to
parity ($47\%$). (c)~cross-backbone $\Delta\mathrm{ECE}_{15}$ at $250$
(thick: seed$\to$draw $95\%$ CI; thin: seed-cluster interval): vs.\
SMART+BCE zero is excluded on all three backbones; vs.\ TvA-TS on DeiT-S
and Swin-T, with flagship ViT-B/16 directional (interval crosses zero).
Full tables, the Tiny-ImageNet/DeiT-S and Swin-T
replications, and the Tiny-ImageNet/ViT-B/16 boundary:
App.~\ref{app:budget-resampling}.}
\label{fig:budget_resampling}
\end{figure*}

\noindent\textbf{Small-budget stability separates the low- and
higher-capacity maps} (Fig.~\ref{fig:budget_resampling}). We resample the
calibration split at budgets $\{250,625,1250,2500\}$; draws within a seed
share the model and test set, so inference is a two-level (seed$\to$draw)
hierarchical bootstrap, reported alongside a conservative seed-cluster
interval (between-seed spread only). At $n{=}250$ SRTS-BCE beats the
higher-capacity SMART+BCE on all three CIFAR-100 backbones: the
SRTS$-$SMART interval $[-0.59,-0.28]$ excludes zero (every seed favours
SRTS-BCE; the seed-cluster interval also excludes zero on ViT-B/16 and
Swin-T, but not DeiT-S), and on ViT-B/16 SRTS-BCE is lower on $81\%$ of
draws while SMART+BCE beats the scalar on only $47\%$. The comparison with
TvA-TS is weaker: the $250$-sample interval excludes zero on DeiT-S and
Swin-T, while on the flagship ViT-B/16 it is \emph{directional} --- four of
five seeds favour SRTS-BCE but the interval $[-0.72,+0.01]$ includes zero
--- becoming interval-supported by $n{=}625$
(Tab.~\ref{tab:budget_hierarchical}).

\noindent\textbf{The separation is not specific to one metric or dataset.}
The AdaECE$_{15}$ and official smECE seed$\to$draw intervals give the same
SRTS$-$SMART ordering on all three CIFAR-100 backbones
(Tab.~\ref{tab:budget_multimetric}), and the advantage also holds
over the nearest low-capacity neighbour, entropy-conditioned HTS
\citep{balanya2024adaptive}, on identical draws (SRTS$-$HTS-BCE at $250$
excludes zero on all three backbones; Tab.~\ref{tab:hts_comparison}).
A protocol-frozen Tiny-ImageNet replication (DeiT-S, Swin-T --- cells frozen
from previously reported full-budget direction, not from any small-budget
result) reproduces both $250$-sample separations beyond CIFAR-100, also
under AdaECE$_{15}$ and smECE
(Tabs.~\ref{tab:tiny_replication}--\ref{tab:tiny_multimetric}).
Swin-T shows the clearest reversal: SMART+BCE is better at the largest
evaluated $2{,}500$-sample budget, whereas SRTS-BCE is better at $n{=}250$;
on CIFAR-100 the full-budget gap closes on ViT-B/16 and Swin-T while DeiT-S
retains a smaller SRTS-BCE advantage.

\noindent\textbf{The reversal reflects a finite-sample capacity effect, not
a fixed method ordering.} On identical draws a frozen matched-map test finds
PWLinear-3 indistinguishable from SRTS-BCE at $250$ on all three backbones
(linear and spline maps on the same risk score are worse;
Tab.~\ref{tab:p1_matched_maps}), so the result is not specific to
discrete grouping. Tiny-ImageNet/ViT-B/16 shows no routing benefit at any
evaluated budget, so shrinking it only exposes the finite-budget variance
cost (App.~\ref{app:budget-resampling}). Together these cells indicate
that low capacity helps when a structured residual is present but the
calibration data are too limited to estimate the higher-capacity
alternative reliably; SRTS-BCE also fits ${\approx}31\times$ faster than
SMART+BCE (Tab.~\ref{tab:efficiency}).

\noindent\textbf{SRTS-NLL (ablation)} reaches only $1.79 \pm 0.62$ with
the same infrastructure --- the objective is the bottleneck --- and the
\textbf{unrestricted adaptive heads} stay at
$\mathrm{ECE}_{15} = 6.04$--$6.71$, a $3$--$4\times$ regression vs.\ scalar TS
on every seed; this capacity-overfit failure replicates across a
single-checkpoint diagnostic, a val-NLL sweep, and the objective-matched BCE
nested check (App.~\ref{app:fairness-sweep}).

\subsection{RQ2: Which components of SRTS-BCE matter?}
\label{sec:results-rq2}

\noindent We next isolate the roles of routing signal, group count, and map
family while holding the fitting objective and, where applicable, parameter
count fixed.

\begin{figure}[t]
\centering
\includegraphics[width=\columnwidth]{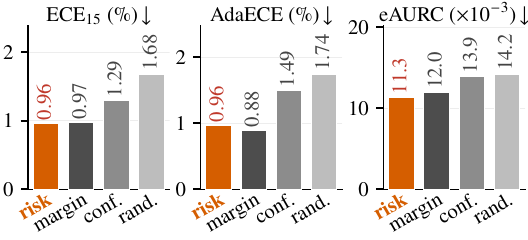}
\caption{\textbf{Routing-rule ablation} ($K{=}3$, BCE objective;
five-seed ViT-B/16 / CIFAR-100). Conditioning on a
genuine reliability signal matters: risk and margin beat
confidence-only and random partitions, but are not distinguished from
each other (risk lower on $\mathrm{ECE}_{15}$/eAURC, margin on AdaECE;
paired CI includes zero); under NLL every rule collapses to one band
(App.~\ref{app:router-causality}).}
\label{fig:router_causality}
\end{figure}

\noindent\textbf{Routing signal.} Fixing $K{=}3$ and varying only the routing
rule (Fig.~\ref{fig:router_causality}), conditioning on a genuine reliability
signal matters: risk and margin routing beat confidence-only and
random partitions, and arbitrary grouping is inert. Under NLL every rule
collapses to one band ($\mathrm{ECE}_{15}\in[1.71,1.82]$;
App.~\ref{app:router-causality}). Margin is the strongest of SRTS-BCE's
six signals and recovers most of the effect; the paired CI
$[-0.51,+0.15]$ does not distinguish the learned router from margin-only
routing (App.~\ref{app:margin-not-sufficient}); SRTS-BCE extends
margin routing to a learned, cross-fitted router instead of one
hand-picked statistic.

% from results/matched_capacity/20260724_150636 (six-signal regen).
% Regenerate with --write; verify with --check. Do not hand-edit numbers.
\begin{table}[t]
\centering
\small
\setlength{\tabcolsep}{0.6pt}
\begin{tabular}{lrrrr}
\toprule
Method & C100 & Tiny & IN-100 & C100-C \\
\midrule
TvA-TS ($K{=}1$) & $1.65{\pm}0.44$ & $1.26{\pm}0.18$ & $1.75{\pm}0.15$ & $5.72{\pm}1.13$ \\
Margin-K3 & $0.97{\pm}0.32$ & $0.99{\pm}0.16$ & $1.83{\pm}0.30$ & $4.70{\pm}0.78$ \\
SMART+BCE & $0.95{\pm}0.29$ & $0.97{\pm}0.22$ & $1.71{\pm}0.26$ & $4.75{\pm}0.79$ \\
\midrule
LinearRiskTemp & $1.29{\pm}0.48$ & $1.35{\pm}0.08$ & $1.74{\pm}0.26$ & $4.89{\pm}0.76$ \\
PWLinear-3 & $1.01{\pm}0.32$ & $1.26{\pm}0.11$ & $1.86{\pm}0.26$ & $4.58{\pm}0.86$ \\
\textbf{SRTS-BCE (Ours)} & $0.96{\pm}0.32$ & $1.11{\pm}0.15$ & $1.84{\pm}0.22$ & $4.63{\pm}0.84$ \\
\bottomrule
\end{tabular}
\caption{\textbf{RQ2: matched-parameterisation comparison} ($\mathrm{ECE}_{15}$, mean${\pm}$std; columns are ViT-B/16 except IN-100, which is DeiT-S; C100-C: per-seed means over the 95 corruption cells, deployable protocol). Within each training seed all methods share the same fixed calibration split; all use the top-label BCE objective. Top block: conditioning signal varies. Bottom block: same OOF risk score, objective, and fitting discipline; map family varies (PWLinear-3 exactly matches SRTS-BCE's fitted router and parameter count). (Seeds: five on C100, three on Tiny and the protocol-frozen IN-100 subset. Full method set, five-seed IN-100, and three-tier CIs: Tabs.~\ref{tab:matched_capacity_full}, \ref{tab:in100}, \S\ref{app:matched-capacity}.)}
\label{tab:matched_capacity_main}
\end{table}

\noindent\textbf{Number of groups.} A BCE $K$-ablation
($\mathrm{ECE}_{15}=1.65/1.07/0.96/0.94/1.01$ at $K{=}1/2/3/5/10$) confirms
that $K{=}3$ captures the gain and $K{=}10$ degrades it. Holding signal and
capacity fixed, Tab.~\ref{tab:matched_capacity_main} varies only the map
family (lower block, which fixes the OOF risk score, objective, and fitting
discipline; PWLinear-3 matches SRTS-BCE's fitted router and temperature-map
parameter count exactly) against external peers (top block).

\noindent\textbf{Map result.} PWLinear-3, with the same fitted router and
parameter count as SRTS-BCE, is statistically indistinguishable from the
grouped map under the primary inference on both clean CIFAR-100 (hierarchical
CI $[-0.14,+0.13]$) and CIFAR-100-C ($[-0.01,+0.14]$), despite a slightly
lower shifted point estimate ($4.58$ vs.\ $4.63$). The linear map is less
effective ($1.29$/$4.89$); the spline matches the grouped map on the clean
cell yet converges toward the linear map under shift ($4.89$). The benefit
appears under both discrete and continuous maps; SRTS-BCE adopts the discrete
map because it is simpler to inspect, fit, and deploy, not because
discretisation is optimal. These results agree with P1--P2 of the
approximation.

\noindent\textbf{Objective replication (QaTS, HTS).} Two independent map
families reproduce the objective axis. QaTS \citep{chakraborty2026qats}
(equation-faithful reimplementation) trails SRTS-BCE on CIFAR-100 (clean CI
$[-0.51,-0.06]$) and CIFAR-100-C (all tiers exclude zero,
hierarchical $[-1.05,-0.13]$), and is unresolved on
Tiny-ImageNet/ViT-B/16 and IN-100/DeiT-S; its QaTS-BCE$\,{-}\,$QaTS-NLL gap
($1.35$ vs.\ $1.90$) reproduces the BCE-over-NLL improvement outside our
router, and HTS shows the same pattern (HTS-BCE $1.63$ vs.\ HTS-NLL
$2.07$ clean; $5.41$ vs.\ $6.96$ shifted). Full method set and tiers:
App.~\ref{app:matched-capacity}, Tab.~\ref{tab:hts_comparison}.

\subsection{RQ3: When does clean-fitted SRTS-BCE transfer?}
\label{sec:results-protocol-c}

On \textbf{CIFAR-100-C} (19 corruptions $\times$ 5 severities) every
corrupted sample is rerouted from its own logits (protocol: Setup).
Per-cell then across-seed means: NoTS $9.13$; TS-NLL $6.75$; TvA-TS
$5.72$; \textbf{SRTS-BCE} $\mathbf{4.63}$; SMART+BCE $4.75$; SATS-Logit
$15.45$.

\noindent\textbf{SRTS-BCE attains a lower mean CIFAR-100-C ECE than
SMART+BCE under deployable rerouting; the $0.12$-pp difference is
directional, not two-sided significant}: under the primary
seed$\times$corruption$\times$severity hierarchical bootstrap the interval
includes zero ($[-0.33,+0.03]$, $\Pr_{\mathrm{boot}}(\Delta<0)=0.93$; more
conservative than the cell and corruption-type tiers,
App.~\ref{app:matched-capacity}). Among matched maps PWLinear-3 is
slightly lower (RQ2). Shift eAURC keeps the ordering; per-corruption gains
anti-correlate with miscalibration magnitude ($\rho{=}{-}0.68$) --- the
clean-fitted router transfers under mild shift, degrades under severe
(App.~\ref{app:routing-heterogeneity}).

\begin{figure}[t]
\centering
\includegraphics[width=\columnwidth]{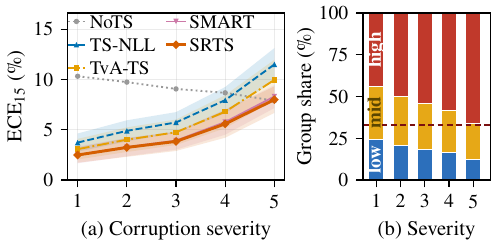}
\caption{\textbf{CIFAR-100-C shift under deployable rerouting.}
(a) Among the plotted methods, SRTS-BCE has the lowest mean
$\mathrm{ECE}_{15}$ point estimate at every severity ($\pm$std; CIs in
text).
(b) corrupted examples shift into higher-risk groups; dashed:
clean-test high-risk share ($32.7\%$). Legend abbreviations: SMART denotes
SMART+BCE; SRTS denotes SRTS-BCE.}
\label{fig:srts-protocol-c-severity}
\end{figure}

\noindent\textbf{Risk-group redistribution tracks the severity trend}
(Fig.~\ref{fig:srts-protocol-c-severity}): the high-risk share grows
$44\%{\to}66\%$. \emph{Temperatures stay fixed; the group assignment
adapts} --- forcing equal test-time occupancy \emph{worsens} ECE
monotonically ($+2.3$\,pp at severity 5;
App.~\ref{app:routing-heterogeneity}), so the gain comes from
test-time rerouting on shifted logits.

\noindent\textbf{Representation and training boundary.}
\label{sec:results-cross-dataset}
Across the 16 fixed-$K{=}3$ cross-regime cells
(Tab.~\ref{tab:cross_dataset_ece_summary}), SRTS-BCE beats TvA-TS
in $12$ cells, ties one CIFAR-10 ConvNet cell, and loses all three
fine-tuned IN-100 cells; vs.\ SMART the ordering is backbone- and
objective-dependent (Tab.~\ref{tab:cross_dataset_full_metrics_fixedk3}).
The two ConvNets replicate the structured-residual pattern beyond transformers.

\noindent\textbf{When does routing transfer?} A residual is
\emph{structured and transferable} when a clean-fitted reliability ordering
separates groups with distinct calibration optima and retains that ordering
on deployment data; routing gains concentrate where this holds, the scalar
is preferable where it does not (Tab.~\ref{tab:regime_winloss}).
Under natural shift the objective transfers on fine-tuned ImageNet-V2 but
calibration harms on pretrained ImageNet-R/-A (App.~\ref{app:natural-shift}).

%==========================================================================
% 7. Limitations
%==========================================================================
\section{Discussion and Limitations}
\label{sec:limitations}

\noindent\textbf{When conditioning helps.} Residual magnitude alone
does not predict the gain: in low-data and pretrained regimes a routing
signal may exist for the fitted objective yet fail to transfer to ECE.
Objective alignment and capacity control act independently: BCE improves
the scalar anchor where NLL does not; low-dimensional conditioning adds a
further gain only under that objective. The budget-dependent capacity trade-off from RQ1b
holds across CIFAR-100 and Tiny-ImageNet: the gain is positive only
where a transferable residual dominates. This describes where the method
works; it is not a rule for selecting capacity from validation data alone
(App.~\ref{app:budget-resampling},\S\ref{app:routing-heterogeneity}).
The fitted-objective criterion indicates whether routing has signal;
ECE transfer also needs the BCE-to-ECE link and router stability
(App.~\ref{app:bsigma-empirical}).

\noindent\textbf{What improves? Primarily top-label confidence.}
NLL and Brier are similar across calibrators, the conditioned maps adding
a modest ${\approx}0.005$ TopBCE gain
(Tab.~\ref{tab:phi_proper_scores}). SRTS-BCE lowers HCFP
(high-confidence errors); selective risk at $80/90/95\%$ coverage is
nearly unchanged vs.\ TvA-TS (within $0.04$ pp clean, $0.03$ pp shifted)
--- confidence alignment, not lower selective error. It reduces the
confidence gap vs.\ the scalar at two of three pre-specified thresholds
on clean and shifted data and survives coverage matching
(App.~\ref{app:deployment-metrics},
Tabs.~\ref{tab:decision_thresholds}--\ref{tab:matched_coverage}).

\noindent\textbf{Which components are essential?}
\looseness=-1
Margin-only routing and a matched continuous map each recover most of the
effect ($[-0.51,+0.15]$), and SRTS-BCE ties SMART+BCE on clean CIFAR-100
($[-0.19,+0.12]$), so the gain does not depend on the learned router or
the discrete map. Selecting the capacity from validation data alone remains
out of reach: the 1-SE selector still fails on the low-data regime
(App.~\ref{app:selector}).

%==========================================================================
% 8. Conclusion
%==========================================================================

\noindent\textbf{Conclusion.}
\looseness=-1
Where conditional calibration is useful, the preferred map capacity changes
with the amount of calibration data. SRTS-BCE provides a simple low-capacity
instance of this trade-off; selecting capacity reliably from validation data
alone remains open. Our results suggest that post-hoc calibrator capacity is
better treated as a finite-sample statistical choice than as a fixed
architectural preference.

%==========================================================================
% Bibliography
%==========================================================================
\bibliography{references}

%==========================================================================
% Appendix (supplementary material)
%==========================================================================
\clearpage
\setcounter{secnumdepth}{1}
\renewcommand{\thesection}{\Alph{section}}
\setcounter{section}{0}
% Top-align floats on float-only pages (see original layout note): fix the
% top offset and inter-float gap so float pages begin at the normal top text
% boundary; leftover space collects at the bottom.
\makeatletter
\setlength{\@fptop}{0pt}
\setlength{\@fpsep}{20pt plus 0fil}
\setlength{\@dblfptop}{0pt}
\setlength{\@dblfpsep}{20pt plus 0fil}
\makeatother

\onecolumn\twocolumn[\centering{\Large\bf Appendix}\vspace{1.0em}]

This appendix provides implementation details, additional ablations,
complete statistical analyses, distribution-shift results, and extended
experiments supporting the main text. SRTS-BCE is a top-label-BCE
calibrator with an out-of-fold risk router, fixed $K{=}3$ groups, and one
scalar temperature per group.

Section~A gives implementation details and metric definitions.
Sections~B--D examine the fitting objective, the map capacity, and the
routing signal in turn. Section~E reports the calibration-budget
experiments and the Tiny-ImageNet replication. Section~F covers
distribution shift, including the deployable CIFAR-100-C protocol and the
natural-shift boundaries. Section~G reports cross-backbone results,
Section~H the negative results and extended limitations, Section~I the
second-order BCE approximation, and Section~J the full result tables.

A code-and-data artifact reproduces the primary clean result, the $n{=}250$
budget resampling, and the HTS comparison from packaged logits with fixed
seeds, and reconstructs the CIFAR-100-C table from frozen per-run rows.
NLL-objective results, NLL-recovery analyses, and single-checkpoint
diagnostics support the main clean and corruption-shift comparisons rather
than forming part of them; each is marked where it appears.

%==========================================================================
% A. Reproducibility and implementation notes
%==========================================================================

\section{Reproducibility and Implementation Notes}
\label{app:reproducibility}

\noindent\textbf{Training details (ID clean evaluation recipe).} All
fine-tuned backbones use the identical, fixed recipe: AdamW,
learning rate $5{\times}10^{-5}$, weight decay $0.05$, batch size $64$,
$100$ epochs with $5$ warmup epochs (warmup start
$10^{-6}$) and cosine decay to $0$; MixUp $\alpha{=}0.8$ / CutMix
$\alpha{=}1.0$ (batch mode), RandAugment, RandomErasing, and label
smoothing $0.1$. The flagship CIFAR-100 / ViT-B/16 evaluation uses
seeds 0--4; unless stated otherwise, the cross-backbone and
supplementary diagnostic suites use seeds 0--2. The controlled
single-checkpoint diagnostic uses the separate single-seed recipe
stated alongside it (\S\ref{app:nll-history}). Calibrators are fit on the
$5\%$ stratified validation split; all post-hoc fitting runs on CPU
(Table~\ref{tab:efficiency}).

\noindent\textbf{Router and temperature fitting details.} The OOF risk
scores use five-fold stratified cross-fitting (folds shuffled and seeded
per training seed). The router is an $L_2$-regularised logistic
regression (inverse regularisation strength $C{=}1.0$), fit by a
quasi-Newton method capped at $1{,}000$ iterations; the six signals are
z-scored with per-signal
mean and scale computed on the clean validation split (scale floor
$10^{-12} \to 1$). All temperatures (global TvA-TS and per-group) are
bounded one-dimensional minimisations of the top-label BCE over
$T \in [0.05, 20]$ with convergence tolerance $10^{-8}$. Groups with fewer than $50$ validation
samples fall back to the global TvA-TS temperature. After the OOF fit,
the router is refit once on the full validation split for deployment;
group thresholds are the OOF-risk tertile boundaries saved from clean
validation.

\noindent\textbf{Calibrator cost.} Learned parameter counts, fit wall-clock,
and per-sample inference cost are in Table~\ref{tab:efficiency}. Measurement
protocol: CPU limited to four threads; fit on the $n{=}2{,}500$ clean
CIFAR-100 / ViT-B/16 validation split; inference on the $10{,}000$-sample
test split; mean over three seeds.

% Calibrator efficiency (CPU, torch limited to 4 threads; fit on the
% n=2,500 clean CIFAR-100/ViT-B/16 validation split, inference on the
% 10,000-sample test split; mean over 3 seeds).
% Source: results/efficiency/run.log. Requires: \usepackage{booktabs}
% 2026-07-05: moved from Supplementary §A into the main Method section
% (caption compressed; the ~31x explanation lives in the Capacity para).
\begin{table}[b]
\centering
\small
\setlength{\tabcolsep}{5pt}
\begin{tabular}{lrrr}
\toprule
Method & Params & Fit (s) & Inference ($\mu$s/sample) \\
\midrule
TS-NLL       & 1       & 0.019 & 1.19 \\
TvA-TS       & 1       & 0.030 & 0.82 \\
SRTS-BCE $K{=}3$ & 10  & 0.137 & 7.00 \\
SMART+BCE    & 49      & 4.256 & 1.44 \\
SATS-Logit   & 13{,}057 & 5.066 & 1.55 \\
\bottomrule
\end{tabular}
\caption{\textbf{Calibrator cost} (CPU, 4 threads; fit on the validation
split, inference on $10{,}000$ test; 3-seed mean).}
\label{tab:efficiency}
\end{table}

\subsection{Metric definitions}
\label{app:metric-definitions}

All metrics are computed exactly as below (matching the released code).
TopBCE denotes the test-set value of the top-label binary cross-entropy
(the Method's fitting objective) evaluated on the calibrated
probabilities: with $b_i = \mathbf{1}[\hat y_i = y_i]$ and top-label
confidence $\hat p_i$, $\mathrm{TopBCE} = -\tfrac1n \sum_i [\, b_i \log
\hat p_i + (1-b_i)\log(1-\hat p_i)\,]$.
Notation: $N$ test samples with label $y_i$ and calibrated probability
vector $p_i \in \Delta^{C}$; prediction $\hat y_i = \arg\max_c p_{ic}$;
top-label confidence $c_i = \max_c p_{ic}$; correctness
$a_i = \mathbf{1}[\hat y_i = y_i]$. ECE-family metrics are reported
$\times 100$ (percentage points).

\noindent\textbf{Top-1 accuracy} $= \tfrac{100}{N}\sum_i a_i$.
\quad\textbf{NLL} $= -\tfrac{1}{N}\sum_i \log p_{i,y_i}$ (probabilities
clipped at $10^{-12}$).
\quad\textbf{Brier} $= \tfrac{1}{N}\sum_i \sum_c
\big(p_{ic} - \mathbf{1}[c{=}y_i]\big)^2$ (multi-class, summed over
classes).

\noindent\textbf{ECE$_{15}$} \citep{naeini2015bbq}: partition $[0,1]$ into
$15$ equal-width bins $B_b$ over $c_i$ and report
\begin{equation}
\mathrm{ECE}_{15} \;=\; \sum_{b=1}^{15} \frac{|B_b|}{N}\,
\Big|\overline{a}(B_b) - \overline{c}(B_b)\Big| \times 100,
\label{Seq:ece15}
\end{equation}
where $\overline{a}(B_b)$ and $\overline{c}(B_b)$ are the mean correctness
and mean confidence in bin $b$ (empty bins skipped; the last bin is closed
at $1$).

\noindent\textbf{AdaECE$_{15}$} \citep{nixon2019measuring}: the same
statistic with $15$ \emph{equal-mass} bins (samples sorted by confidence
and split into $15$ contiguous groups of $\lfloor N/15 \rfloor$-balanced
size).

\noindent\textbf{ClasswiseECE$_{15}$ (CW-ECE$_{15}$)}
\citep{kull2019dirichlet}: the one-vs-rest classwise ECE --- for each
class $c$, bin the per-class probability $p_{ic}$ over \emph{all}
samples into $15$ equal-width bins and accumulate
$\sum_b \tfrac{|B_b|}{N} |\overline{p_c}(B_b) -
\overline{\mathbf{1}[y{=}c]}(B_b)|$, then average over the $C$ classes.
At $C{=}100$ most per-class probabilities are near zero, so this metric
is small and only weakly discriminative among calibrated maps.

\noindent\textbf{smECE} \citep{blasiok2023smooth}: the smooth
calibration error, computed with the authors' released estimator
(reflected Gaussian kernel on the top-label confidence--correctness
pairs, bandwidth chosen by the estimator's self-consistent rule);
reported $\times 100$.

\noindent\textbf{AUROC (of correctness)} \citep{geifman2017selective}: the
area under the ROC curve of the confidence $c_i$ as a score for predicting
correctness $a_i$.

\noindent\textbf{AURC / eAURC} \citep{geifman2019biasreduced}: sort samples
by confidence (descending); at coverage $k/N$ the selective risk is the
error rate among the $k$ most-confident samples,
$r_k = \tfrac{1}{k}\sum_{j \le k} (1 - a_{(j)})$, and
$\mathrm{AURC} = \tfrac{1}{N}\sum_{k=1}^{N} r_k$. eAURC subtracts the
oracle value obtained by ordering all correct predictions first:
$\mathrm{eAURC} = \mathrm{AURC} - \mathrm{AURC}^{\ast}$.

\noindent\textbf{HCFP@$0.90$}: the fraction of \emph{wrong} predictions
whose top-label confidence is at least $0.90$ (``confidently wrong''
rate); reported only in this appendix's NLL-era diagnostics.

\noindent\textbf{Implementation note (router signals).} Historical
experiment artifacts constructed a seven-channel signal vector in which the
confidence channel appeared twice (a duplicated alias), and the archived
fitting code passed all seven columns to the $L_2$ logistic router. The
duplicate therefore changed the effective regularisation of the confidence
signal; it was not inert storage metadata. The released
implementation removes the duplicate and uses
exactly the six distinct statistics of Algorithm~1; every
reported router-dependent result in this paper was regenerated
with this six-signal implementation under identical seeds,
folds, draws, and bootstrap design. Point estimates change only slightly
(flagship clean five-seed mean $\mathrm{ECE}_{15}$ $0.9489 \to 0.9559$;
all six clean cell means move by $\le 0.010$ pp; CIFAR-100-C $4.626 \to
4.628$), but one marginal interval crosses zero: the flagship $n{=}250$
SRTS$-$TvA-TS seed$\to$draw interval moves from $[-0.714,-0.001]$ to
$[-0.724,+0.006]$, and the main text reports that comparison as
directional rather than interval-supported.

\noindent\textbf{Code and data availability.} A calibration-stage code
and data package accompanies this work and will be released publicly.
It \emph{includes}: the post-hoc calibration and evaluation
implementations used by the packaged analyses (the SRTS-BCE router; the
scalar, grouped, and continuous risk-conditioned temperature maps; HTS;
SMART+BCE; and metric/reconstruction code --- the unrestricted neural heads
are documented but not shipped); the packaged float32 evaluation logits for
the clean CIFAR-100 / ViT-B/16 cells (all five seeds), which re-evaluate the
main clean, flagship $n{=}250$ budget-resampling, and HTS results; the
frozen per-run result rows for CIFAR-100-C and for the cross-backbone and
Tiny-ImageNet replications; reference result files, a manifest with SHA-256
checksums, and a documented clean-environment reproduction script
that re-derives the main clean, $n{=}250$
budget-resampling, and HTS results from the packaged logits, and
\emph{reconstructs} the CIFAR-100-C table from the frozen per-run rows.
Reproduction is tolerance-based, not bitwise: each check prints reference,
reproduced, absolute difference, tolerance, and PASS/FAIL under
method-specific absolute ECE tolerances --- $0.011$ pp for the
deterministic post-hoc methods (float32 packaging and BLAS drift), $0.15$
pp for SMART+BCE only (its torch-trained MLP drifts across torch builds),
and $0.05$ pp for the CIFAR-100-C reconstruction --- stated identically in
the package documentation and the reproduction script. It \emph{excludes}, due to package-size limits: training code and checkpoints; the raw CIFAR-100-C corruption
logits; and the raw logits of the cross-backbone (DeiT-S/Swin-T) and other
regimes (Tiny-ImageNet, IN-100, pretrained, natural shift). Consequently the
CIFAR-100-C and cross-backbone tables are \emph{reconstructed from frozen
per-run rows, not re-evaluated from raw logits} in the package. It is
therefore a \emph{partial reproduction package} covering the
calibration-stage evidence; training configurations are documented in
this section for reimplementation from scratch. The table-level
aggregation and hierarchical inference for the Tiny-ImageNet follow-up
are independently auditable from the package: the complete per-seed/%
per-draw statistics, draw identifiers, subsample-RNG formula, frozen
protocol document, hierarchical-bootstrap code, and a verification
script that re-derives every
Table~\ref{tab:tiny_replication}--\ref{tab:tiny_multimetric} mean, win
rate, and CI from the per-draw rows are included; auditing the
per-draw rows themselves would require the excluded Tiny logits.

%==========================================================================
% B. Objective and capacity
%==========================================================================

\section{Objective and Capacity}
\label{app:objective-capacity}

This section collects evidence that the calibration benefit depends
jointly on the fitting objective and on the capacity of the temperature
map: the BCE $K$-ablation and the objective-matched capacity comparisons,
the Brier-objective routing and BCE-trained neural-head checks, and the
mixed-objective, single-checkpoint, and NLL analyses.

\textbf{BCE-objective $K$-ablation.} On the modern-recipe
ViT-B/16 / CIFAR-100 base (five seeds), SRTS-BCE $K \in \{1, 2, 3, 5, 10\}$ gives
mean $\mathrm{ECE}_{15} = 1.65 / 1.07 / \mathbf{0.96} / 0.94 / 1.01$. The
$K{=}1 \to K{=}3$ step closes most of the gain ($1.65 \to 0.96$, $-0.70$ pp);
$K{=}5$ has a marginally lower mean ECE ($0.94$) without stable evidence
of additional benefit, and $K{=}10$ degrades ECE ($1.01$). We select
$K{=}3$ as the BCE-objective default.

\subsection{Objective-conditional routing ablation}
\label{app:router-causality}

The objective-conditional routing table
(Table~\ref{tab:router_causality}) holds the $K{=}3$ constraint
fixed and varies the routing rule across four deployable variants (risk,
random, shuffled, confidence-only) under both NLL and BCE calibration
objectives, plus a BCE oracle diagnostic (the main text shows the $\mathrm{ECE}_{15}$ slice
as a bar chart; the full numbers are in
Table~\ref{tab:router_causality}). This subsection expands on it;
the excess-risk condition stated in the main Method is derived in
\S\ref{app:routing-theory}.

% Full objective-conditional routing ablation (supplement §B).
% BCE block five-seed; NLL block three-seed. K=3 group-wise temp,
% objectives. Same modern-recipe ViT-B/16 / CIFAR-100 base as main Table 2.
% The main paper shows the ECE15 slice as a grouped bar chart.
\begin{table*}[t]
\centering
\small
\setlength{\tabcolsep}{3.5pt}
\begin{tabular}{lrrrr}
\toprule
Routing rule & NLL & Brier & $\mathrm{ECE}_{15}$ & eAURC \\
\midrule
\multicolumn{5}{l}{\textit{NLL objective}} \\
\textit{SRTS-NLL K=3 risk}      & $0.3217 \pm 0.0274$ & $0.1333 \pm 0.0070$ & $1.7058 \pm 0.6483$ & $0.0115 \pm 0.0016$ \\
K=3 confidence                   & $0.3243 \pm 0.0302$ & $0.1340 \pm 0.0074$ & $1.8149 \pm 0.6923$ & $0.0131 \pm 0.0033$ \\
K=3 random equal-frequency       & $0.3241 \pm 0.0296$ & $0.1342 \pm 0.0076$ & $1.8181 \pm 0.7090$ & $0.0134 \pm 0.0033$ \\
K=3 shuffled risk-score          & $0.3239 \pm 0.0297$ & $0.1342 \pm 0.0076$ & $1.7970 \pm 0.7138$ & $0.0132 \pm 0.0034$ \\
\midrule
\multicolumn{5}{l}{\textit{BCE objective}} \\
TvA-TS $\equiv$ K=1 BCE           & $0.3321 \pm 0.0296$ & $0.1362 \pm 0.0081$ & $1.6542 \pm 0.4407$ & $0.0142 \pm 0.0033$ \\
\textit{SRTS-BCE K=3 risk (main)} & $0.3297 \pm 0.0257$ & $0.1348 \pm 0.0070$ & $\mathbf{0.9559 \pm 0.3188}$ & $\mathbf{0.0113 \pm 0.0011}$ \\
K=3 confidence                    & $0.3323 \pm 0.0294$ & $0.1357 \pm 0.0077$ & $1.2915 \pm 0.4317$ & $0.0139 \pm 0.0030$ \\
K=3 random equal-frequency        & $0.3322 \pm 0.0297$ & $0.1362 \pm 0.0081$ & $1.6789 \pm 0.4347$ & $0.0142 \pm 0.0032$ \\
K=3 shuffled risk-score           & $0.3325 \pm 0.0292$ & $0.1363 \pm 0.0080$ & $1.5867 \pm 0.4836$ & $0.0143 \pm 0.0031$ \\
\midrule
K=3 oracle risk $1{-}p_y$ (BCE) & $0.3214 \pm 0.0231$ & $0.1346 \pm 0.0066$ & $1.6949 \pm 0.2768$ & $0.0083 \pm 0.0008$ \\
SMART+BCE & $0.3312 \pm 0.0290$ & $0.1350 \pm 0.0070$ & $0.9483 \pm 0.2865$ & $0.0123 \pm 0.0015$ \\
\bottomrule
\end{tabular}%
\caption{\textbf{Objective-conditional routing ablation (full table).}
Modern-recipe ViT-B/16 / CIFAR-100, mean $\pm$ std; $K{=}3$ fixed, only
the routing rule varies. The BCE block is over five seeds (matching the
main paper's routing-rule ablation figure, Fig.~3, which plots these BCE
routing rules); the NLL block is an archived three-seed diagnostic. \emph{NLL objective}: deployable
routing variants collapse to one ECE band. \emph{BCE objective}: risk
routing beats confidence and random (risk $<$ confidence $<$ shuffled
$\approx$ random on ECE). SMART+BCE uses the SMART logit-gap architecture
\citep{guo2026smart} retrained with top-label BCE. The oracle-risk row
routes on the label-informed true-label risk $1{-}p_y$ (not a binary
correct/incorrect partition); it uses labels and is diagnostic only, not
an upper bound.}
\label{tab:router_causality}
\end{table*}

\textbf{Under NLL.} All deployable routing variants land in a tight band
($\mathrm{ECE}_{15} \in [1.71, 1.82]$, NLL within $\pm 3 \times 10^{-3}$).
The routing signal has no measurable effect under NLL fitting; we report the
oracle diagnostic under the BCE objective only.

\textbf{Under BCE.} The same routing rules now separate on ECE:
risk ($0.96$) $<$ confidence ($1.29$) $<$ shuffled ($1.59$) $\approx$
random ($1.68$). The OOF risk router beats even the
label-informed true-label-risk router $1{-}p_y$ ($1.69$), which
--- though distinct from the binary-correctness oracle of the
$B/\bar\sigma^2$ table --- induces extreme per-group BCE optima that
generalise worse than the smoother OOF risk partition. Under BCE fitting the
routing signal separates the variants, and the OOF logistic risk signal is a
competitive default.

\textbf{Interpretation.} The "constraint-not-router" reading is
specific to the NLL fitting objective. The main text's SRTS-BCE result
shows that the same $K{=}3$ infrastructure becomes a strong adaptive
calibrator once paired with a top-label BCE objective and an OOF risk
router. Routing matters when the per-group temperature optimisation has
gradient room to exploit it; NLL fitting on small validation slices does
not.

\textbf{Objective-agnostic routing check: top-label Brier score.}
To verify that the routing benefit is not specific to the BCE objective,
we repeat the routing ablation with per-group temperatures fitted by
minimising the \emph{top-label Brier score}
$\mathcal{L}_{\mathrm{Brier\text{-}top}}(T;\mathcal{G})
  = |\mathcal{G}|^{-1}\!\sum_{i\in\mathcal{G}}(\hat p_i - b_i)^2$,
where $\hat p_i = \max_c \mathrm{softmax}(z_i/T)_c$ is the top-label
confidence and $b_i = \mathbf{1}[\hat{y}_i = y_i]$ is correctness.
This is a proper scoring rule distinct from BCE and NLL; the $K{=}1$
limit is scalar Brier TS (``TvA-Brier'').
Under top-label Brier fitting, risk routing again outperforms
confidence and random alternatives
(risk $\mathrm{ECE}_{15}$ = $1.05$, confidence = $1.17$,
random = $1.68$; 3-seed mean), replicating the BCE ordering.
TvA-Brier ($K{=}1$) attains $\mathrm{ECE}_{15} = 1.68$,
confirming that the routing benefit is not an artefact of the
top-label BCE loss function but rather a property of the OOF
risk signal's ability to partition the validation set into
groups with distinct calibration optima.
The objective-conditional pattern replicates under top-label BCE and
top-label Brier: genuine reliability routing outperforms random routing.
Under NLL the routing rules are largely indistinguishable, while the
SoftECE diagnostic (below) shows that risk grouping can improve over its
scalar anchor but does not provide a full routing-rule comparison.
Full per-seed statistics are in Table~\ref{tab:brier_routing}.

% Objective-agnostic routing check: top-label Brier score objective.
% Three-seed K=3 group-wise temperature calibration, same base as router causality table.
% Key finding: risk routing beats confidence/random under Brier, replicating BCE ordering.

\begin{table*}[t]
\centering
\small
\setlength{\tabcolsep}{3.5pt}
\begin{tabular}{lrrrr}
\toprule
Routing rule & NLL & Brier & $\mathrm{ECE}_{15}$ & eAURC \\
\midrule
\multicolumn{5}{l}{\textit{Brier objective}} \\
TvA-Brier $\equiv$ K=1 Brier       & $0.3276 \pm 0.0228$ & $0.1332 \pm 0.0058$ & $1.6830 \pm 0.6586$ & $0.0137 \pm 0.0030$ \\
\textit{SRTS-Brier K=3 risk}       & $0.3233 \pm 0.0211$ & $0.1326 \pm 0.0054$ & $\mathbf{1.0518 \pm 0.3480}$ & $\mathbf{0.0114 \pm 0.0014}$ \\
K=3 confidence                     & $0.3270 \pm 0.0234$ & $0.1331 \pm 0.0057$ & $1.1723 \pm 0.2711$ & $0.0135 \pm 0.0027$ \\
K=3 random equal-frequency         & $0.3282 \pm 0.0229$ & $0.1333 \pm 0.0059$ & $1.6811 \pm 0.5510$ & $0.0139 \pm 0.0031$ \\
\bottomrule
\end{tabular}%
\caption{\textbf{Objective-agnostic routing check: $K{=}3$ group-wise temperature calibration under the top-label Brier objective.} Same modern-recipe ViT-B/16 / CIFAR-100 base as Table~\ref{tab:router_causality}, mean $\pm$ std over seeds 0/1/2. Top-label Brier ($\mathcal{L} = \overline{(q_i - b_i)^2}$) is a proper scoring rule distinct from both BCE and NLL. OOF risk routing again outperforms confidence and random routing by $0.12$--$0.63$ pp ECE, replicating the ordering from the BCE objective and confirming that the routing benefit is a property of the OOF risk signal rather than an artefact of the BCE loss function. TvA-Brier ($K{=}1$) attains ECE $\approx$ random ($1.68$ vs.\ $1.68$), consistent with random routing degenerating to near-scalar behaviour.}
\label{tab:brier_routing}
\end{table*}

\textbf{Objective still matters: per-group SoftECE.} We also fit each
$K{=}3$ group's temperature by minimising the soft-binned ECE (SoftECE)
objective of SMART, holding the router and groups fixed. Routing again
learns three distinct group temperatures (seed-0 span $0.60$--$0.83$) and
beats scalar TvA-TS, so the routing benefit is not specific to BCE. But
SoftECE is a weaker \emph{per-group} objective than top-label BCE at this
budget: clean CIFAR-100 / ViT-B/16 $\mathrm{ECE}_{15}$ is
$0.95 \pm 0.28$ for SRTS-SoftECE versus $0.83 \pm 0.28$ for SRTS-BCE
refit under the identical pipeline (both three-seed; the five-seed
value is $0.96$), and SoftECE is also higher-variance across seeds
(worst seed $1.04$ vs.\ $0.67$). The soft-binned objective on
${\approx}833$-sample groups is noisier to fit than the smooth top-label
log-loss, which is why the deployed method fits groups by BCE.

\subsection{Mixed-objective diagnostic}
\label{app:mixed-objective}

The main method uses pure top-label BCE; mixed NLL-plus-BCE fits are
reported only as a diagnostic connecting the two objectives. On the same
three-seed ViT-B/16 / CIFAR-100 setting, increasing the BCE weight moves the
$K{=}3$ risk-routed variant toward the pure-BCE result while preserving the
same OOF risk router:

\begin{table}[t]
\centering
\small
\setlength{\tabcolsep}{5pt}
\begin{tabular}{lrr}
\toprule
Method & NLL & $\mathrm{ECE}_{15}$ \\
\midrule
SRTS-Mixed $K{=}3$ risk, $\lambda=0.1$ & 0.3217 & 1.6617 \\
SRTS-Mixed $K{=}3$ risk, $\lambda=0.3$ & 0.3218 & 1.5023 \\
SRTS-Mixed $K{=}3$ risk, $\lambda=1$   & 0.3222 & 1.1884 \\
SRTS-Mixed $K{=}3$ risk, $\lambda=3$   & 0.3231 & 0.9698 \\
SRTS-Mixed $K{=}3$ risk, $\lambda=10$  & 0.3241 & 0.8667 \\
\bottomrule
\end{tabular}
\caption{\textbf{Mixed-objective SRTS diagnostic} (three-seed mean;
$\mathcal{L}=\mathcal{L}_{\mathrm{NLL}}+\lambda
\mathcal{L}_{\mathrm{BCE}}$). These are ablations, not deployed-method
rows; as $\lambda$ grows the mixed fit approaches the
pure-BCE configuration (three-seed value $0.83{\pm}0.28$; the five-seed value is $0.96{\pm}0.32$).}
\label{tab:mixed-objective-diagnostic}
\end{table}

\subsection{Adaptive-head fairness sweep (capacity-control diagnostic)}
\label{app:fairness-sweep}

% Item 3 — Adaptive-head fairness sweep (Stage-1 supplementary).
% Best-val-selected hyperparameter cells from a 48-cell grid:
%   head_width   in {linear, 16, 32, 128}
%   weight_decay in {0, 1e-4, 1e-3, 1e-2}
%   learning rate in {1e-3, 1e-4, 1e-5}
% Early stopping on validation NLL with patience 20.
% Selection rule: lowest validation NLL per (method, seed). No test-set tuning.

\begin{table*}[t]
\centering
\small
\setlength{\tabcolsep}{3.5pt}
\begin{tabular}{lrrr}
\toprule
Method (best-val cell)                & val NLL & test $\mathrm{ECE}_{15}$ & test NLL \\
\midrule
\textit{Reference: scalar TS-NLL}      & ---     & $1.7950 \pm 0.6977$ & $0.3239 \pm 0.0297$ \\
\textit{Reference: SRTS-NLL $K{=}3$ (ablation)} & --- & $1.7058 \pm 0.5294$ & $0.3217 \pm 0.0274$ \\
\textit{Reference: SRTS-BCE $K{=}3$ risk (main)} & --- & $0.8276 \pm 0.2791$ & $0.3246 \pm 0.0206$ \\
\textit{Reference: SMART+BCE}          & ---     & $0.8573 \pm 0.2871$ & $0.3245 \pm 0.0217$ \\
\midrule
SATS-Logit  (best-val)                 & $0.2151 \pm 0.0074$ & $5.7416 \pm 0.4385$ & $0.6917 \pm 0.0726$ \\
SATS-Features (best-val)               & $0.2099 \pm 0.0042$ & $6.4659 \pm 0.4906$ & $0.8418 \pm 0.2136$ \\
LTS-feature (best-val)                 & $0.2099 \pm 0.0042$ & $6.4659 \pm 0.4906$ & $0.8418 \pm 0.2136$ \\
\bottomrule
\end{tabular}%
\caption{\textbf{Adaptive-head fairness sweep as a capacity-control diagnostic on modern-recipe ViT-B/16 / CIFAR-100, three seeds.} For each unrestricted adaptive-head family, a $48$-cell grid (head width $\in \{$linear$, 16, 32, 128\} \times$ weight decay $\in \{0, 10^{-4}, 10^{-3}, 10^{-2}\} \times$ learning rate $\in \{10^{-3}, 10^{-4}, 10^{-5}\}$) is fit with early stopping on val NLL (patience 20). The cell with the lowest \emph{validation} NLL is selected per (method, seed) --- no test-set tuning. \textbf{Across all nine (method, seed) combinations the val-only selection picks the largest head} (width $128$, learning rate $10^{-3}$). The constrained and SMART+BCE rows are references from the objective-aligned comparison, not cells in this NLL sweep. \emph{Notes.} All $9 / 9$ best-val cells select head width $128$ and learning rate $10^{-3}$. Weight decay variation is essentially ignored by the val-NLL criterion because lower-$\ell_2$ cells reach lower val NLL by over-fitting. SATS-Features and LTS-feature share the same feature-input MLP topology in our implementation and select the same cell, so their best-val rows are identical.}
\label{tab:fairness_sweep}
\end{table*}

Table~\ref{tab:fairness_sweep} reports a 48-cell
hyperparameter sweep on SATS-Logit, SATS-Features, and LTS-feature,
alongside constrained SRTS-NLL, SRTS-BCE, and SMART+BCE references.
The sweep tests the capacity-control motivation: the neural adaptive
heads have far more degrees of freedom than the $K{=}3$ routed family.
Across all nine (method, seed) combinations the val-only selection
picks the width-$128$ head with learning rate $10^{-3}$ --- the largest,
least-regularised cell --- consistent with these heads overfitting the
val-NLL selection objective at this budget. Best-val-selected test
$\mathrm{ECE}_{15}$ remains $\geq 3\times$ scalar TS on every
method-seed.

\subsection{Cross-backbone adaptive-head failure summary}
\label{app:adaptive-head-cross-backbone-failure}

% Supplementary: unrestricted adaptive-head ECE15 failure summary across
% backbones and datasets (NLL-trained heads).
% Source: phaseb_deit_swin_full.tex (C100 DeiT/Swin),
%         sota_3seed_summary (C100 ViT, SATS-Logit only; other heads
%         from the single-seed Table 2 / fairness sweep),
%         tinyimagenet_3seed_summary (TinyIN ViT),
%         tinyimagenet_deit_swin_full.tex (TinyIN DeiT/Swin).
% All values: mean +/- std, 3 seeds, ddof=0 (except where noted).
% Requires: \usepackage{booktabs}
%
\begin{table*}[t]
\centering
\small
\setlength{\tabcolsep}{8pt}
\begin{tabular}{l r r r}
\toprule
Regime / backbone & SATS-Logit & LTS-feat.\ / Feat-SATS / UTempH & Range \\
\midrule
C100 / ViT-B/16 & $6.17 {\pm} 0.31$ & $6.71 {\pm} 0.44$  & $6.2$--$6.7$ \\
C100 / DeiT-S   & $7.02 {\pm} 1.05$ & $8.31 {\pm} 0.65$  & $7.0$--$8.3$ \\
C100 / Swin-T   & $7.24 {\pm} 0.49$ & $8.46 {\pm} 0.44$  & $7.2$--$8.5$ \\
TinyIN / ViT-B/16 & $7.40 {\pm} 0.35$ & $7.05 {\pm} 0.46$ & $7.1$--$7.4$ \\
TinyIN / DeiT-S  & $9.75 {\pm} 0.16$ & $10.41 {\pm} 0.07$ & $9.8$--$10.4$ \\
TinyIN / Swin-T  & $10.06 {\pm} 0.30$ & $11.02 {\pm} 0.15$ & $10.1$--$11.0$ \\
\bottomrule
\end{tabular}
\caption{\textbf{Unrestricted adaptive-head $\mathrm{ECE}_{15}$ failure summary} (\%, mean $\pm$ std over three seeds). All four methods are NLL-trained MLP temperature heads with varying input representations (logits or CLS features). Every unrestricted head produces $\mathrm{ECE}_{15} \geq 6\%$ across all six backbone$\times$dataset settings, far exceeding both scalar TS ($1$--$3\%$) and the constrained adaptive calibrators SRTS-BCE and SMART+BCE (approximately $1.25\%$ or below), motivating the deliberate capacity constraint in SRTS-BCE ($K + (D{+}1) = 10$ parameters) and SMART+BCE ($49$ parameters). Full per-method tables are in \S\ref{app:backbone-generalisation}.}
\label{tab:adaptive_head_failure_cross_backbone}
\end{table*}

Table~\ref{tab:adaptive_head_failure_cross_backbone} collects the
unrestricted adaptive-head $\mathrm{ECE}_{15}$ across all six
backbone$\times$dataset settings in the paper. The failure mode is
consistent: every unrestricted head remains at $\mathrm{ECE}_{15}
\geq 6\%$, far above the ${\leq}1.2\%$ achieved by SRTS-BCE $K{=}3$ and
SMART+BCE on the same splits. This table provides the per-method detail
behind the main-text statement that unrestricted adaptive heads overfit
under the small-validation protocol. Among the reported heads,
LTS-feature, Feature-SATS, and Frozen-feature UTempH produce nearly
identical $\mathrm{ECE}_{15}$ within each cell (typically ${\leq}0.02$ pp
apart), so the column reports their common value; SATS-Logit often attains
a lower $\mathrm{ECE}_{15}$ than the feature-based heads because it uses
logits (100- or 200-dim) instead of CLS features (768-dim), giving fewer
overfit-prone parameters, but it still regresses well beyond scalar TS.

\subsection{BCE-trained unrestricted adaptive-head fairness check}
\label{app:sats-bce}

% BCE-trained unrestricted adaptive-head fairness check.
% SATS-Logit-BCE and Feature-SATS-BCE trained and selected under top-label BCE
% (36-config grid, Cal-Fit/Cal-Select 80/20, refit on full val).
% Three-seed ViT-B/16 / CIFAR-100 Protocol B, mean +/- std (ddof=0).
\begin{table*}[t]
\centering
\small
\setlength{\tabcolsep}{4pt}
\begin{tabular}{lrrrr}
\toprule
Method & Top-1 & ECE$_{15}$ & NLL & eAURC \\
\midrule
\multicolumn{5}{l}{\textit{Constrained scalar / grouped baselines}} \\
TS-NLL              & $91.17$ & $1.80{\pm}0.57$ & $0.324{\pm}0.024$ & $0.01324{\pm}0.0028$ \\
TvA-TS ($K{=}1$ BCE) & $91.17$ & $1.54{\pm}0.42$ & $0.325{\pm}0.023$ & $0.01349{\pm}0.0029$ \\
SRTS-BCE $K{=}3$    & $91.17$ & $0.83{\pm}0.28$ & $0.325{\pm}0.021$ & $0.01147{\pm}0.0013$ \\
SMART+BCE           & $91.17$ & $0.86{\pm}0.29$ & $0.325{\pm}0.022$ & $0.01187{\pm}0.0014$ \\
\midrule
\multicolumn{5}{l}{\textit{BCE-trained unrestricted adaptive heads}} \\
SATS-Logit-BCE      & $91.17$ & $8.12{\pm}0.84$ & $0.939{\pm}0.049$ & $0.03197{\pm}0.0067$ \\
Feature-SATS-BCE    & $91.17$ & $9.61{\pm}0.44$ & $1.007{\pm}0.044$ & $0.02861{\pm}0.0082$ \\
\bottomrule
\end{tabular}
\\[2pt]
{\small\raggedright\noindent Per-seed ECE$_{15}$: SATS-Logit-BCE $\{7.88, 9.25, 7.23\}$;
Feature-SATS-BCE $\{9.74, 10.07, 9.01\}$.
Best selected configs: SATS-Logit-BCE hidden=16--128, lr=$10^{-3}$, wd=$10^{-2}$;
Feature-SATS-BCE hidden=16--128, lr=$10^{-3}$, wd=$10^{-2}$.
Top-1 is unchanged because temperature scaling is argmax-preserving.\par}
\caption{\textbf{BCE-trained unrestricted adaptive-head fairness check.}
SATS-Logit-BCE and Feature-SATS-BCE are trained and selected under the
same top-label BCE objective as SRTS-BCE and SMART+BCE, using an 80/20
Cal-Fit/Cal-Select split and a 36-config hyperparameter sweep
(hidden $\in\{16,32,128\}$, lr $\in\{10^{-3},10^{-4},10^{-5}\}$,
weight-decay $\in\{0,10^{-4},10^{-3},10^{-2}\}$);
the best Cal-Select configuration is refit on the full calibration split.
Both MLP variants share the same input$\to$hidden$\to$GELU$\to$1$\to$softplus
architecture; they differ only in input dimension (100 logits vs.\ 768 features).
ECE$_{15} \geq 8$ confirms the failure mode is not specific to the NLL objective.}
\label{tab:sats_bce_fairness}
\end{table*}

Table~\ref{tab:sats_bce_fairness} reports the BCE-trained adaptive-head
fairness check. SATS-Logit-BCE and Feature-SATS-BCE are trained and
selected under the same top-label BCE objective as SRTS-BCE and SMART+BCE,
using an 80/20 Cal-Fit/Cal-Select stratified split of the ${\approx}2{,}500$
calibration samples and a 36-config hyperparameter sweep
(hidden $\in\{16,32,128\}$, lr $\in\{10^{-3},10^{-4},10^{-5}\}$,
weight-decay $\in\{0,10^{-4},10^{-3},10^{-2}\}$). The best
Cal-Select configuration is refit on the full calibration split before
a single test-set evaluation.

Both MLP architectures share the same input$\to$hidden$\to$GELU$\to$scalar
topology and differ only in input dimension (100 logits vs.\ 768 CLS features).
SATS-Logit-BCE attains $\mathrm{ECE}_{15} = 8.12 \pm 0.84$ and
Feature-SATS-BCE attains $9.61 \pm 0.44$ across three seeds, compared to
the three-seed TvA-TS ($1.54$) and SRTS-BCE ($0.83$).
Both methods also substantially degrade NLL ($0.94$ and $1.01$ vs.\ $0.32$
for TvA-TS). The poor calibration is therefore not specific to the NLL
training criterion: these heads remain poorly calibrated under the evaluated
nested BCE-selection protocol as well, consistent with a capacity/overfit
failure given ${\geq}1{,}600$--$98{,}000$ parameters fitted on a
${\approx}2{,}000$-sample Cal-Fit split. SRTS-BCE uses $K{+}(D{+}1) = 10$
scalar parameters on the same split and stays out of this regime. We do not
claim every regularised or ensembled head must fail; we report that the
heads and protocols we evaluated did.

\subsection{NLL-objective diagnostics}
\label{app:nll-history}

This subsection reports the SRTS-NLL variant as ablation evidence for the
objective-conditional story (under NLL the routing signal collapses; under
BCE it separates the variants). None of it is the deployed method.

\subsubsection{Controlled single-checkpoint diagnostic}
\label{app:protocol-a-controlled-diagnostic}

% Edit the generator, not this file; rerun the script.
%
% Requires: \usepackage{booktabs}
%
\begin{table*}[t]
\centering
\small
\setlength{\tabcolsep}{2pt}
\begin{tabular}{lrrrrrrrr}
\toprule
Method & Top-1 & NLL & Brier & ECE\textsubscript{15} & AdaECE\textsubscript{15} & AUROC & AURC & eAURC \\
\midrule
CE / NoTS & 53.71 & 1.7179 & 0.5975 & 3.1788 & 3.2246 & 0.8089 & 0.2246 & 0.0956 \\
CE + TS-NLL & 53.71 & 1.7065 & 0.5963 & 2.3578 & 2.3451 & \underline{0.8101} & \underline{0.2241} & \underline{0.0950} \\
CE + LTS\textsubscript{scalar} & 53.71 & \underline{1.7063} & 0.5962 & \underline{2.1861} & \underline{2.2012} & 0.8100 & 0.2241 & 0.0950 \\
CE + SATS\textsubscript{logits} & 53.71 & 2.2788 & 0.6483 & 12.4806 & 12.5604 & 0.7481 & 0.2689 & 0.1399 \\
CE + LTS\textsubscript{feature} & 53.71 & 2.5166 & 0.6584 & 13.5758 & 13.6473 & 0.7401 & 0.2746 & 0.1455 \\
CE + Feature-SATS\tablefootnote{\textsc{Feature-SATS} uses the same feature-input SATS fitting path as the frozen-feature variant in our implementation and therefore reduces to one row.} & 53.71 & 2.5185 & 0.6585 & 13.5917 & 13.6543 & 0.7400 & 0.2746 & 0.1455 \\
\textbf{CE + SRTS-NLL (logits; ablation)} & 53.71 & \textbf{1.7050} & \textbf{0.5959} & \textbf{2.1434} & \textbf{2.1631} & 0.8100 & 0.2241 & 0.0950 \\
CE + SRTS-NLL (features; ablation)\tablefootnote{SRTS-NLL feature routing is reported as an ablation because feature routing can overfit at the $\sim$2{,}500-sample validation scale; see the Limitations section.} & 53.71 & 1.7099 & \underline{0.5960} & 2.6193 & 2.5868 & \textbf{0.8114} & \textbf{0.2240} & \textbf{0.0949} \\
\bottomrule
\end{tabular}
\caption{\textbf{Controlled single-checkpoint diagnostic under NLL fitting} (CIFAR-100, ViT-B/16). Exposes the validation-overfit failure of unrestricted sample-adaptive heads under matched checkpoint and protocol; failure-mode evidence, \emph{not} the five-seed headline performance estimate reported in the main paper's ID-clean summary table. SRTS-NLL logits/features are NLL-objective ablations. All rows are post-hoc calibrators applied to the same CE base checkpoint and therefore preserve top-1 accuracy. \textbf{Bold}: best per column; \underline{underline}: second-best.}
\label{tab:table1_protocol_a}
\end{table*}

Table~\ref{tab:table1_protocol_a} is a mechanism diagnostic, not
a main performance estimate. It uses a matched CIFAR-100 / ViT-B/16
checkpoint to isolate the validation-overfit failure mode of unrestricted
adaptive temperature heads under NLL fitting: the mild initial
miscalibration ($\mathrm{ECE}_{15} = 3.18$) is amplified to $\geq 12$ by
unrestricted heads, while scalar and constrained routed calibrators remain
stable. The checkpoint uses a separate single-seed recipe ($350$ epochs
SGD, crop/flip augmentation only; top-1 $53.71$); the main claims use the
five-seed ID clean and three-seed deployable shift-evaluation results. The
bootstrap-aggregated post-recovery $\mathrm{ECE}_{15}$ is
$1.71$ $[1.12,\,2.38]$ ($B{=}1000$;
Table~\ref{tab:stage4_srts_bootstrap}).

\begin{figure}[t]
\centering
\includegraphics[width=0.95\columnwidth]{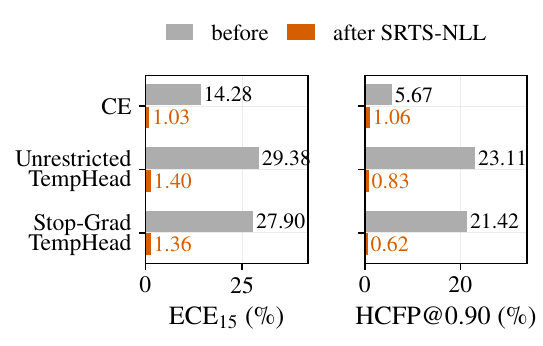}
\caption{\textbf{Recovery case study: NLL-objective SRTS applied to a
severely miscalibrated Unrestricted TempHead.} Diagnostic figure
generated with the SRTS-NLL ablation variant. Single-run point estimates:
pre-$\mathrm{ECE}_{15} = 29.38 \to$ post $1.40$; pre-NLL $4.253 \to$
post $1.281$; pre-HCFP@$0.90$ $23.11 \to$ post $0.83$
(\S\ref{app:metric-definitions}). Recovery uses only logit
signals --- no further training, no feature signals, and no access to
the miscalibrated head's learned $T(x)$. This plot documents an
NLL-objective diagnostic only; it is not evidence of a BCE-objective
recovery run.}
\label{fig:srts-recovery-case-study}
\end{figure}

% Archived stage4 asset (no data file / generator in this repo); shown at
% full text width because the natural canvas is ~11.5 in and a
% single-column inclusion renders its fonts unreadably small.
\begin{figure*}[t]
\centering
\includegraphics[width=\textwidth]{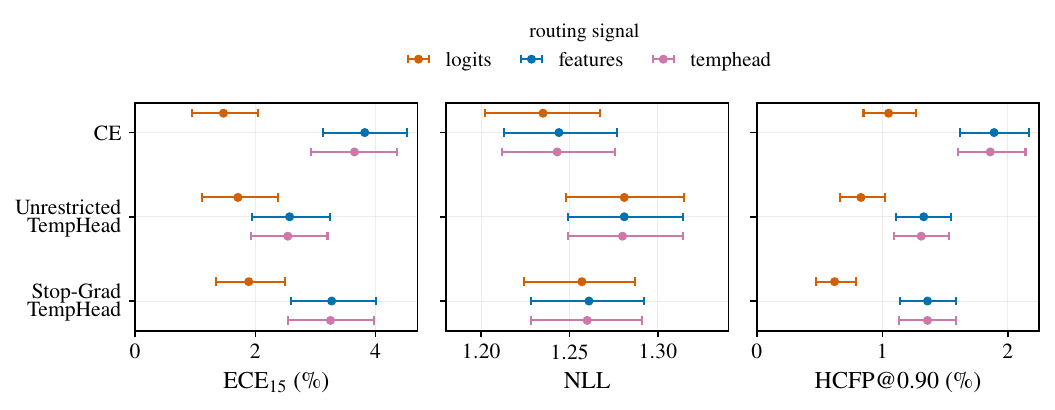}
\caption{\textbf{Signal-routing ablation (NLL-objective diagnostic).}
Logits-only routing is the reliable default; feature-input routing achieves
higher val risk-AUROC but regresses on test $\mathrm{ECE}_{15}$
(controlled setting: $2.62$ vs.\ $2.14$), consistent with small-val
feature-routing overfit. Routing on a Stop-Grad TempHead's learned $T(x)$
does not improve over scalar TS; the signal collapses to near-global
scaling. The signal-input choice (logits vs.\ features) applies equally
to SRTS-NLL and SRTS-BCE; the main text's SRTS-BCE uses the same
logits-only routing because of this diagnostic.
\emph{x-axis bases:} CE+Brier (CE+Brier baseline),
UTempH (Unrestricted TempHead), SG-TempH (Stop-Grad TempHead).}
\label{fig:srts-signal-ablation}
\end{figure*}

% K-group ablation moved to main text (formerly appendix Table 4 / Stage-4
% bundle). Numerical content unchanged; layout slightly compressed to fit
% the main-text page budget (\small, narrower tabcolsep).
\begin{table*}[t]
\centering
\small
\setlength{\tabcolsep}{3pt}
\begin{tabular}{lrrrrrr}
\toprule
Base & $K$ & ECE$_{15}$ & NLL & HCFP@0.9 & $T_g$ range & $T_{\max}/T_{\min}$ \\
\midrule
CE & 1 & 1.06 & 1.236 & 0.97 & [1.62, 1.62] & 1.00\,\(\times\) \\
CE & 2 & 1.02 & 1.235 & 1.05 & [1.60, 1.63] & 1.02\,\(\times\) \\
CE & 3 & 1.03 & 1.235 & 1.06 & [1.58, 1.67] & 1.06\,\(\times\) \\
CE & 5 & 1.25 & 1.236 & 1.27 & [1.54, 1.71] & 1.11\,\(\times\) \\
CE & 10 & 0.89 & 1.237 & 1.22 & [1.46, 1.81] & 1.24\,\(\times\) \\
\addlinespace[2pt]
Unrestricted TempHead & 1 & 1.48 & 1.282 & 0.75 & [5.61, 5.61] & 1.00\,\(\times\) \\
Unrestricted TempHead & 2 & 0.78 & 1.281 & 0.90 & [5.45, 5.73] & 1.05\,\(\times\) \\
Unrestricted TempHead & 3 & 1.40 & 1.281 & 0.83 & [5.48, 5.67] & 1.03\,\(\times\) \\
Unrestricted TempHead & 5 & 1.00 & 1.282 & 0.95 & [5.41, 5.81] & 1.08\,\(\times\) \\
Unrestricted TempHead & 10 & 0.83 & 1.282 & 1.02 & [5.17, 6.02] & 1.16\,\(\times\) \\
\addlinespace[2pt]
Stop-Grad TempHead & 1 & 1.31 & 1.257 & 0.75 & [5.13, 5.13] & 1.00\,\(\times\) \\
Stop-Grad TempHead & 2 & 1.50 & 1.257 & 0.73 & [5.12, 5.15] & 1.00\,\(\times\) \\
Stop-Grad TempHead & 3 & 1.36 & 1.257 & 0.62 & [5.04, 5.23] & 1.04\,\(\times\) \\
Stop-Grad TempHead & 5 & 1.36 & 1.258 & 0.54 & [4.86, 5.38] & 1.11\,\(\times\) \\
Stop-Grad TempHead & 10 & 1.25 & 1.259 & 0.55 & [4.63, 5.50] & 1.19\,\(\times\) \\
\bottomrule
\end{tabular}%
\caption{\textbf{$K$-group ablation for SRTS-NLL (NLL-objective ablation)}
on three matched single-checkpoint bases (CE, Unrestricted TempHead, Stop-Grad TempHead).
$K=1$ reduces to global scalar TS-NLL by construction. Numbers are percent except the
group-temperature spread $T_g$ range / $T_{\max}/T_{\min}$. The per-base
$\mathrm{ECE}_{15}$ band across $K \in \{1, 2, 3, 5, 10\}$ is at most
$\approx 0.6$ percentage points, while the per-group temperature spread
$T_{\max}/T_{\min}$ grows monotonically with $K$; we read this as
evidence that larger $K$ expands the parameter budget without delivering
matching calibration improvement on this protocol.}
\label{tab:stage4_srts_k}
\end{table*}

\textbf{NLL-objective $K$-ablation.} The $K$-group ablation
(Table~\ref{tab:stage4_srts_k}) reports SRTS-NLL at
$K \in \{1, 2, 3, 5, 10\}$ on three matched single-checkpoint bases. The per-group
temperature spread $T_{\max}/T_{\min}$ grows monotonically with $K$, but
the calibration metrics stay within a tight band: per-base
$\mathrm{ECE}_{15}$ differences are at most $\approx 0.6$ pp. Under NLL
fitting, $K$ alone has no measurable effect.

\textbf{Feature and Stop-Grad TempHead signals are ablations, not main methods.}
The signal-routing ablation (Figure~\ref{fig:srts-signal-ablation}) shows that
logits-only routing is the reliable default under both NLL and BCE objectives;
feature routing achieves higher val risk-AUROC but can regress on test
$\mathrm{ECE}_{15}$ at the $\sim 2{,}500$-val regime; routing on a
stop-gradient Unrestricted TempHead's learned $T(x)$ fails to improve over
scalar TS.

\subsubsection{Bootstrap robustness}

% Stage-4 SRTS robustness appendix table — auto-generated.
\begin{table*}[t]
\centering
\small
\setlength{\tabcolsep}{4pt}
\begin{tabular}{llrrrr}
\toprule
Base & Variant & ECE$_{15}$ & NLL & HCFP@0.9 & AUROC$_{\text{corr}}$ \\
\midrule
CE & raw & 14.27\,[13.50,15.05] & 1.441\,[1.393,1.487] & 5.67\,[5.23,6.14] & 0.850\,[0.843,0.858] \\
CE & \textbf{SRTS-NLL (logits)} & 1.47\,[0.95,2.05] & 1.235\,[1.202,1.267] & 1.05\,[0.85,1.27] & 0.854\,[0.846,0.861] \\
CE & SRTS-NLL (features) & 3.82\,[3.12,4.52] & 1.244\,[1.213,1.277] & 1.89\,[1.62,2.17] & 0.853\,[0.845,0.861] \\
CE & SRTS-NLL (temphead) & 3.65\,[2.93,4.35] & 1.243\,[1.212,1.276] & 1.86\,[1.60,2.14] & 0.854\,[0.846,0.861] \\
\addlinespace[3pt]
Unrestricted TempHead & raw & 29.39\,[28.48,30.29] & 4.250\,[4.095,4.413] & 23.11\,[22.31,23.97] & 0.825\,[0.817,0.832] \\
Unrestricted TempHead & \textbf{SRTS-NLL (logits)} & 1.71\,[1.12,2.38] & 1.281\,[1.248,1.315] & 0.83\,[0.66,1.02] & 0.845\,[0.837,0.852] \\
Unrestricted TempHead & SRTS-NLL (features) & 2.57\,[1.95,3.24] & 1.281\,[1.249,1.314] & 1.33\,[1.11,1.55] & 0.851\,[0.844,0.858] \\
Unrestricted TempHead & SRTS-NLL (temphead) & 2.54\,[1.93,3.20] & 1.280\,[1.249,1.314] & 1.31\,[1.09,1.53] & 0.851\,[0.844,0.858] \\
\addlinespace[3pt]
Stop-Grad TempHead & raw & 27.88\,[27.10,28.80] & 3.760\,[3.615,3.901] & 21.40\,[20.63,22.21] & 0.823\,[0.815,0.831] \\
Stop-Grad TempHead & \textbf{SRTS-NLL (logits)} & 1.89\,[1.34,2.49] & 1.257\,[1.224,1.287] & 0.62\,[0.47,0.79] & 0.842\,[0.835,0.850] \\
Stop-Grad TempHead & SRTS-NLL (features) & 3.27\,[2.60,4.01] & 1.261\,[1.228,1.292] & 1.36\,[1.14,1.59] & 0.846\,[0.839,0.854] \\
Stop-Grad TempHead & SRTS-NLL (temphead) & 3.25\,[2.55,3.98] & 1.260\,[1.228,1.291] & 1.36\,[1.13,1.59] & 0.846\,[0.839,0.854] \\
\addlinespace[3pt]
\bottomrule
\end{tabular}
\caption{SRTS-NLL single-checkpoint bootstrap 95\% CIs on CIFAR-100 test (NLL-objective ablation; $B{=}1000$ resamples; the fitted routing is held fixed). Numbers are percent.}
\label{tab:stage4_srts_bootstrap}
\end{table*}

% K-group ablation table moved to the main text (see
% tables/main_srts_k_ablation.tex; label tab:stage4_srts_k
% retained). The appendix narrative that previously discussed K=1..10
% has been moved with the table.

Table~\ref{tab:stage4_srts_bootstrap} reports $B = 1000$ bootstrap $95\%$
confidence intervals on test-set calibration and selective-classification metrics
across three backbone bases (CE, Unrestricted TempHead, Stop-Grad TempHead) and
four post-hoc rows for the NLL-objective SRTS variant. SRTS-NLL is the only
post-hoc variant that moves every base into the $\mathrm{ECE}_{15} \le 2$
regime; the confidence intervals exclude the pre-recovery regime by a wide
margin. These recovery CIs are an SRTS-NLL ablation; the
BCE-objective comparison is separately measured on the modern-recipe
benchmark in the main text's five-seed ID clean summary table.

\subsubsection{ID clean result visualisation}

\begin{figure*}[t]
\centering
\begin{subfigure}[t]{0.48\textwidth}
\centering
\includegraphics[width=\textwidth]{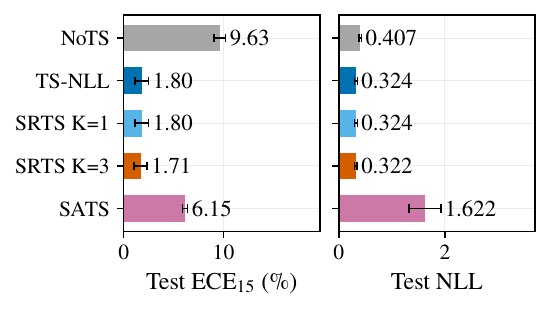}
\caption{Global calibration metrics ($\mathrm{ECE}_{15}$, NLL).}
\label{fig:srts-protocol-b-bars-a}
\end{subfigure}\hfill
\begin{subfigure}[t]{0.48\textwidth}
\centering
\includegraphics[width=\textwidth]{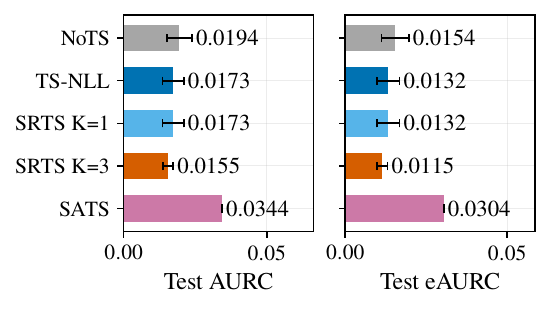}
\caption{Risk-coverage metrics (AURC, eAURC).}
\label{fig:srts-protocol-b-bars-b}
\end{subfigure}
\caption{\textbf{ID clean results over three modern-recipe ViT seeds
(NLL-objective rows).} Bars show mean $\pm$ standard deviation over seeds
0/1/2 (top-1 unchanged within each reported post-hoc seed; 3-seed mean
$91.17 \pm 0.46$); the SATS-Logit AURC/eAURC bars show the 3-seed mean
without whiskers (per-seed spread not archived). Under the NLL objective, SRTS-NLL $K{=}1$ recovers TS-NLL
and $K{=}3$ adds only modest gain ($-0.09$ pp). Under the BCE objective
(see the main text's five-seed ID clean summary table and
Table~\ref{tab:router_causality}; this NLL-objective figure is
an archived three-seed diagnostic), SRTS-BCE $K{=}3$ risk attains
$\mathrm{ECE}_{15} = 0.96$ (five seeds), beating TvA-TS ($K{=}1$ BCE) by $0.70$ pp.
SATS-Logit remains substantially worse across seeds, illustrating the
validation-overfit failure mode of unrestricted adaptive temperature
heads.}
\label{fig:srts-protocol-b-bars}
\end{figure*}

%==========================================================================
% C. Matched low-capacity maps
%==========================================================================

\section{Matched Low-Capacity Maps: Grouped versus Continuous Risk-Conditioned Scaling}
\label{app:matched-capacity}

This appendix reports the protocol-frozen three-seed
matched-parameterisation run behind the main RQ2 (design and
implementation fixed before the run) together with a protocol-frozen
two-seed extension for the flagship clean CIFAR-100 / ViT-B/16 cell. The
original run reproduced its reference values --- the archived three-seed
SRTS-BCE value of $0.8276$ clean and
$4.6283$ on CIFAR-100-C --- and all 27 continuous fits converged;
Table~\ref{tab:matched_capacity_full} reports the combined five-seed
clean results. No method definition, hyperparameter, knot placement,
fitting rule, or analysis criterion was changed for the two-seed
extension. Sign
convention throughout: $\Delta = $ SRTS-BCE $-$ comparator, so negative
values favour SRTS-BCE. Every temperature map shares the calibration
split, the top-label BCE objective, the OOF fitting discipline
(coefficients fit on out-of-fold risk scores; deployment uses the
full-split-refit router), and the initialisation at the global TvA-TS
temperature.

% from results/matched_capacity/20260724_150636 (six-signal regen).
% Regenerate with --write; verify with --check. Do not hand-edit numbers.
\begin{table*}[t]
\centering
\small
\setlength{\tabcolsep}{4pt}
\begin{tabular}{lrrrr}
\toprule
Method & C100/ViT & Tiny/ViT & IN-100/DeiT & C100-C/ViT \\
\midrule
TvA-TS ($K{=}1$) & $1.65{\pm}0.44$ & $1.26{\pm}0.18$ & $1.75{\pm}0.15$ & $5.72{\pm}1.13$ \\
QaTS-NLL & $1.90{\pm}0.61$ & $1.33{\pm}0.24$ & $2.56{\pm}0.20$ & $6.51{\pm}0.57$ \\
QaTS-BCE & $1.35{\pm}0.31$ & $1.11{\pm}0.06$ & $2.03{\pm}0.20$ & $5.18{\pm}0.68$ \\
Margin-K3+BCE & $0.97{\pm}0.32$ & $0.99{\pm}0.16$ & $1.83{\pm}0.30$ & $4.70{\pm}0.78$ \\
LinearRiskTemp+BCE & $1.29{\pm}0.48$ & $1.35{\pm}0.08$ & $1.74{\pm}0.26$ & $4.89{\pm}0.76$ \\
PWLinear-3+BCE & $1.01{\pm}0.32$ & $1.26{\pm}0.11$ & $1.86{\pm}0.26$ & $4.58{\pm}0.86$ \\
SplineRiskTemp+BCE & $1.18{\pm}0.47$ & $1.27{\pm}0.10$ & $1.74{\pm}0.26$ & $4.89{\pm}0.77$ \\
SMART+BCE & $0.95{\pm}0.29$ & $0.97{\pm}0.22$ & $1.71{\pm}0.26$ & $4.75{\pm}0.79$ \\
\textbf{SRTS-BCE (Ours)} & $0.96{\pm}0.32$ & $1.11{\pm}0.15$ & $1.84{\pm}0.22$ & $4.63{\pm}0.84$ \\
\bottomrule
\end{tabular}
\caption{\textbf{Matched-capacity comparison, full method set} ($\mathrm{ECE}_{15}$, mean${\pm}$std; C100/ViT five seeds, others three; C100-C as in the main RQ2 table). LinearRiskTemp and SplineRiskTemp use the same router and fitting discipline; the spline's natural cubic basis with the three frozen knots has three coefficients and converges to a near-linear solution on every setting.}
\label{tab:matched_capacity_full}
\end{table*}

% from results/matched_capacity/20260724_150636/bootstrap_cis.json.
\begin{table*}[t]
\centering
\small
\setlength{\tabcolsep}{4pt}
\begin{tabular}{lrrrr}
\toprule
vs.\ comparator & point & cell & corr.-type & hierarchical \\
\midrule
TvA-TS ($K{=}1$) & $-1.09$ & $[-1.28,-0.88]$ & $[-1.64,-0.49]$ & $[-2.02,-0.36]$ \\
QaTS-NLL & $-1.88$ & $[-2.11,-1.64]$ & $[-2.56,-1.15]$ & $[-2.90,-0.88]$ \\
QaTS-BCE & $-0.55$ & $[-0.65,-0.44]$ & $[-0.83,-0.25]$ & $[-1.04,-0.13]$ \\
Margin-K3+BCE & $-0.07$ & $[-0.09,-0.04]$ & $[-0.11,-0.03]$ & $[-0.21,+0.02]$ \\
LinearRiskTemp+BCE & $-0.26$ & $[-0.32,-0.20]$ & $[-0.36,-0.15]$ & $[-0.43,-0.06]$ \\
PWLinear-3+BCE & $+0.05$ & $[+0.03,+0.08]$ & $[+0.01,+0.10]$ & $[-0.01,+0.14]$ \\
SplineRiskTemp+BCE & $-0.27$ & $[-0.31,-0.22]$ & $[-0.38,-0.14]$ & $[-0.43,-0.09]$ \\
SMART+BCE & $-0.12$ & $[-0.17,-0.07]$ & $[-0.25,+0.00]$ & $[-0.33,+0.03]$ \\
\bottomrule
\end{tabular}
\caption{\textbf{CIFAR-100-C three-tier bootstraps for $\Delta = $ SRTS-BCE $-$ comparator} ($\mathrm{ECE}_{15}$; negative favours SRTS-BCE; $B{=}20{,}000$). Tiers: paired cell-level; corruption-type cluster; full seed$\times$corruption$\times$severity hierarchical (primary).}
\label{tab:matched_capacity_tiers}
\end{table*}

\textbf{Methods.} LinearRiskTemp fits $T(x) = \mathrm{softplus}(a + b\,
q(x))$ on the OOF risk probability $q$; PWLinear-3 fits three positive
anchor temperatures at the OOF-$q$ $1/6$, $1/2$, $5/6$ quantiles with
linear interpolation and clamped ends --- the same fitted router and
temperature-map parameter count as SRTS-BCE (three anchors stored vs.\
two group thresholds); SplineRiskTemp uses a natural cubic basis on the
three frozen $25/50/75\%$ knots (three coefficients) and converges to a
near-linear solution on every setting; Margin-K3 replaces the learned
score with raw logit-margin tertiles.

\textbf{Uncertainty.} On the flagship C100/ViT clean cell the
seed$\times$sample hierarchical bootstrap over five seeds
($B{=}20{,}000$) gives SRTS-BCE $-$ TvA-TS $[-0.88,-0.36]$ (excludes
zero) while SRTS-BCE vs.\ PWLinear-3 $[-0.17,+0.09]$, vs.\ Margin-K3
$[-0.09,+0.08]$ and vs.\ SMART+BCE $[-0.19,+0.12]$ all include zero;
Tiny and IN-100 clean cells use three-seed paired bootstraps
($B{=}2{,}000$). On Tiny-ImageNet/ViT-B/16, SRTS-BCE improves over TvA-TS
and LinearRiskTemp. On the three-seed IN-100/DeiT-S matched-map diagnostic,
SRTS-BCE is statistically unresolved from the scalar and continuous
alternatives and has a slightly higher point estimate.
Table~\ref{tab:matched_capacity_tiers} reports all three CIFAR-100-C
tiers; the hierarchical tier is primary. For SRTS-BCE vs.\ PWLinear-3
the hierarchical CI includes zero while the cell-level and
corruption-type tiers exclude zero in PWLinear-3's favour --- the mirror
image, at the same evidential grade, of the SRTS-vs-SMART+BCE
directional comparison.

\textbf{Score-scale diagnostics.} Out-of-fold and full-refit validation
risk scores agree closely on every seed (mean shift $\leq 0.0004$;
standard-deviation ratio $\approx 0.99$), so the protocol-frozen
fold-router ensemble sensitivity was not triggered. Under shift the
test-score mass moves upward (mean $q$: $0.093$ clean, $0.125$ at
severity 1, $0.226$ at severity 5), which both the grouped and the
continuous maps track from the corrupted logits without refitting.
Fitted temperatures stay in $[0.60, 0.90]$ on the clean flagship with no
non-finite values and no excursions outside $[0.05, 20]$ in any
method or setting.

\textbf{QaTS comparison (protocol-frozen extension).} We additionally
evaluate Quantile-Adaptive Temperature Scaling
\citep{chakraborty2026qats} --- an equation-faithful reimplementation
(we did not locate official code) --- under the identical protocol: $q(x)$ is the
calibration-set empirical CDF of the max-softmax confidence, test
samples are mapped through the same CDF, and $T(x) = a\,(1-q(x)) + b$
with $a, b > 0$ is fit by full NLL (QaTS-NLL, as originally specified) or by
top-label BCE (QaTS-BCE, objective-matched), initialised at the
matching-objective scalar temperature with $a{=}0.01$; all 24 fits
converged. Results (tables above): QaTS-BCE improves
substantially over QaTS-NLL on every setting --- an external
replication of the objective axis. QaTS-BCE trails SRTS-BCE on
CIFAR-100, IN-100/DeiT-S, and CIFAR-100-C, and the two are
statistically unresolved on Tiny-ImageNet/ViT-B/16 ($1.11 \pm 0.06$
vs.\ $1.12 \pm 0.17$); against SRTS-BCE every CIFAR-100-C tier excludes
zero (hierarchical $[-1.05,-0.13]$ objective-matched;
$[-2.90,-0.89]$ as originally specified) and the five-seed clean C100 hierarchical CI excludes
zero as well ($[-0.51,-0.06]$). Our corruption evaluation uses the
broader 19-corruption CIFAR-100-C protocol (the original reports the
standard 15 corruptions); restricting to the standard 15 changes no
conclusion (QaTS-NLL $6.77 \pm 0.57$, QaTS-BCE $5.45 \pm 0.70$,
SRTS-BCE $4.91 \pm 0.87$). The fitted slope $a$ frequently lands
on its positivity bound (five of twelve clean fits), degenerating
toward a global scalar: at this budget the confidence quantile alone
separates less risk-relevant structure than the learned six-signal
score, consistent with the confidence-only routing ablation of \S\ref{app:router-causality}.
QaTS remains attractive for its rank-based shift stability and
two-parameter budget; the comparison isolates the conditioning signal,
not the quantile construction, as the differentiator.

% results/hts_baseline/{clean,budget_hier,c100c}.json. No hand-edits.
\begin{table*}[t]
\centering
\footnotesize
\setlength{\tabcolsep}{5pt}
\caption{\textbf{HTS comparison} (equation-faithful reimplementation of the
entropy-conditioned two-parameter map of \citet{balanya2024adaptive}, fitted
with its original NLL objective and with this paper's top-label BCE). Clean:
CIFAR-100/ViT-B/16 $\mathrm{ECE}_{15}$, five seeds. C100-C: deployable
protocol, three seeds, with the three-tier hierarchical CI for
SRTS-BCE $-$ HTS (negative favours SRTS-BCE). Budget $250$: seed-level
hierarchical CIs of SRTS-BCE $-$ HTS on the identical resampling draws.
\textbf{Bold}: interval excludes zero.}
\label{tab:hts_comparison}
\begin{tabular}{lcccccc}
\toprule
 & & & C100-C & \multicolumn{3}{c}{SRTS $-$ HTS at budget $250$} \\
\cmidrule(lr){5-7}
Variant & Clean ECE$_{15}$ & C100-C ECE$_{15}$ & SRTS$-$HTS (hier.) & ViT-B/16 & DeiT-S & Swin-T \\
\midrule
HTS-NLL & $2.07{\pm}0.51$ & $6.96{\pm}0.30$ & \textbf{$[-3.70,-1.12]$} & \textbf{$[-0.95,-0.40]$} & \textbf{$[-1.58,-1.24]$} & \textbf{$[-1.74,-1.07]$} \\
HTS-BCE & $1.63{\pm}0.25$ & $5.41{\pm}0.54$ & \textbf{$[-1.47,-0.22]$} & \textbf{$[-0.60,-0.13]$} & \textbf{$[-0.68,-0.20]$} & \textbf{$[-0.86,-0.38]$} \\
\bottomrule
\end{tabular}
\end{table*}

\textbf{HTS comparison.} HTS \citep{balanya2024adaptive} is the closest
existing low-capacity conditional map: a two-parameter temperature function
$T(z)=\sigma_{\mathrm{SP}}(w_H\log\bar H(z)+b)$ of the normalized prediction
entropy $\bar H(z)=H(\sigma_{\mathrm{SM}}(z))/\log K$, proposed specifically
for data-scarce calibration. We reimplement it equation-faithfully and fit
the two parameters by Nelder--Mead, initialised at the matching-objective
scalar temperature, under both its original NLL objective (HTS-NLL) and this
paper's top-label BCE (HTS-BCE); temperatures are strictly positive, so
top-1 is preserved (verified: zero argmax flips). Implementation details:
entropy is normalized by $\log K$ and floored at $10^{-12}$ before the
logarithm; the temperature is $\sigma_{\mathrm{SP}}(w_H u + b)$ with a
$10^{-4}$ numerical floor; both variants are fit by Nelder--Mead
($x_{\mathrm{atol}}{=}10^{-6}$, $f_{\mathrm{atol}}{=}10^{-10}$, max $2{,}000$
iterations, run to convergence, no early stopping), initialised at
$w_H{=}0$, $b{=}\sigma_{\mathrm{SP}}^{-1}(T_0)$ with $T_0$ the
matching-objective scalar temperature. Neither HTS nor SRTS-BCE receives any
hyperparameter search; both use deterministic optimizers on the identical
calibration splits. Table~\ref{tab:hts_comparison}
reports the full comparison matrix. On clean CIFAR-100/ViT-B/16, HTS-BCE ($1.63$)
matches scalar TvA-TS ($1.65$): under the aligned objective, entropy
conditioning alone does not close the gap to SRTS-BCE ($0.96$). On the
identical budget-resampling draws, the SRTS$-$HTS-BCE seed-level interval at
$250$ samples excludes zero on all three backbones, and under the deployable
CIFAR-100-C protocol every tier of the SRTS$-$HTS comparison excludes zero
(HTS recomputes each corrupted sample's temperature from its own entropy).
HTS-BCE improves on HTS-NLL everywhere, a further external replication of
the objective axis. HTS retains its attractions --- two parameters and no
router to store --- and its entropy signal is one-dimensional; the
comparison indicates that at this budget the learned six-signal risk
ordering separates more calibration-relevant structure than entropy alone.

\textbf{Proper-score improvements.}
Table~\ref{tab:phi_proper_scores} reports
$\Phi_{\ell} = \ell(\text{NoTS}) - \ell(\text{calibrated})$ for the
flagship cell under TopBCE, NLL and Brier. Every calibrator recovers an
essentially identical NLL and Brier improvement; the conditioned maps
add ${\approx}0.005$ TopBCE over the scalar anchors and are mutually
indistinguishable at seed noise. The paper's benefit is therefore
primarily top-label calibration-error reduction, not a general
improvement in multiclass probabilistic fit.

% Proper-score improvements Phi_ell = ell(NoTS) - ell(calibrated), flagship
% ViT-B/16 / CIFAR-100, three seeds (per-seed differences; mean+-std).
% Source: canonical deployed-pipeline recomputation archived with the
% matched-capacity run (phi_proper_scores.json); ECE gates reproduced the
% frozen registry values for every row (TvA 1.5370, SRTS 0.8276,
% SMART+BCE 0.8573, SRTS-NLL 1.70, PWLinear-3 0.87).
\begin{table*}[t]
\centering
\small
\setlength{\tabcolsep}{10pt}
\begin{tabular}{lrrr}
\toprule
Method & $\Phi_{\text{TopBCE}}$ & $\Phi_{\text{NLL}}$ & $\Phi_{\text{Brier}}$ \\
\midrule
TS-NLL     & $0.0577 \pm 0.0057$ & $0.0828 \pm 0.0078$ & $0.0100 \pm 0.0026$ \\
TvA-TS     & $0.0587 \pm 0.0053$ & $0.0814 \pm 0.0069$ & $0.0106 \pm 0.0026$ \\
SRTS-NLL   & $0.0613 \pm 0.0027$ & $\mathbf{0.0850 \pm 0.0056}$ & $0.0108 \pm 0.0020$ \\
SMART+BCE  & $0.0628 \pm 0.0029$ & $0.0822 \pm 0.0063$ & $0.0113 \pm 0.0020$ \\
PWLinear-3 & $\mathbf{0.0643 \pm 0.0023}$ & $0.0822 \pm 0.0049$ & $\mathbf{0.0116 \pm 0.0018}$ \\
SRTS-BCE   & $0.0637 \pm 0.0020$ & $0.0821 \pm 0.0049$ & $\mathbf{0.0116 \pm 0.0018}$ \\
\bottomrule
\end{tabular}
\caption{\textbf{Proper-score improvements over the uncalibrated model}
($\Phi_{\ell} = \ell(\text{NoTS}) - \ell(\text{calibrated})$; higher is
better; ViT-B/16 / CIFAR-100, per-seed differences, mean$\pm$std over
three seeds). All calibrators recover an essentially identical NLL and
Brier improvement; the conditioned maps add only ${\approx}0.005$ TopBCE
over the scalars and are mutually indistinguishable.}
\label{tab:phi_proper_scores}
\end{table*}

\textbf{Input-scale sensitivity.} Refitting the continuous maps on the
pre-sigmoid router score $s(x)$ instead of $q(x)$, with no other change,
leaves the conclusions unchanged (PWLinear-3[s] reaches $4.52$ on
CIFAR-100-C; all fits converged): the result is not an artifact of the
score parameterisation. Per-seed metrics, per-cell CIFAR-100-C values,
and diagnostics are archived with the released code.

%==========================================================================
% D. Routing-signal ablations
%==========================================================================

\section{Routing-Signal Ablations}
\label{app:routing-ablations}

This section isolates \emph{which} conditioning signal carries the
benefit, holding the top-label BCE objective and the $K{=}3$ budget
fixed throughout.

\subsection{Motivating diagnostics: why risk-conditioned scaling is needed}
\label{app:motivating-diagnostics}

\begin{figure*}[t]
\centering
\includegraphics[width=\textwidth]{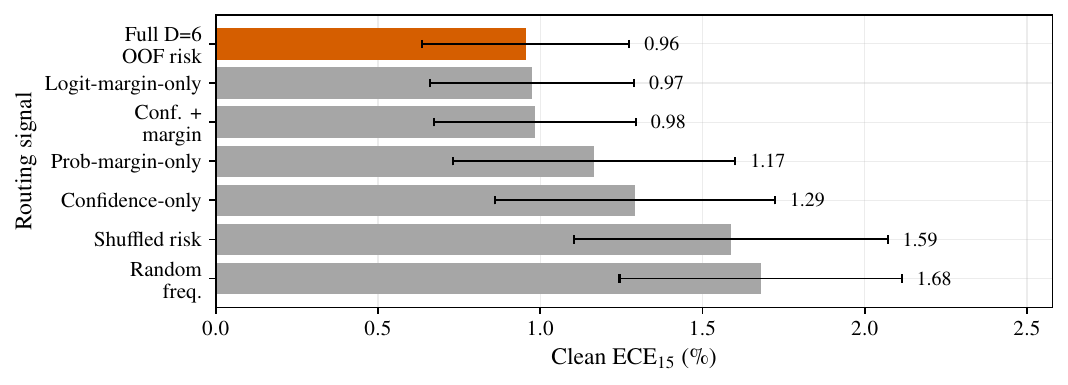}
\caption{\textbf{Motivating diagnostics for risk-conditioned scaling}
(clean CIFAR-100 / ViT-B/16; five seeds).
(a) At matched scalar-TvA confidence, empirical accuracy differs across
OOF-risk groups, exposing residual heterogeneity hidden by a global
confidence value.
(b) Scalar TvA-TS leaves the largest residual in the high-risk group,
while the fitted group temperatures differ across risk levels.
(c) The five-seed $K$-ablation illustrates the capacity trade-off:
$K{=}1$ underfits, $K{=}3$ captures most of the gain, and larger $K$
provides no stable additional benefit.
These diagnostics motivate risk conditioning but do not establish a
unique advantage for the six-signal router or the piecewise-constant map.}
\label{fig:motivation-why-routing}
\end{figure*}

Figure~\ref{fig:motivation-why-routing} summarises the motivating
diagnostics referenced in the main text: at matched scalar confidence,
empirical accuracy varies across OOF-risk groups (panel a); scalar TvA-TS
leaves its largest residual in the high-risk group (panel b); and the
$K$-ablation shows that $K{=}3$ captures most of the available gain
(panel c). Which routing signal actually carries the benefit is examined
separately below.

\subsection{Routing-signal ablation (which signal carries the benefit)}
\label{app:routing-signal}

This is the detailed evidence behind the one-line routing-signal claim in
the main text (the motivating-diagnostics figure and paragraph immediately
before the Method section). We hold the objective at top-label BCE and
$K{=}3$ fixed and
vary only the \emph{routing signal}; all rows share the identical $K{=}3$
group-wise BCE temperature fit on clean CIFAR-100 / ViT-B/16 (five seeds).

\begin{table*}[t]
\centering
\small
\setlength{\tabcolsep}{4pt}
\begin{tabular}{lrrrr}
\toprule
Routing signal & ECE$_{15}$ & NLL & Brier & eAURC \\
\midrule
Confidence-only & 1.29\,$\pm$\,0.43 & 0.3323\,$\pm$\,0.0294 & 0.1357\,$\pm$\,0.0077 & 0.0139\,$\pm$\,0.0030 \\
Logit-margin-only & 0.97\,$\pm$\,0.31 & 0.3302\,$\pm$\,0.0264 & 0.1349\,$\pm$\,0.0071 & 0.0120\,$\pm$\,0.0013 \\
Prob-margin-only & 1.17\,$\pm$\,0.44 & 0.3321\,$\pm$\,0.0287 & 0.1355\,$\pm$\,0.0076 & 0.0139\,$\pm$\,0.0030 \\
Confidence + logit-margin & 0.98\,$\pm$\,0.31 & 0.3305\,$\pm$\,0.0269 & 0.1350\,$\pm$\,0.0072 & 0.0122\,$\pm$\,0.0016 \\
\textbf{Full OOF risk (six signals)} & \textbf{0.96\,$\pm$\,0.32} & \textbf{0.3297\,$\pm$\,0.0257} & \textbf{0.1348\,$\pm$\,0.0070} & \textbf{0.0113\,$\pm$\,0.0011} \\
Random equal-frequency & 1.68\,$\pm$\,0.43 & 0.3322\,$\pm$\,0.0297 & 0.1362\,$\pm$\,0.0081 & 0.0142\,$\pm$\,0.0032 \\
Shuffled risk & 1.59\,$\pm$\,0.48 & 0.3325\,$\pm$\,0.0292 & 0.1363\,$\pm$\,0.0080 & 0.0143\,$\pm$\,0.0031 \\
\bottomrule
\end{tabular}
\caption{\textbf{Routing-signal ablation on clean CIFAR-100 / ViT-B/16 (top-label BCE, $K{=}3$, five-seed mean~$\pm$~std).} All rows share the identical $K{=}3$ group-wise BCE temperature fit and differ only in the routing signal. Lower is better for all metrics.}
\label{tab:routing_signal_ablation}
\end{table*}

Table~\ref{tab:routing_signal_ablation} confirms three points. (i)~The
signal matters: \emph{random} ($\mathrm{ECE}_{15}=1.68$) and
\emph{shuffled} ($1.59$) routing collapse to scalar TvA-TS behaviour
($K{=}1$ BCE $=1.65$), so the gain is from routing on a genuine reliability
signal, not from partitioning. (ii)~Logit-margin-only routing ($0.97$)
essentially matches the margin-based SMART+BCE head ($0.95$), but SRTS-BCE
uses the same signal under far tighter capacity control ($K{+}(D{+}1)$
scalars vs.\ a trained MLP). (iii)~Margin is a strong low-capacity routing signal: in this
clean routing-signal table, the full six-signal OOF risk router has slightly
lower point estimates than margin-only ($\mathrm{ECE}_{15}$/eAURC
$0.96$/$0.0113$ vs.\ $0.97$/$0.0120$; confidence-only $1.29$,
probability-margin-only $1.17$), but the paired comparison does not
statistically separate the two (\S\ref{app:margin-not-sufficient}). The
six-signal router is a sensible default, not a demonstrated improvement over
margin-only routing; the matched-margin CIFAR-100-C diagnostic below shows
margin routing matches SRTS in the high-margin region.

\begin{figure*}[t]
\centering
\includegraphics[width=\textwidth]{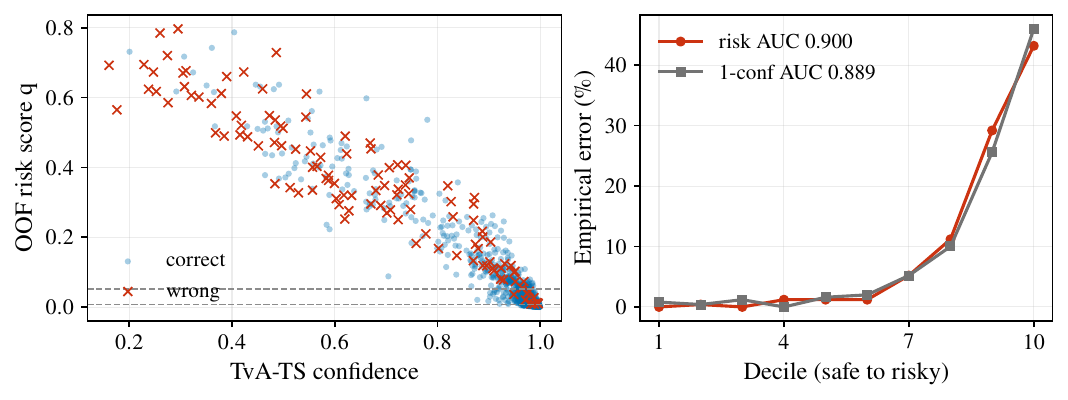}
\caption{\textbf{OOF risk vs.\ confidence (clean CIFAR-100 / ViT-B/16, seed 0).}
(a) The OOF risk score is tightly coupled to confidence; (b) it is a monotone
reliability ordering with only a \emph{modest} incremental edge over confidence
at ranking errors (AUROC $0.900$ vs.\ $0.889$). We do not claim a strong
risk-beyond-confidence separation from this view; the incremental value of the
full six-signal router is better quantified by the routing-signal ablation
(Table~\ref{tab:routing_signal_ablation}).}
\label{fig:risk-vs-confidence}
\end{figure*}

\subsection{Margin-only vs.\ learned-router routing: fixed-budget ablation}
\label{app:margin-not-sufficient}

We sharpen the previous point with two diagnostics that isolate the
\emph{logit margin} $m = z_{(1)} - z_{(2)}$ as a routing signal, using only
cached CIFAR-100 / CIFAR-100-C logits. These diagnostics use the main
deployable protocol: clean columns are held-out clean-test metrics;
CIFAR-100-C columns are corruption-test metrics over the full
$19\times5$ corruption/severity cells per seed, averaged per seed and then
reported as mean$\pm$std over three seeds. All temperatures and routers are
fit on clean validation only. SRTS uses the same clean-validation-fitted OOF
logistic risk router, clean-validation risk thresholds, and deployable
rerouting rule as the main text, so each corrupted sample is routed from its
own corrupted logits. Under this protocol, the SRTS-BCE rows reproduce the
three-seed fixed-$K{=}3$ values (clean $\mathrm{ECE}_{15}=0.817$,
CIFAR-100-C $4.63$, eAURC $0.0431$).

\begin{table*}[t]
\centering
\small
\setlength{\tabcolsep}{5pt}
\begin{tabular}{l rr rr}
\toprule
& \multicolumn{2}{c}{Clean CIFAR-100} & \multicolumn{2}{c}{CIFAR-100-C} \\
\cmidrule(lr){2-3}\cmidrule(lr){4-5}
Grouping rule ($K{=}3$) & ECE$_{15}$ & eAURC & ECE$_{15}$ & eAURC \\
\midrule
Margin tertiles & 0.837\,$\pm$\,0.247 & 0.0117\,$\pm$\,0.0013 & 4.695\,$\pm$\,0.778 & 0.0438\,$\pm$\,0.0042 \\
\textbf{SRTS risk tertiles} & \textbf{0.817\,$\pm$\,0.285} & \textbf{0.0114\,$\pm$\,0.0013} & \textbf{4.626\,$\pm$\,0.839} & \textbf{0.0431\,$\pm$\,0.0045} \\
Shuffled risk & 1.538\,$\pm$\,0.471 & 0.0136\,$\pm$\,0.0029 & 5.713\,$\pm$\,1.173 & 0.0458\,$\pm$\,0.0054 \\
Oracle correctness & 0.249\,$\pm$\,0.006 & 0.0000\,$\pm$\,0.0000 & 0.514\,$\pm$\,0.007 & 0.0000\,$\pm$\,0.0000 \\
\bottomrule
\end{tabular}
\caption{\textbf{Margin is a strong routing signal: fixed-budget routing ablation} (ViT-B/16, 3 seeds). Clean held-out and CIFAR-100-C corruption-test $\mathrm{ECE}_{15}$/eAURC (mean$\pm$std over seeds); lower is better. Objective, validation split, and $K{=}3$ budget are held fixed and only the grouping rule varies (protocol, statistical test, and within-table-comparison scope in the text). The oracle-correctness row routes by the true correct/incorrect label (label-leaking, non-deployable, diagnostic reference only); its near-zero eAURC is an artifact of using correctness labels for routing.}
\label{tab:router_ablation_margin}
\end{table*}

Table~\ref{tab:router_ablation_margin} fixes the objective, the validation
split, and the $K{=}3$ budget, and varies only the grouping rule. At the same
$K{=}3$ budget and objective, margin tertiles are already strong, confirming
that the logit gap is a powerful reliability signal. The full six-signal OOF
risk tertiles have slightly lower point estimates on clean and corrupted
$\mathrm{ECE}_{15}$/eAURC than margin tertiles (not statistically separated;
next paragraph), while shuffled routing collapses. Thus SRTS does not ignore
the margin; it uses the margin together with confidence,
entropy, and logit-scale signals inside a lower-capacity OOF risk router.
These diagnostic numbers are intended for within-table comparison among
routing signals; except for the SRTS-BCE row, they are not replacements for
the main-paper aggregate metrics.

\paragraph{The full router does not \emph{significantly} beat margin-only.}
The point-estimate improvement of the full OOF risk router over margin-only
tertiles is real but small, so we test it directly. Holding the $K{=}3$
grouped-BCE fitting identical and varying \emph{only} the partition (full
six-signal OOF logistic risk router vs.\ logit-margin tertiles), we
paired-bootstrap the test set across CIFAR-100 / ViT-B/16 (three seeds) and
four CIFAR-100 ConvNets (ResNet-20/32/44/56). Per-cell differences straddle
zero in both directions (e.g.\ one ViT seed favours risk by $0.75$ pp, another
favours margin by $0.48$ pp), and the cluster bootstrap across the seven cells
gives a mean $\Delta(\text{risk}-\text{margin})=-0.17$ pp with 95\% CI
$[-0.51,\,+0.15]$ --- \emph{including zero}. The full multi-signal router is
therefore not a statistically significant improvement over routing by the
logit margin alone. The ingredients that matter are conditioning on a
genuine reliability signal (margin or its risk generalisation) and the BCE
objective, not the specific choice of a multi-signal logistic router over
margin.

\begin{table*}[t]
\centering
\small
\setlength{\tabcolsep}{4pt}
\begin{tabular}{ll rrrr}
\toprule
Margin tertile & Calibrator & ECE$_{15}$ & NLL & AURC & eAURC \\
\midrule
\multirow{4}{*}{low} & TvA-TS & 8.668 & 1.365 & 0.2105 & 0.1242 \\
 & Margin-K3 & 6.327 & 1.355 & 0.2120 & 0.1258 \\
 & SRTS-K3 & 6.297 & 1.355 & 0.2103 & 0.1241 \\
 & SMART+BCE & 6.433 & 1.355 & 0.2122 & 0.1260 \\
\midrule
\multirow{4}{*}{mid} & TvA-TS & 3.135 & 0.370 & 0.0482 & 0.0450 \\
 & Margin-K3 & 3.424 & 0.379 & 0.0473 & 0.0441 \\
 & SRTS-K3 & 3.295 & 0.371 & 0.0422 & 0.0390 \\
 & SMART+BCE & 3.401 & 0.378 & 0.0396 & 0.0364 \\
\midrule
\multirow{4}{*}{high} & TvA-TS & 1.222 & 0.117 & 0.0157 & 0.0154 \\
 & Margin-K3 & 0.971 & 0.122 & 0.0154 & 0.0150 \\
 & SRTS-K3 & 1.014 & 0.122 & 0.0132 & 0.0129 \\
 & SMART+BCE & 0.997 & 0.127 & 0.0154 & 0.0151 \\
\bottomrule
\end{tabular}
\caption{\textbf{Matched-margin CIFAR-100-C risk (deployable rerouting, 3 seeds).} These are corruption-test metrics only, computed per corruption/severity cell within a logit-margin tertile and averaged over the full 19 corruptions $\times$ 5 severities for each seed, then over three seeds. Margin-tertile thresholds are fixed on clean validation. All calibrators, including the diagnostic Margin-K3 temperatures and SRTS-K3 temperatures/router, are fit or refit on clean validation only and applied unchanged to CIFAR-100-C; SRTS uses deployable rerouting from each corrupted sample's own logits. No oracle row or corrupted labels are used. In the high-margin region margin routing is competitive on $\mathrm{ECE}_{15}$; in the low-margin region the full six-signal router has slightly lower point-estimate ECE$_{15}$ and eAURC than margin routing. These within-tertile diagnostics are descriptive and do not establish router superiority; they are not replacements for the main-paper aggregate metrics. Lower is better.}
\label{tab:matched_margin_c100c}
\end{table*}

\begin{figure*}[t]
\centering
\includegraphics[width=0.82\textwidth]{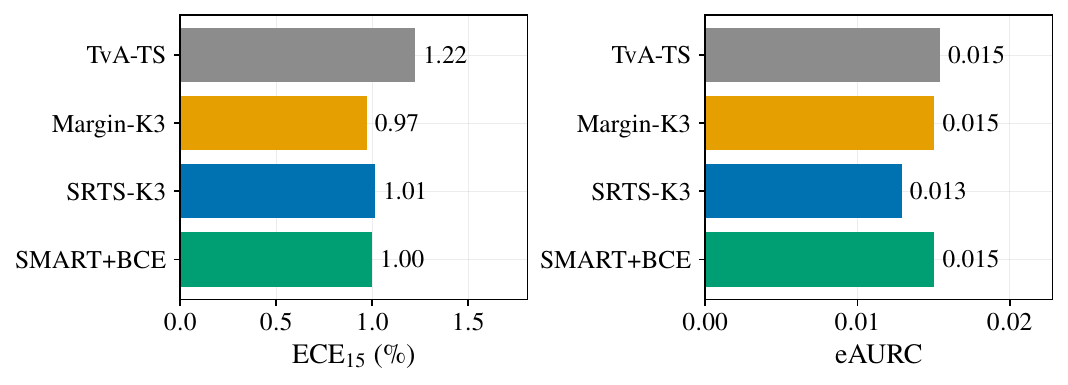}
\caption{\textbf{High-margin CIFAR-100-C aggregate diagnostic.} Within the
high-margin tertile, margin routing matches SRTS on $\mathrm{ECE}_{15}$, while
SRTS retains the lower eAURC. This 3-seed corruption-test diagnostic averages
the full $19\times5$ CIFAR-100-C protocol inside the high-margin subset:
margin tertiles are fixed on clean validation, temperatures/routers are fit on
clean validation only, and SRTS uses deployable rerouting from each corrupted
sample's own logits. No corrupted labels or oracle routing are used; the panel
is intended for within-tertile mechanism comparison, not as a replacement for
the aggregate main-paper metrics.}
\label{fig:high-margin-c100c}
\end{figure*}

Table~\ref{tab:matched_margin_c100c} and
Figure~\ref{fig:high-margin-c100c} give the matched-margin view. Margin is
nearly sufficient in the high-margin region, where margin routing can match or
slightly beat SRTS on $\mathrm{ECE}_{15}$. SRTS has lower point estimates
for low-margin $\mathrm{ECE}_{15}$ and for eAURC across all margin tertiles
(descriptive; the overall paired difference is not statistically resolved),
so margin-only routing recovers most, but on these point estimates not all,
of the effect.

\subsection{Alternative Low-Capacity Routing Targets}
\label{app:residual-routing}

As a further diagnostic we vary the \emph{routing target} at the same fixed
$K{=}3$ budget, top-label-BCE objective, and clean-validation split
(Table~\ref{tab:residual_router_ablation}), comparing the deployed OOF
wrongness-risk router against margin-tertile routing and two residual-based
routers that group by the signed or absolute top-label calibration residual.

% Alternative low-capacity routing-target ablation (clean CIFAR-100 / ViT-B/16,
% 3 seeds, K=3, top-label BCE, same clean-validation split). Values from the
% residual-router diagnostic rerun 20260713_232538 (official smECE). Requires: \usepackage{booktabs}.
\begin{table*}[t]
\centering
\small
\setlength{\tabcolsep}{6pt}
\begin{tabular}{l cccccc}
\toprule
Routing target & ECE$_{15}$ & AdaECE$_{15}$ & smECE & NLL & Brier & eAURC \\
\midrule
TvA-TS / TS-BCE ($K{=}1$)          & $1.537 \pm 0.417$ & $1.596 \pm 0.673$ & $1.562 \pm 0.417$ & $0.3253$ & $0.1335$ & $0.0135$ \\
\textbf{OOF wrongness-risk (SRTS-BCE)} & $\mathbf{0.814 \pm 0.264}$ & $0.957 \pm 0.134$ & $1.073 \pm 0.151$ & $0.3247$ & $0.1326$ & $0.0115$ \\
Margin-tertile ($K{=}3$)           & $0.836 \pm 0.256$ & $\mathbf{0.823 \pm 0.216}$ & $\mathbf{1.037 \pm 0.179}$ & $0.3246$ & $0.1326$ & $0.0117$ \\
Signed-residual ($K{=}3$)          & $1.093 \pm 0.148$ & $0.968 \pm 0.306$ & $1.140 \pm 0.145$ & $\mathbf{0.3201}$ & $\mathbf{0.1322}$ & $\mathbf{0.0099}$ \\
Absolute-residual ($K{=}3$)        & $0.862 \pm 0.276$ & $0.905 \pm 0.249$ & $1.072 \pm 0.195$ & $0.3241$ & $0.1327$ & $0.0112$ \\
\bottomrule
\end{tabular}
\caption{\textbf{Alternative low-capacity routing targets} (clean CIFAR-100 /
ViT-B/16, 3 seeds; $K{=}3$, top-label BCE). All routed rows fix the objective,
the $K{=}3$ budget, and the clean-validation split, and vary only the routing
target; TvA-TS ($K{=}1$) is the no-routing reference. The OOF wrongness-risk
router (the deployed SRTS-BCE design) is best on the primary $\mathrm{ECE}_{15}$
metric; margin-tertile routing is highly competitive and better on
AdaECE$_{15}$/smECE; signed-residual routing improves NLL/Brier/eAURC but
worsens $\mathrm{ECE}_{15}$; absolute-residual routing is close but does not
improve $\mathrm{ECE}_{15}$ over wrongness-risk. This is a metric trade-off
under the present small-validation ID clean setting, not a universal ranking.
Bold marks the best (lowest) value per column. Lower is better.}
\label{tab:residual_router_ablation}
\end{table*}

\begin{itemize}
\item The OOF wrongness-risk router remains best on the primary
$\mathrm{ECE}_{15}$ metric.
\item Margin-tertile routing is highly competitive and even better on
AdaECE$_{15}$/smECE, confirming that the margin is a strong reliability signal.
\item Signed-residual routing improves NLL, Brier, and eAURC, but worsens
$\mathrm{ECE}_{15}$, so it is not suitable as the main ECE-oriented router.
\item Absolute-residual routing is close but still does not improve over
wrongness-risk routing on mean $\mathrm{ECE}_{15}$.
\end{itemize}

Direct residual routing did not improve the primary $\mathrm{ECE}_{15}$ metric
over the more stable OOF wrongness-risk proxy under $n \approx 2{,}500$. We
therefore retain OOF wrongness-risk routing as the main SRTS-BCE design; the
margin and residual routers are metric-specific trade-offs under the present
small-validation ID clean setting not universal improvements.

% Archived stage4 asset (bar values unlabeled, no data file / generator in
% this repo); shown at full text width because the natural canvas is ~11 in
% and a single-column inclusion renders its fonts unreadably small.
\begin{figure*}[t]
\centering
\includegraphics[width=\textwidth]{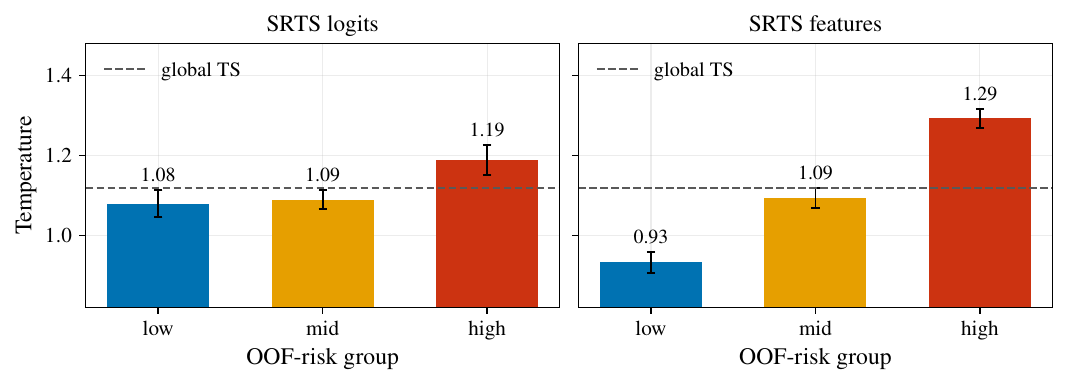}
\caption{\textbf{Per-group temperatures learned by SRTS variants
(NLL-objective diagnostic).} Under NLL fitting, SRTS-NLL group temperatures
cluster tightly around the global-TS value. Under BCE fitting
(Table~\ref{tab:router_causality}, \S\ref{app:router-causality}), the per-group temperature
spread is substantially larger (range $0.6$--$0.9$ on seed 0 vs.\ TvA-TS
single $T \approx 0.76$), which is consistent with the BCE-objective
$K{=}3$ risk router achieving the $0.73$ pp gain over TvA-TS.}
\label{fig:appendix-group-temperatures}
\end{figure*}

% Archived stage4 asset — same full-width treatment as the figure above.
\begin{figure*}[t]
\centering
\includegraphics[width=\textwidth]{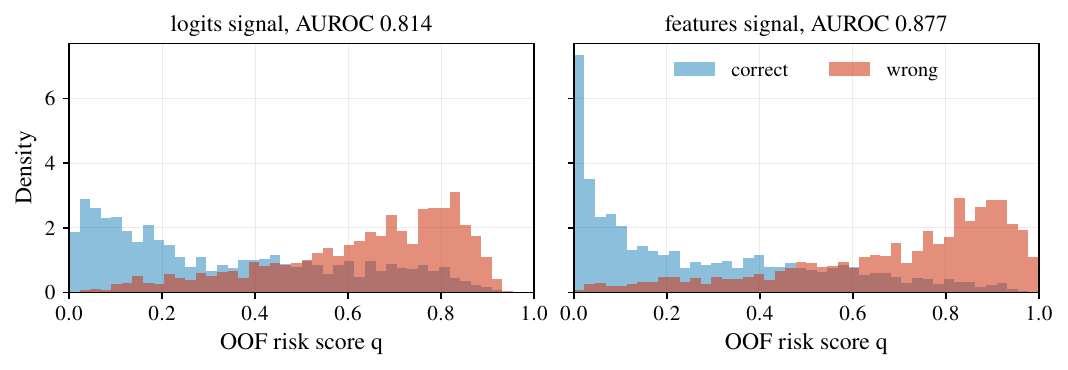}
\caption{\textbf{Risk-score density by variant: correct vs.\ wrong predictions.}
Logit-derived signals separate correct from wrong predictions on the
validation out-of-fold split. Under NLL fitting, stronger separation does
not translate to lower test-side $\mathrm{ECE}_{15}$
(\S~\ref{app:router-causality}, NLL block); under BCE fitting, the same
separation is exploited to achieve the $0.73$ pp improvement over TvA-TS
(\S~\ref{app:router-causality}, BCE block).}
\label{fig:appendix-risk-score-separation}
\end{figure*}

\begin{figure*}[t]
\centering
\includegraphics[width=\textwidth]{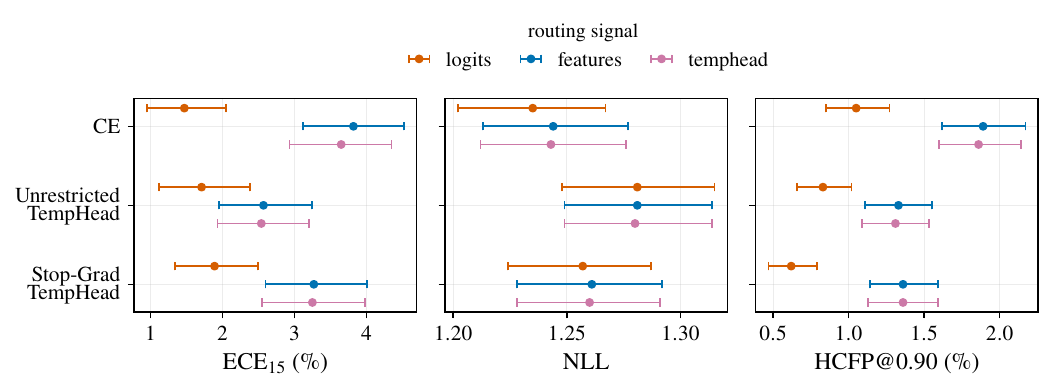}
\caption{\textbf{Bootstrap $95\%$ CIs across signal-routing variants
(NLL-objective diagnostic; $B{=}1000$, values as in
Table~\ref{tab:stage4_srts_bootstrap}).} On NLL the three signal variants
are indistinguishable (near-complete CI overlap), whereas on
$\mathrm{ECE}_{15}$ and HCFP@$0.90$ the logit-input variant is clearly
preferable (CIs disjoint from the feature/temphead variants on the CE and
Stop-Grad bases); logit-input routing is the chosen default for the main
paper's SRTS-BCE method on these overfitting-safety grounds.}
\label{fig:appendix-signal-variant-ci}
\end{figure*}

%==========================================================================
% E. Budget sensitivity
%==========================================================================

\section{Budget Sensitivity}
\label{app:budget}

This section reports validation-budget sensitivity for the flagship
CIFAR-100 / ViT-B/16 cell.

\begin{table}[t]
\centering
\small
\setlength{\tabcolsep}{4pt}
\begin{tabular}{lrrrr}
\toprule
Method & $10\%$ & $25\%$ & $50\%$ & $100\%$ \\
\midrule
TvA-TS & 1.81 & 1.65 & 1.56 & 1.54 \\
\textbf{SRTS-BCE $K{=}3$ risk} & \textbf{1.45} & \textbf{1.18} & \textbf{0.93} & \textbf{0.81} \\
SMART+BCE & 1.95 & 1.41 & 1.00 & 0.83 \\
\bottomrule
\end{tabular}
\caption{\textbf{BCE-objective validation-size sensitivity on ViT-B/16 /
CIFAR-100.} Mean $\mathrm{ECE}_{15}$ over the three backbone seeds and
five calibration-subsamples per seed. SRTS-BCE has the lowest absolute
ECE at every tested fraction ($10$--$100\%$), including at $10\%$
($1.45$ vs.\ TvA-TS $1.81$, SMART+BCE $1.95$). However, the absolute
ECE gap relative to TvA-TS shrinks from $0.73$ pp at $100\%$ to $0.36$ pp
at $10\%$, because SRTS-BCE's ECE worsens by $0.64$ pp as budget falls
from $100\%$ to $10\%$ while TvA-TS worsens by only $0.27$ pp.
TvA-TS is the most rate-stable single-parameter fallback at extreme budgets.}
\label{tab:bce-valsize-sensitivity}
\end{table}

\begin{figure}[t]
\centering
\includegraphics[width=0.95\columnwidth]{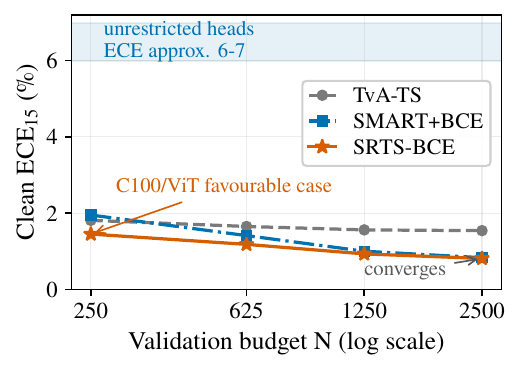}
\caption{\textbf{Validation-budget trade-off on CIFAR-100 / ViT-B/16}
(companion plot to Table~\ref{tab:bce-valsize-sensitivity}). SRTS-BCE has
the lowest ECE at the smallest validation budget and converges with
SMART+BCE at the full budget, while unrestricted adaptive heads remain
poorly calibrated across budgets. This supports the fixed low-capacity
routing design under small validation budgets.}
\label{fig:validation-budget}
\end{figure}

\paragraph{Budget breadth.} The flagship sweep
(Table~\ref{tab:bce-valsize-sensitivity}) and the second-order
approximation's $O(K/n)$ variance term motivate the working rule that risk grouping needs a few
hundred effective validation samples \emph{per group}; the resampling
experiment below adds a paired analysis and a Tiny/ViT-B/16 boundary.

\subsection{Calibration-resample budget experiment}
\label{app:budget-resampling}

% from results/budget_resampling/aggregate.json. Do not hand-edit numbers.
\begin{table}[t]
\centering
\footnotesize
\setlength{\tabcolsep}{2.8pt}
\caption{\textbf{Calibration-budget resampling on CIFAR-100 / ViT-B/16}
($\mathrm{ECE}_{15}$ \%). Each calibrator is fit on $20$ stratified
calibration subsamples per training seed (five seeds; the $2500$ row is the
full split, no resampling variance). \emph{SD}: standard deviation over
calibration resamples. \emph{CI}: draw-level bootstrap $95\%$ interval of
the paired mean difference vs.\ TvA-TS (negative favours the method);
descriptive only --- seed-level inference is in
Table~\ref{tab:budget_hierarchical}. \emph{Win}: fraction of resamples with
lower $\mathrm{ECE}_{15}$ than TvA-TS.}
\label{tab:budget_resampling}
\begin{tabular}{llrrcr}
\toprule
Budget & Method & Mean & SD & Paired CI vs TvA & Win \\
\midrule
250 & TvA-TS & $1.96$ & $0.72$ & --- & --- \\
 & Margin-K3 & $1.55$ & $0.64$ & $[-0.53,-0.28]$ & 77\% \\
 & SMART+BCE & $2.03$ & $0.72$ & $[-0.06,+0.23]$ & 47\% \\
 & \textbf{SRTS-BCE} & $1.60$ & $0.70$ & $[-0.48,-0.23]$ & 81\% \\
\addlinespace
625 & TvA-TS & $1.81$ & $0.62$ & --- & --- \\
 & Margin-K3 & $1.24$ & $0.53$ & $[-0.66,-0.47]$ & 86\% \\
 & SMART+BCE & $1.43$ & $0.50$ & $[-0.48,-0.28]$ & 76\% \\
 & \textbf{SRTS-BCE} & $1.23$ & $0.53$ & $[-0.68,-0.49]$ & 89\% \\
\addlinespace
1250 & TvA-TS & $1.75$ & $0.55$ & --- & --- \\
 & Margin-K3 & $1.11$ & $0.46$ & $[-0.70,-0.58]$ & 99\% \\
 & SMART+BCE & $1.18$ & $0.45$ & $[-0.63,-0.50]$ & 93\% \\
 & \textbf{SRTS-BCE} & $1.12$ & $0.46$ & $[-0.69,-0.57]$ & 98\% \\
\addlinespace
2500 & TvA-TS & $1.65$ & $0.44$ & --- & --- \\
 & Margin-K3 & $0.97$ & $0.32$ & $[-0.82,-0.56]$ & 100\% \\
 & SMART+BCE & $0.95$ & $0.29$ & $[-0.87,-0.53]$ & 100\% \\
 & \textbf{SRTS-BCE} & $0.96$ & $0.32$ & $[-0.84,-0.55]$ & 100\% \\
\bottomrule
\end{tabular}
\end{table}

To probe the bias--variance account directly we resample the calibration
split itself. For each training seed we draw $20$ stratified subsamples at
each budget $\in\{250,625,1250,2500\}$, fit TvA-TS, Margin-K3, SMART+BCE and
SRTS-BCE on each, and evaluate on the fixed test set
(Table~\ref{tab:budget_resampling}; the table's draw-level paired CIs are
descriptive --- inference is the seed-level hierarchical analysis of
Table~\ref{tab:budget_hierarchical}). The two low-capacity conditional maps
(SRTS-BCE, Margin-K3) degrade gracefully: at the smallest budget ($250$)
SRTS-BCE still beats TvA-TS on $81\%$ of resamples. The higher-capacity
SMART+BCE head instead becomes unstable at $250$ samples --- its mean
$\mathrm{ECE}_{15}$ ($2.03$) is no better than the scalar ($1.96$) and it
wins only $47\%$ of resamples --- consistent with a capacity/variance
penalty that the ten-parameter map avoids. As the budget grows the three
adaptive methods converge and all beat the scalar at $2500$ (win rate
$100\%$).

% from results/budget_resampling/hierarchical.json. Do not hand-edit numbers.
\begin{table*}[t]
\centering
\footnotesize
\setlength{\tabcolsep}{5pt}
\caption{\textbf{Hierarchical (training-seed $\to$ calibration-draw)
$95\%$ CIs for the budget experiment} ($\Delta\,\mathrm{ECE}_{15}$, negative
favours SRTS-BCE; $B{=}20{,}000$). Draws within a training seed share the
model and test set, so seeds are the top-level resampling unit within each
backbone; the aggregate row adds a backbone level
(backbone$\to$seed$\to$draw) and is a secondary summary --- the per-backbone
intervals are the primary replication evidence. \textbf{Bold}: interval
excludes zero.}
\label{tab:budget_hierarchical}
\begin{tabular}{lcccc}
\toprule
 & \multicolumn{2}{c}{Budget $250$} & \multicolumn{2}{c}{Budget $2500$} \\
\cmidrule(lr){2-3}\cmidrule(lr){4-5}
Cell & SRTS $-$ TvA & SRTS $-$ SMART & SRTS $-$ TvA & SRTS $-$ SMART \\
\midrule
C100 / ViT-B/16 (5 seeds) & $[-0.72,+0.01]$ & \textbf{$[-0.59,-0.28]$} & \textbf{$[-0.84,-0.55]$} & $[-0.03,+0.06]$ \\
C100 / DeiT-S (3 seeds) & \textbf{$[-1.94,-1.17]$} & \textbf{$[-1.34,-0.30]$} & \textbf{$[-2.88,-1.79]$} & \textbf{$[-0.29,-0.15]$} \\
C100 / Swin-T (3 seeds) & \textbf{$[-1.30,-0.94]$} & \textbf{$[-0.59,-0.25]$} & \textbf{$[-2.11,-1.54]$} & $[-0.39,+0.14]$ \\
\midrule
Aggregate (backbone$\to$seed$\to$draw; secondary) & \textbf{$[-1.57,-0.39]$} & \textbf{$[-0.87,-0.35]$} & \textbf{$[-2.42,-0.72]$} & $[-0.24,+0.02]$ \\
\bottomrule
\end{tabular}
\end{table*}

\noindent\textbf{Hierarchical inference and cross-backbone replication.}
Draws within a training seed share the model and the fixed test set, so they
are not independent units; Table~\ref{tab:budget_hierarchical} therefore
reports a two-level (training-seed $\to$ calibration-draw) bootstrap with
method pairing preserved within each draw, alongside replications of the
identical protocol on CIFAR-100 / DeiT-S and Swin-T (three seeds each). All three backbones give a
SRTS$-$SMART interval at $250$ that excludes zero (ViT $[-0.59,-0.28]$,
DeiT-S $[-1.34,-0.30]$, Swin-T $[-0.59,-0.25]$). Against the scalar the
seed-level SRTS$-$TvA interval at $250$ excludes zero on DeiT-S
($[-1.94,-1.17]$) and Swin-T ($[-1.30,-0.94]$); on the flagship ViT-B/16 it
is directional --- $[-0.72,+0.01]$ narrowly includes zero, with four of five
seeds negative and an $81\%$ draw-level win rate. A secondary three-level
aggregate that also resamples the backbone level
(backbone$\to$seed$\to$draw) gives $[-1.57,-0.39]$ vs.\ TvA-TS and
$[-0.87,-0.35]$ vs.\ SMART+BCE at $250$; the per-backbone intervals remain
the primary replication evidence. At the full budget SRTS$-$SMART is
not separable on ViT or Swin (intervals include zero; DeiT-S excludes),
matching the clean-cell tie: the small-budget separation, not the full-budget
comparison, is where capacity control shows.

% from results/budget_resampling/hierarchical_multi.json. No hand-edits.
\begin{table*}[t]
\centering
\footnotesize
\setlength{\tabcolsep}{5pt}
\caption{\textbf{Metric sensitivity of the $250$-sample result}
(seed-level hierarchical $95\%$ CIs of $\Delta$; negative favours SRTS-BCE;
same draws as the $\mathrm{ECE}_{15}$ analysis, official smECE estimator).
\textbf{Bold}: interval excludes zero.}
\label{tab:budget_multimetric}
\begin{tabular}{lcccc}
\toprule
 & \multicolumn{2}{c}{AdaECE$_{15}$} & \multicolumn{2}{c}{smECE} \\
\cmidrule(lr){2-3}\cmidrule(lr){4-5}
Cell & SRTS $-$ TvA & SRTS $-$ SMART & SRTS $-$ TvA & SRTS $-$ SMART \\
\midrule
ViT-B/16 & $[-0.71,+0.07]$ & \textbf{$[-0.54,-0.22]$} & $[-0.31,+0.27]$ & \textbf{$[-0.70,-0.34]$} \\
DeiT-S & \textbf{$[-2.11,-1.30]$} & \textbf{$[-1.23,-0.27]$} & \textbf{$[-1.20,-0.73]$} & \textbf{$[-1.39,-0.42]$} \\
Swin-T & \textbf{$[-1.38,-0.90]$} & \textbf{$[-0.57,-0.19]$} & \textbf{$[-0.92,-0.43]$} & \textbf{$[-0.73,-0.32]$} \\
\bottomrule
\end{tabular}
\end{table*}

\noindent\textbf{Metric sensitivity.} The $250$-sample result is not
specific to the $15$-bin estimator (Table~\ref{tab:budget_multimetric};
same draws, AdaECE$_{15}$ and official smECE, seed-level hierarchical
inference). SRTS$-$SMART excludes zero on all three backbones under both
alternative estimators. SRTS$-$TvA excludes zero on DeiT-S and Swin-T under
both; on ViT-B/16 the alternative-metric intervals include zero (point
estimates remain negative), so the flagship scalar-gap at $250$ leans on the
binned estimator while the cross-backbone replication does not.

\noindent\textbf{Two honest boundaries.} First, the calibration-error gain
is not a proper-score gain: on TopBCE the win rate at $250$
samples is low for every adaptive method (SRTS-BCE $0.21$, SMART+BCE
$0.06$). Second, in the boundary Tiny-ImageNet/ViT-B/16 cell
($10$ resamples/seed, three seeds) conditioning offers no advantage at any
budget --- every conditional map is within noise of or worse than TvA-TS,
and SMART+BCE is again the worst at $250$ samples ($2.61$ vs.\ $1.64$) ---
the same variance penalty without a transferable residual to offset it. The
bias--variance account therefore reads the scope map as a difference of two
terms --- routing gain $=$ (structured-residual benefit) $-$ (finite-budget
variance cost) --- positive on CIFAR-100/ViT-B/16 where a transferable
residual dominates and non-positive on Tiny-ImageNet/ViT-B/16 where it is
absent.

\subsection{Small-budget matched continuous-map test}
\label{app:p1-matched-maps}

% results/p1_matched_maps/hier.json. No hand-edits.
\begin{table*}[t]
\centering
\small
\setlength{\tabcolsep}{4pt}
\begin{tabular}{lcccccc}
\toprule
 & \multicolumn{2}{c}{ViT-B/16} & \multicolumn{2}{c}{DeiT-S} & \multicolumn{2}{c}{Swin-T} \\
\cmidrule(lr){2-3}\cmidrule(lr){4-5}\cmidrule(lr){6-7}
Map & $\Delta$ CI & win & $\Delta$ CI & win & $\Delta$ CI & win \\
\midrule
PWLinear-3 & $[-0.19,+0.06]$ & $73\%$ & $[-0.56,+0.03]$ & $93\%$ & $[-0.16,+0.07]$ & $98\%$ \\
LinearRiskTemp & \textbf{$[-0.48,-0.12]$} & $57\%$ & \textbf{$[-1.26,-0.48]$} & $83\%$ & \textbf{$[-0.79,-0.52]$} & $90\%$ \\
SplineRiskTemp & \textbf{$[-0.37,-0.18]$} & $57\%$ & \textbf{$[-0.83,-0.11]$} & $88\%$ & \textbf{$[-0.43,-0.06]$} & $92\%$ \\
\bottomrule
\end{tabular}
\caption{\textbf{Small-budget matched continuous-map test} ($n{=}250$; the
exact calibration draws, training seeds, router pipeline, optimizer, knots,
initialisation, and bounds of the budget-resampling protocol; frozen before
execution). Per backbone: seed$\to$draw hierarchical $95\%$ CI of
$\Delta\mathrm{ECE}_{15}=$ SRTS-BCE $-$ map (negative favours SRTS-BCE;
bold: CI excludes zero) and the map's win rate against TvA-TS. SRTS-BCE
vs.\ the matched three-anchor PWLinear-3 is statistically unresolved on all
three backbones; the linear and spline parameterizations of the same risk
score are significantly worse.}
\label{tab:p1_matched_maps}
\end{table*}

The full-budget matched-map control (\S\ref{app:matched-capacity}) leaves
open whether the grouped form matters specifically when the budget is
small. We therefore reran the matched continuous maps at $n{=}250$ on the exact
stored calibration draws of the budget-resampling experiment, with the
methods, seeds, metrics, and inference specified before this comparison was
run: the unchanged router pipeline, optimizer, knots, initialisation, and
bounds, with a per-draw gate requiring refit TvA-TS and SRTS-BCE to
reproduce the stored rows to $10^{-6}$.
Table~\ref{tab:p1_matched_maps} reports seed$\to$draw hierarchical CIs of
SRTS-BCE minus each map. Two facts emerge. First, SRTS-BCE vs.\ the
matched three-anchor PWLinear-3 is statistically unresolved on all three
backbones (ViT $[-0.19,+0.06]$, DeiT-S $[-0.56,+0.04]$, Swin-T
$[-0.16,+0.07]$; point estimates mildly favour SRTS-BCE, descriptive
only). Second, the linear and spline parameterizations of the same risk
score are significantly worse than SRTS-BCE on all three backbones (e.g.\
LinearRiskTemp: $[-0.48,-0.12]$, $[-1.26,-0.48]$, $[-0.79,-0.52]$).
Secondary metrics (AdaECE$_{15}$, smECE) show no systematic reversal of
these directions; TopBCE differences are small and slightly favour the
continuous maps (${\approx}0.02$--$0.03$, descriptive). The small-budget
evidence supports the three-level risk-conditioned family --- discrete grouping or
piecewise-linear interpolation between three anchors --- not the
piecewise-constant SRTS map specifically; SRTS-BCE remains the discrete instance we deploy, and map shape matters more than raw parameter count
(the two-parameter linear map is the weakest).

\subsection{Protocol-frozen cross-dataset small-budget follow-up (Tiny-ImageNet)}
\label{app:tiny-replication}

% results/budget_resampling/tiny_{deit,swin}_{rows,hier_p13}.json. No hand-edits.
\begin{table*}[t]
\centering
\footnotesize
\setlength{\tabcolsep}{5pt}
\caption{\textbf{Protocol-frozen cross-dataset small-budget follow-up (Tiny-ImageNet)}
(200 classes; three training seeds $\times$ 20 draws per budget). Mean test $\mathrm{ECE}_{15}$ (\%) over
seeds$\times$draws per method, hierarchical seed$\to$draw $95\%$ CIs of
$\Delta\mathrm{ECE}_{15}$ (negative favours SRTS-BCE; bold: CI excludes
zero), and SRTS-BCE's draw-paired win rate against TvA-TS. Cells, budgets,
draws, methods, metrics, inference, and outcome interpretation were frozen
before execution; no method, metric, optimizer, or draw selection changed
after inspection. On Swin-T the higher-capacity SMART+BCE is significantly
\emph{better} at the largest evaluated $2{,}500$-sample budget and the
ordering reverses as the budget shrinks. Tiny-ImageNet/ViT-B/16 remains the boundary result reported in the
budget-resampling section.}
\label{tab:tiny_replication}
\begin{tabular}{llcccccc}
\toprule
Backbone & $n$ & TvA-TS & SMART+BCE & SRTS-BCE & SRTS$-$TvA CI & SRTS$-$SMART CI & win \\
\midrule
DeiT-S & $250$ & $3.08$ & $2.51$ & $2.01$ & \textbf{$[-1.35,-0.80]$} & \textbf{$[-0.71,-0.28]$} & $95\%$ \\
DeiT-S & $625$ & $3.05$ & $1.74$ & $1.53$ & \textbf{$[-1.92,-1.22]$} & \textbf{$[-0.36,-0.07]$} & $100\%$ \\
DeiT-S & $1250$ & $3.06$ & $1.48$ & $1.28$ & \textbf{$[-2.11,-1.55]$} & \textbf{$[-0.30,-0.10]$} & $100\%$ \\
DeiT-S & $2500$ & $3.06$ & $1.16$ & $1.12$ & \textbf{$[-2.31,-1.73]$} & $[-0.14,+0.05]$ & $100\%$ \\
\addlinespace
Swin-T & $250$ & $2.85$ & $2.95$ & $2.03$ & \textbf{$[-1.02,-0.60]$} & \textbf{$[-1.27,-0.58]$} & $92\%$ \\
Swin-T & $625$ & $2.71$ & $1.73$ & $1.56$ & \textbf{$[-1.35,-0.87]$} & \textbf{$[-0.27,-0.07]$} & $98\%$ \\
Swin-T & $1250$ & $2.75$ & $1.48$ & $1.46$ & \textbf{$[-1.50,-1.02]$} & $[-0.12,+0.06]$ & $100\%$ \\
Swin-T & $2500$ & $2.73$ & $1.17$ & $1.30$ & \textbf{$[-1.59,-1.17]$} & \textbf{$[+0.08,+0.19]$} & $100\%$ \\
\addlinespace
\bottomrule
\end{tabular}
\end{table*}

The budget evidence above is CIFAR-100-specific. Before any Tiny-ImageNet
small-budget result was computed, we froze two replication cells ---
Tiny-ImageNet DeiT-S and Swin-T ($200$ classes,
$n_{\mathrm{val}}{=}5{,}000$) --- using only their already-reported
full-budget direction in the scope map; no small-budget statistic or
budget curve was inspected before freezing, and Tiny-ImageNet/ViT-B/16
(whose full-budget gain is absent) was excluded by the same rule and
remains reported as the boundary cell. The protocol is the identical
budget-resampling pipeline: budgets $\{250,625,1250,2500\}$, $20$
stratified draws per seed with subsample RNG
$\mathrm{seed}\cdot 10^{6}+\mathrm{budget}\cdot 10^{3}+\mathrm{draw}$,
unchanged fitting code and metrics, and seed$\to$draw hierarchical
bootstrap inference ($B{=}20{,}000$, bootstrap RNG $20260721$).

Table~\ref{tab:tiny_replication} shows that the small-budget separation
replicates on both frozen cells at $n{=}250$: SRTS $-$ TvA-TS and SRTS $-$
SMART+BCE exclude zero in SRTS-BCE's favour on both backbones (DeiT-S
$[-1.36,-0.80]$ / $[-0.73,-0.29]$; Swin-T $[-1.01,-0.59]$ /
$[-1.26,-0.58]$), with $n{=}250$ win rates against the scalar of $95\%$
and $92\%$.

Tiny-ImageNet/Swin-T provides the clearest within-cell capacity-control
example. SMART+BCE is significantly \emph{better} at the largest
evaluated $2{,}500$-sample budget ($[+0.08,+0.17]$; the full Tiny
validation split has $5{,}000$ samples), but the ordering reverses as the
budget
decreases (point estimates $+0.13 \to -0.02 \to -0.16 \to -0.91$), with
SRTS-BCE significantly better at $n{=}250$. This rules out an
interpretation based on SRTS-BCE being uniformly better for this backbone
and instead supports a budget-dependent capacity trade-off; on DeiT-S the
$2{,}500$-sample comparison is a tie ($[-0.14,+0.05]$), mirroring the
CIFAR pattern. The small-budget separation replicates on Tiny-ImageNet
DeiT-S and Swin-T, while ViT-B/16 remains the boundary cell; this is a
two-backbone replication, not a claim of cross-dataset universality.

% results/budget_resampling/tiny_multimetric_p13.json. No hand-edits.
\begin{table}[t]
\centering
\small
\setlength{\tabcolsep}{2pt}
\caption{\textbf{Metric sensitivity of the Tiny-ImageNet replication}
($n{=}250$; the exact Table~\ref{tab:tiny_replication} predictions, seeds,
and draws; protocol frozen before execution, run once). Seed$\to$draw
hierarchical $95\%$ CIs of $\Delta$AdaECE$_{15}$ and $\Delta$smECE
(official estimator); SMART denotes SMART+BCE; negative favours SRTS-BCE;
bold: CI excludes zero. All four comparisons are supported under both
alternative estimators.}
\label{tab:tiny_multimetric}
\begin{tabular}{llcc}
\toprule
Backbone & Comparison & $\Delta$AdaECE$_{15}$ & $\Delta$smECE \\
\midrule
DeiT-S & SRTS $-$ TvA-TS & \textbf{$[-1.45,-0.82]$} & \textbf{$[-0.92,-0.44]$} \\
DeiT-S & SRTS $-$ SMART & \textbf{$[-0.65,-0.25]$} & \textbf{$[-0.84,-0.35]$} \\
Swin-T & SRTS $-$ TvA-TS & \textbf{$[-1.28,-0.70]$} & \textbf{$[-0.67,-0.21]$} \\
Swin-T & SRTS $-$ SMART & \textbf{$[-1.19,-0.52]$} & \textbf{$[-1.45,-0.69]$} \\
\bottomrule
\end{tabular}
\end{table}

\paragraph{Metric sensitivity.} Using the same
Table~\ref{tab:tiny_replication} predictions, seeds, and draws at
$n{=}250$, we repeat both comparisons under AdaECE$_{15}$ and the
official smECE estimator. These alternative metrics were specified before
this comparison was run.
All four hierarchical CIs exclude zero in SRTS-BCE's favour on both
backbones (Table~\ref{tab:tiny_multimetric}), so the cross-dataset
separation does not depend on the $15$-bin ECE estimator.

\subsection{Fitted temperatures and routing stability}
\label{app:auditability}

% results/auditability/audit.json. No hand-edits.
\begin{table*}[t]
\centering
\small
\setlength{\tabcolsep}{3pt}
\begin{tabular}{@{}lll@{}}
\toprule
Diagnostic & Full budget & $n=250$ ($220$ fits) \\
\midrule
Group temps $T_{1..3}$, med.\,[IQR] & $.63\,[.63,.65]$ / $.73\,[.73,.74]$ / $.84\,[.82,.87]$ & $.17\,[.13,.73]$ / $.76\,[.69,.80]$ / $.89\,[.83,.94]$ \\
Router thresholds (mean$\pm$std) & $0.008{\pm}0.001$ / $0.042{\pm}0.007$ & --- \\
Bound saturation & $0\%$ & $0.0\%$ \\
Scalar fallback & $0\%$ & $0.0\%$ \\
Risk ordering monotone & $100\%$ & $93\%$ \\
Temperature ordering monotone & $100\%$ & $64\%$ \\
\midrule
\multicolumn{3}{@{}l}{Occupancy (low/mid/high): clean 34\% / 33\% / 33\%; CIFAR-100-C 19\% / 27\% / 55\%} \\
\multicolumn{3}{@{}l}{High-risk share by severity: s1: 44\%, s2: 50\%, s3: 54\%, s4: 59\%, s5: 66\%} \\
\bottomrule
\end{tabular}
\caption{\textbf{Auditability diagnostics for SRTS-BCE.} Full budget:
CIFAR-100/ViT-B/16, five seeds. $n{=}250$: all $220$ calibration draws
across the three CIFAR-100 backbones. Saturation: fitted group temperatures
at the $[0.05,20]$ bounds. Fallback: groups below the $50$-sample floor
(reverting to the global scalar). Risk-monotone: fits whose empirical
group error rates increase with the routed risk index.}
\label{tab:auditability}
\end{table*}

Table~\ref{tab:auditability} reports what can actually be inspected in a
fitted SRTS-BCE calibrator. At the full budget the three group temperatures
are tight across seeds (medians $0.63/0.73/0.84$, narrow IQRs) and
monotone in the routed risk index on every seed; the two router thresholds
are stable; no fit touches the $[0.05,20]$ temperature bounds; and no group
falls below the $50$-sample floor that triggers the global-scalar fallback.
Across all $220$ calibration draws at $n{=}250$ the picture degrades
without degenerating: saturation and fallback remain at
$0\%$, empirical risk ordering stays monotone in $93\%$ of fits, and the
noise shows up where expected --- the low-risk group's temperature becomes
volatile (median $0.17$, IQR $[0.13,0.73]$: an almost-always-correct group
supports aggressive sharpening) and strict temperature-ordering
monotonicity drops to $64\%$. Under shift the routed occupancy moves
interpretably: from roughly uniform thirds on clean test to a high-risk
share that grows monotonically with severity ($44\%$ at severity~1 to
$66\%$ at severity~5) while temperatures and thresholds stay fixed.

%==========================================================================
% F. Distribution shift
%==========================================================================

\section{Distribution Shift}
\label{app:shift}

\subsection{CIFAR-100-C corruption-shift evidence}
\label{app:cifar100c}

This subsection collects the full numerical evidence behind the
clean-to-corruption shift evaluation. The corruption sweep
covers the standard CIFAR-100-C release
\citep{hendrycks2019benchmarking}: 19 corruptions
$\times$ 5 severities $\times$ 3 seeds $= 285$ corruption cells per method.
Calibrators are fit only on the clean CIFAR-100 validation split
and applied unchanged to each (corruption, severity) cell; no
calibrator parameter is updated using corruption-test data.

% Requires: \usepackage{booktabs}.

\begin{table*}[t]
\centering
\small
\setlength{\tabcolsep}{1pt}
\begin{tabular}{lrrrrrr}
\toprule
Method & Top-1 & NLL & Brier & $\mathrm{ECE}_{15}$ & AURC & eAURC \\
\midrule
CE / NoTS & $78.09 {\pm} 0.76$ & $0.9450 {\pm} 0.0494$ & $0.3191 {\pm} 0.0122$ & $9.1334 {\pm} 0.5508$ & $0.0869 {\pm} 0.0100$ & $0.0487 {\pm} 0.0061$ \\
\midrule
\multicolumn{7}{l}{\textit{NLL objective (ablation)}} \\
TS-NLL & $78.09 {\pm} 0.76$ & $0.9094 {\pm} 0.0642$ & $0.3190 {\pm} 0.0153$ & $6.7545 {\pm} 0.9060$ & $0.0836 {\pm} 0.0094$ & $0.0454 {\pm} 0.0055$ \\
SRTS-NLL $K{=}3$ (ablation) & $78.09 {\pm} 0.76$ & $0.9068 {\pm} 0.0613$ & $0.3178 {\pm} 0.0143$ & $7.0772 {\pm} 0.9234$ & $0.0766 {\pm} 0.0066$ & $0.0384 {\pm} 0.0025$ \\
\midrule
\multicolumn{7}{l}{\textit{BCE objective}} \\
TvA-TS $\equiv$ TS-BCE $K{=}1$ & $78.09 {\pm} 0.76$ & $0.9004 {\pm} 0.0644$ & $0.3163 {\pm} 0.0156$ & $5.7158 {\pm} 1.1275$ & $0.0840 {\pm} 0.0094$ & $0.0457 {\pm} 0.0055$ \\
\textbf{SRTS-BCE $K{=}3$ (rerouted)} & $78.09 {\pm} 0.76$ & $\mathbf{0.8924 {\pm} 0.0581}$ & $\mathbf{0.3124 {\pm} 0.0140}$ & $\mathbf{4.6283 {\pm} 0.8410}$ & $\mathbf{0.0813 {\pm} 0.0084}$ & $\mathbf{0.0431 {\pm} 0.0045}$ \\
SMART+BCE & $78.09 {\pm} 0.76$ & $0.8950 {\pm} 0.0586$ & $0.3132 {\pm} 0.0142$ & $4.7479 {\pm} 0.7924$ & $0.0823 {\pm} 0.0085$ & $0.0441 {\pm} 0.0046$ \\
\bottomrule
\end{tabular}%
\caption{\textbf{CIFAR-100-C corruption-shift calibration under deployable routing} (3-seed mean $\pm$ standard deviation over seeds 0/1/2; each per-seed value is the mean over 95 corruption/severity cells). Calibrators are fit only on clean CIFAR-100 validation data and applied unchanged under shift. For SRTS-BCE, each corrupted sample is routed using its own corrupted logits through the clean-validation-fitted risk router; no clean-test routing group is reused. SRTS-NLL is included as an NLL-objective ablation. Among the headline calibrators in this table, SRTS-BCE has the lowest ECE under the deployable BCE protocol and retains a small eAURC advantage over SMART+BCE (the matched PWLinear-3 diagnostic attains a slightly lower point estimate, \S\ref{app:matched-capacity}); the SRTS--SMART ECE margin is small: cell-level and corruption-type bootstraps exclude zero, while the full seed$\times$corruption$\times$severity hierarchical bootstrap (\S\ref{app:best-per-criterion}) marginally includes zero; we treat the gain as directional rather than two-sided significant.}
\label{tab:cifar100c_summary}
\end{table*}

Table~\ref{tab:cifar100c_summary} reports the deployable
corruption evaluation. Aggregating the same 95 (corruption, severity) cells
per seed across three seeds, TvA-TS has $\mathrm{ECE}_{15} = 5.72$,
\textbf{SRTS-BCE $K{=}3$ deployable rerouting $4.63$}, and SMART+BCE
$4.75$. SRTS-BCE also has the lowest severity-4--5 ECE point estimate
($6.79$ vs.\ SMART+BCE $7.03$, TvA-TS $8.38$) and retains a small aggregate
eAURC advantage ($0.0431$ vs.\ SMART+BCE $0.0441$, TvA-TS $0.0457$).
The corrected ECE/eAURC trade-off referenced in the main text is
therefore: SRTS-BCE is competitive with SMART+BCE on clean ECE and,
under deployable rerouting, attains a slightly lower CIFAR-100-C mean
ECE while retaining the eAURC advantage. At the paired cell level the SRTS--SMART ECE margin excludes zero:
$\Delta=-0.1218$ pp with 95\% CI $[-0.1695,-0.0755]$. As in
\S\ref{app:best-per-criterion} and
Table~\ref{tab:cifar100c_summary}, the full hierarchical
bootstrap marginally includes zero, so the deployable CIFAR-100-C gain
is treated as directional rather than two-sided significant.

For completeness, the paired clean-group (non-deployable) protocol gives
SRTS-BCE $\mathrm{ECE}_{15}=5.7864 \pm 1.0246$ (per seed:
$4.9579$, $7.2301$, $5.1711$) and severity-4--5 ECE $8.7102$.
Paired clean-group reuse is possible only because CIFAR-100-C provides
corruptions paired with clean CIFAR-100 test images. It is not the
deployable protocol and is reported only as a diagnostic. The main text
uses deployable rerouting from corrupted logits.

\begin{figure}[t]
\centering
\includegraphics[width=0.95\columnwidth]{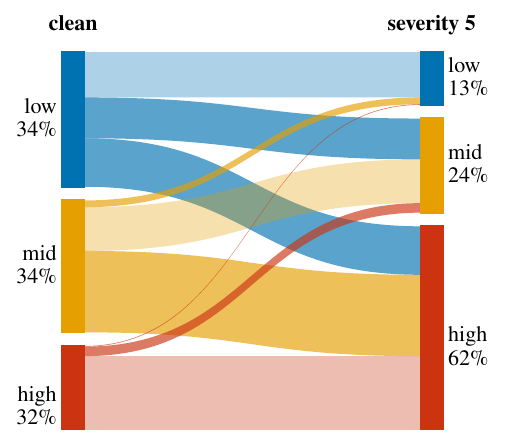}
\caption{\textbf{Corruption risk-flow diagnostic (clean $\to$ severity-5).}
Flow of samples between OOF risk tertiles from clean CIFAR-100 to severity-5
corruption (seed 0, averaged over 19 corruptions): the high-risk share grows
from ${\approx}32\%$ (clean) to ${\approx}62\%$ (severity 5). \emph{This
diagram uses the clean/corrupted image pairing to trace per-sample movement
and is therefore a mechanism diagnostic only; it is not used for any
deployable evaluation}, which reroutes each corrupted input from its own
logits without any clean pairing.}
\label{fig:corruption-risk-sankey}
\end{figure}

\paragraph{Test-side re-quantization ablation (why clean-anchored thresholds
matter).} A natural alternative to the deployed protocol is to
keep the clean-fitted temperatures but re-quantize the $K{=}3$ group
thresholds on each corruption cell's own risk-score distribution
(label-free, per-cell batch), restoring equal group occupancy under shift.
This \emph{hurts}, and monotonically more with severity: aggregate
$\mathrm{ECE}_{15}$ degrades from $4.63 \pm 0.84$ (deployed) to
$5.73 \pm 1.08$ (re-quantized), with the severity-5 mean rising from
$7.99$ to $10.30$ ($+2.31$ pp); eAURC also worsens slightly
($0.0431 \to 0.0434$). The absolute clean-anchored thresholds are therefore
doing the work under shift: letting shifted samples flood the high-risk group
(${\approx}62\%$ at severity 5; Figure~\ref{fig:corruption-risk-sankey})
\emph{is} the correct adaptation, because the absolute risk score carries
cross-distribution information that relative (per-batch quantile)
re-grouping destroys.

\begin{figure}[t]
\centering
\includegraphics[width=0.95\columnwidth]{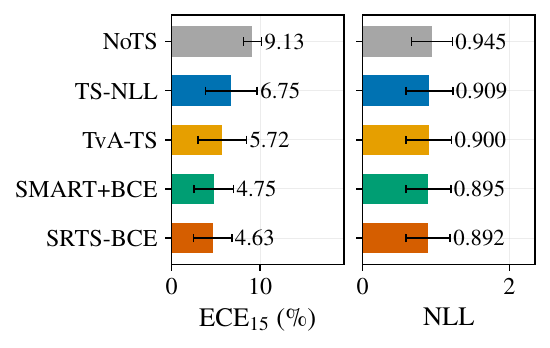}
\caption{\textbf{CIFAR-100-C 3-seed aggregate calibration metrics
(deployable BCE comparison).} Mean $\pm$ variability over seed$\times$severity
aggregates of the deployable corruption evaluation (each corrupted sample
rerouted from its own logits). The panel compares the methods of the final
deployable protocol: TvA-TS ($K{=}1$ BCE), SMART+BCE, and SRTS-BCE $K{=}3$
alongside NoTS/TS-NLL references.}
\label{fig:appendix-c100c-overall}
\end{figure}

\begin{figure}[t]
\centering
\includegraphics[width=0.95\columnwidth]{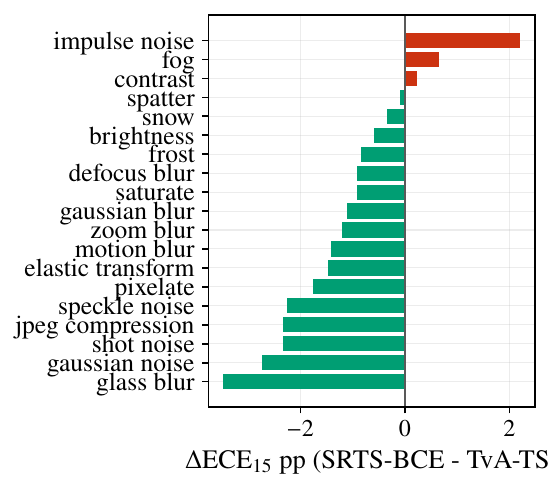}
\caption{\textbf{Per-corruption $\Delta\mathrm{ECE}_{15}$ of SRTS-BCE
$K{=}3$ minus TvA-TS on CIFAR-100-C.} Negative values indicate that deployable
risk routing improves over scalar BCE on that corruption; positive values
indicate a local regression. This view exposes the heterogeneous corruption
profile that is averaged in Table~\ref{tab:cifar100c_summary}.}
\label{fig:appendix-c100c-k3-minus-k1}
\end{figure}

\begin{figure}[t]
\centering
\includegraphics[width=0.95\columnwidth]{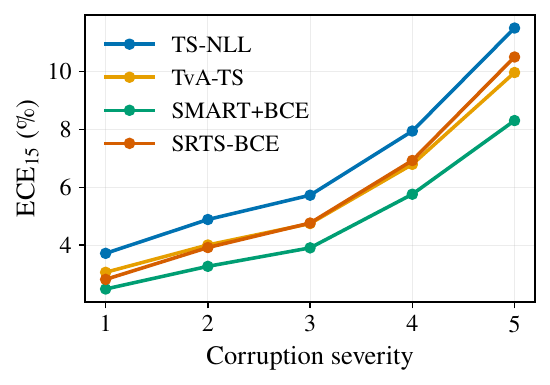}
\caption{\textbf{CIFAR-100-C $\mathrm{ECE}_{15}$ by severity
(deployable BCE comparison).} SRTS-BCE and SMART+BCE track below TvA-TS across
severity, while TS-NLL degrades fastest at high severity. These are the same
clean-validation-fitted calibrators applied unchanged to corrupted inputs, with
SRTS rerouting each corrupted sample from its own logits.}
\label{fig:appendix-c100c-sats-failure}
\end{figure}

\subsection{Deployment-facing selective metrics}
\label{app:deployment-metrics}

% from results/deployment_metrics/deployment_metrics.json. No hand-edited numbers.
\begin{table*}[t]
\centering
\small
\setlength{\tabcolsep}{3pt}
\begin{tabular}{lrrrrr}
\toprule
Method & HCFP@.90 & HCFP@.95 & Risk@80 & Risk@90 & Risk@95 \\
\midrule
\multicolumn{6}{@{}l}{\emph{Clean CIFAR-100}}\\
TvA-TS & $22.94$ & $12.00$ & $2.38$ & $4.59$ & $6.47$ \\
SMART+BCE & $15.41$ & $9.56$ & $2.40$ & $4.60$ & $6.49$ \\
\textbf{SRTS-BCE} & $15.37$ & $7.95$ & $2.42$ & $4.58$ & $6.45$ \\
\midrule
\multicolumn{6}{@{}l}{\emph{CIFAR-100-C (rerouting)}}\\
TvA-TS & $15.43$ & $7.97$ & $12.06$ & $16.65$ & $19.24$ \\
SMART+BCE & $10.69$ & $6.44$ & $12.08$ & $16.64$ & $19.24$ \\
\textbf{SRTS-BCE} & $10.06$ & $5.35$ & $12.07$ & $16.63$ & $19.24$ \\
\bottomrule
\end{tabular}
\caption{\textbf{Deployment-facing selective metrics} (\%; five-seed means
on both clean CIFAR-100 and CIFAR-100-C), from cached calibrated
probabilities. HCFP.$t$: fraction of wrong
predictions with confidence ${\geq}t$ (confidently wrong). R@$c$: selective
error at coverage $c$ (error among the most-confident $c$ fraction). Lower
is better throughout.}
\label{tab:deployment_metrics}
\end{table*}

From the cached calibrated probabilities we compute two deployment-facing
metric families (Table~\ref{tab:deployment_metrics}): HCFP@$t$, the fraction
of wrong predictions that are confidently wrong (confidence ${\geq}t$), and
selective risk at fixed coverage. This table uses all five training seeds on
both clean CIFAR-100 and CIFAR-100-C; the decision-threshold and
matched-coverage analyses below (Tables~\ref{tab:decision_thresholds}
and~\ref{tab:matched_coverage}) use the first three seeds on CIFAR-100-C,
the subset fixed when those protocols were specified. The two tell different stories. On both
clean CIFAR-100 and CIFAR-100-C, SRTS-BCE and SMART+BCE substantially reduce
the confidently-wrong rate relative to scalar TvA-TS (clean HCFP@$0.95$
$7.95$/$9.56$ vs.\ $12.00$; shift $5.35$/$6.44$ vs.\ $7.97$), SRTS-BCE
lowest. Selective risk at $80/90/95\%$ coverage, however, is empirically nearly
identical across the three methods (differences from TvA-TS at most $0.04$
pp on clean data and $0.03$ pp on CIFAR-100-C). Argmax preservation does
not imply preservation of the cross-sample confidence ranking: SRTS-BCE and
SMART+BCE apply different temperatures to different samples and can reorder
confidences across samples, and the lower AURC in
Table~\ref{tab:matched_coverage} reflects these small ranking changes
accumulated over the full risk--coverage curve. At the three reported
coverages the gain is a top-label calibration refinement --- fewer
confidently-wrong predictions not a reduction in selective
risk.

% results/decision_thresholds/decision.json. No hand-edits.
\begin{table*}[t]
\centering
\small
\setlength{\tabcolsep}{3.5pt}
\begin{tabular}{lccccccccc}
\toprule
 & \multicolumn{3}{c}{$t=0.80$} & \multicolumn{3}{c}{$t=0.90$} & \multicolumn{3}{c}{$t=0.95$} \\
\cmidrule(lr){2-4}\cmidrule(lr){5-7}\cmidrule(lr){8-10}
Method & Cov & Risk & Gap & Cov & Risk & Gap & Cov & Risk & Gap \\
\midrule
\multicolumn{10}{@{}l}{\emph{Clean CIFAR-100 (five seeds)}}\\
NoTS & $72.3$ & $1.97$ & $-8.86{\pm}0.47$ & $32.5$ & $1.09$ & $-6.15{\pm}0.41$ & $5.5$ & $2.02$ & $-1.94{\pm}1.76$ \\
TvA-TS & $87.6$ & $3.87$ & $+0.35{\pm}0.57$ & $81.0$ & $2.55$ & $-0.09{\pm}0.42$ & $70.8$ & $1.53$ & $-0.50{\pm}0.18$ \\
HTS-BCE & $87.2$ & $3.74$ & $+0.31{\pm}0.53$ & $80.4$ & $2.48$ & $-0.03{\pm}0.55$ & $70.3$ & $1.57$ & $-0.30{\pm}0.51$ \\
SMART+BCE & $84.4$ & $3.14$ & $+0.19{\pm}0.59$ & $76.0$ & $1.83$ & $+0.17{\pm}0.48$ & $68.1$ & $1.27$ & $+0.24{\pm}0.37$ \\
\textbf{SRTS-BCE} & $84.9$ & $3.23$ & $+0.17{\pm}0.49$ & $75.5$ & $1.84$ & $+0.16{\pm}0.39$ & $67.8$ & $1.06$ & $+0.04{\pm}0.21$ \\
\midrule
\multicolumn{10}{@{}l}{\emph{CIFAR-100-C, deployable rerouting (three seeds)}}\\
NoTS & $50.1$ & $4.20$ & $-6.90{\pm}0.45$ & $21.8$ & $2.41$ & $-4.74{\pm}0.45$ & $4.1$ & $2.81$ & $-1.13{\pm}1.19$ \\
TvA-TS & $70.8$ & $8.36$ & $+3.56{\pm}0.75$ & $60.2$ & $5.46$ & $+2.35{\pm}0.54$ & $48.3$ & $3.50$ & $+1.38{\pm}0.27$ \\
HTS-BCE & $69.8$ & $8.08$ & $+3.37{\pm}0.41$ & $59.3$ & $5.33$ & $+2.34{\pm}0.54$ & $47.9$ & $3.53$ & $+1.53{\pm}0.56$ \\
SMART+BCE & $65.5$ & $6.86$ & $+2.33{\pm}0.22$ & $53.3$ & $4.06$ & $+1.85{\pm}0.14$ & $44.5$ & $2.78$ & $+1.59{\pm}0.16$ \\
\textbf{SRTS-BCE} & $66.2$ & $7.02$ & $+2.51{\pm}0.24$ & $53.9$ & $4.18$ & $+1.95{\pm}0.24$ & $45.5$ & $2.64$ & $+1.39{\pm}0.11$ \\
\bottomrule
\end{tabular}
\caption{\textbf{Confidence-threshold deployment at pre-specified thresholds}
$t\in\{0.80,0.90,0.95\}$ (fixed before inspecting results; existing
predictions only). \emph{Cov}: retained coverage (\%). \emph{Risk}: error
among retained (\%). \emph{Gap}: mean retained confidence $-$ retained
accuracy (pp; positive $=$ overconfident), $\pm$ std over seeds (five clean,
three CIFAR-100-C).}
\label{tab:decision_thresholds}
\end{table*}

\noindent\textbf{Confidence-threshold deployment.} At the pre-specified
thresholds $t\in\{0.80,0.90,0.95\}$ (fixed before inspecting results;
Table~\ref{tab:decision_thresholds}), SRTS-BCE reduces the
nominal-vs-realized confidence gap relative to scalar TvA-TS at two of the
three thresholds on clean data ($|{+}0.17|$ vs.\ $|{+}0.35|$ at $0.80$;
$|{+}0.04|$ vs.\ $|{-}0.50|$ at $0.95$) and at two of three under
CIFAR-100-C ($+2.51$ vs.\ $+3.56$ at $0.80$; $+1.95$ vs.\ $+2.35$ at
$0.90$; tied at $0.95$), without worsening retained error --- at $t=0.95$
on clean data it also has the lowest retained error ($1.06\%$ at $67.8\%$
coverage). Its retained coverage is $3$--$5$ pp below the scalar's
(cooler temperatures retain fewer predictions), so the fixed-threshold
comparison confounds alignment with coverage; the matched-coverage control
below removes the confound. Throughout, this is framed as generic
threshold-based decision utility not abstention quality.

% results/decision_thresholds/matched_coverage.json. No hand-edits.
\begin{table*}[t]
\centering
\small
\setlength{\tabcolsep}{4pt}
\begin{tabular}{lccccccc}
\toprule
 & & \multicolumn{2}{c}{@cov$(t{=}0.80)$} & \multicolumn{2}{c}{@cov$(t{=}0.90)$} & \multicolumn{2}{c}{@cov$(t{=}0.95)$} \\
\cmidrule(lr){3-4}\cmidrule(lr){5-6}\cmidrule(lr){7-8}
Method & AURC & Risk & Gap & Risk & Gap & Risk & Gap \\
\midrule
\multicolumn{8}{@{}l}{\emph{Clean CIFAR-100 (five seeds)}}\\
NoTS & $20.73$ & $3.98$ & $-9.66$ & $2.74$ & $-9.41$ & $1.85$ & $-8.83$ \\
TvA-TS & $18.38$ & $3.87$ & $+0.35$ & $2.55$ & $-0.09$ & $1.53$ & $-0.50$ \\
HTS-BCE & $18.80$ & $3.87$ & $+0.35$ & $2.57$ & $-0.00$ & $1.60$ & $-0.30$ \\
SMART+BCE & $16.49$ & $3.89$ & $+0.18$ & $2.54$ & $+0.20$ & $1.44$ & $+0.21$ \\
\textbf{SRTS-BCE} & $15.49$ & $3.88$ & $+0.21$ & $2.55$ & $+0.17$ & $1.38$ & $+0.14$ \\
\midrule
\multicolumn{8}{@{}l}{\emph{CIFAR-100-C, deployable rerouting (three seeds)}}\\
NoTS & $68.51$ & $8.57$ & $-7.57$ & $5.81$ & $-7.36$ & $3.98$ & $-6.82$ \\
TvA-TS & $65.29$ & $8.36$ & $+3.56$ & $5.46$ & $+2.35$ & $3.50$ & $+1.38$ \\
HTS-BCE & $65.77$ & $8.39$ & $+3.42$ & $5.52$ & $+2.40$ & $3.58$ & $+1.55$ \\
SMART+BCE & $63.23$ & $8.43$ & $+2.54$ & $5.52$ & $+2.12$ & $3.30$ & $+1.68$ \\
\textbf{SRTS-BCE} & $62.24$ & $8.39$ & $+2.71$ & $5.49$ & $+2.22$ & $3.19$ & $+1.64$ \\
\bottomrule
\end{tabular}
\caption{\textbf{Matched-coverage control for the threshold comparison.}
Because different calibrators retain different fractions at a fixed
confidence threshold, every method is evaluated at the \emph{same}
reference coverages --- TvA-TS's coverage at the pre-specified thresholds
$\{0.80,0.90,0.95\}$ (no operating point chosen after inspecting results;
top-$k$ retained set with $k=\operatorname{round}(cN)$, the nearest
attainable coverage).
\emph{AURC}: area under the full risk--coverage curve ($\times 10^{3}$,
lower is better). \emph{Risk}/\emph{Gap}: retained error (\%) and
nominal-vs-realized confidence gap (pp) at the matched coverage.}
\label{tab:matched_coverage}
\end{table*}

\noindent\textbf{Matched-coverage control.} Evaluating every method at
the same reference coverages (TvA-TS's coverage at each pre-specified
threshold; Table~\ref{tab:matched_coverage}), the pattern survives coverage
matching: retained risk remains closely matched to TvA-TS --- a maximum
excess of $0.03$ pp across the six matched points on the un-rounded values
(no non-inferiority margin was pre-specified, so we report the raw maximum
we do not claim uniform non-inferiority) --- and is lower at the
$t{=}0.95$ reference coverage on both settings, the absolute
nominal-vs-realized gap remains smaller than the scalar's at two of the
three matched coverages on both clean and shifted data, and the
coverage-free AURC is lowest for SRTS-BCE on both settings (clean $15.49$
vs.\ $18.38$ scalar and $18.80$ HTS-BCE; CIFAR-100-C $62.24$ vs.\ $65.29$
and $65.77$, $\times 10^{3}$). For a target coverage $c$ over $N$ test
samples, samples are sorted by calibrated confidence (deterministic stable
sort) and retained risk and confidence gap are computed on the top-$k$
retained set with $k=\operatorname{round}(cN)$, i.e.\ at the nearest
attainable coverage $k/N$; no interpolation between adjacent coverage
points is applied.

\subsection{Where deployable rerouting helps, and why}
\label{app:routing-heterogeneity}

The aggregate CIFAR-100-C gain hides strong per-corruption structure that
sharpens the mechanism. We compute, for each of the $19$ corruptions, the
routing gain of SRTS-BCE over scalar TvA-TS (mean $\mathrm{ECE}_{15}$ over
five severities and three seeds). The gain is not uniform: it ranges from
$+3.47$ pp (glass blur) to $-2.20$ pp (impulse noise). Rerouting is
therefore a conditional benefit, not a fixed offset.

\begin{figure}[t]
\centering
\includegraphics[width=0.9\columnwidth]{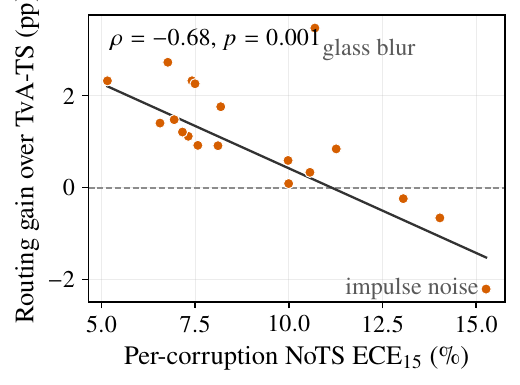}
\caption{\textbf{Per-corruption routing gain vs corruption severity
(CIFAR-100-C, three seeds).} Each point is one of the $19$ corruptions:
$x$ is its NoTS $\mathrm{ECE}_{15}$ (how badly the uncalibrated model is
miscalibrated under that corruption), $y$ is the SRTS-BCE routing gain over
scalar TvA-TS in $\mathrm{ECE}_{15}$ points. The clean-fitted router helps
on milder corruptions and degrades on the most severe ones
(Spearman $\rho = -0.68$, $p = 0.001$; Theil--Sen trend). Points above the
dashed line favour routing.}
\label{fig:appendix-routing-heterogeneity}
\end{figure}

\noindent\textbf{The gain tracks how far the corruption pushes the model
from the clean regime.} The per-corruption gain anti-correlates with the
per-corruption miscalibration magnitude, measured as NoTS
$\mathrm{ECE}_{15}$: Spearman $\rho = -0.68$
($p = 0.001$; Figure~\ref{fig:appendix-routing-heterogeneity}). The router
is fit once on clean validation logits; under a mild corruption the clean
risk ranking still transfers and routing recovers structured residual,
whereas under the most severe corruptions (NoTS $\mathrm{ECE}_{15} > 12$)
the logit distribution leaves the clean regime, the clean-fitted risk
signal degrades, and the group temperatures no longer match the shifted
residual. This is the corruption-axis image of the bias--variance boundary
in the main Method: risk-conditioned scaling helps while the risk signal remains
informative and washes out once it does not.

\noindent\textbf{The benefit is a property of the router, not the
baseline.} The per-corruption gain measured against TvA-TS and the gain
measured against SMART+BCE agree almost exactly (Spearman $\rho = 0.96$),
and SRTS-BCE has the lower $\mathrm{ECE}_{15}$ than \emph{both} peers on
$12$ of $19$ corruptions. The corruptions where it helps least are
the same regardless of which scalar-or-adaptive peer we compare to, so the
heterogeneity reflects transferability of the clean risk signal rather
than an artifact of one baseline.

\subsection{ConvNet CIFAR-100-C Deployable Rerouting}
\label{app:convnet-c100c}

We repeat the deployable shift evaluation on the two ConvNet scope-check
models (ResNet-50 and RegNetY-1.6GF; same frozen recipe, fit on clean
validation only, each corrupted sample rerouted from its own logits through
the clean-fitted router and clean-validation thresholds).
Table~\ref{tab:convnet_c100c} reports the results.

% ConvNet x CIFAR-100-C deployable rerouting (supplement).
% Source: results/convnet_c100c_audit/run.log (2 models x 3 seeds x 95 cells).
% Requires: \usepackage{booktabs}
\begin{table*}[t]
\centering
\small
\setlength{\tabcolsep}{5pt}
\begin{tabular}{lrr}
\toprule
Method & ResNet-50 & RegNetY-1.6GF \\
\midrule
NoTS        & $14.39{\pm}0.75$ & $15.24{\pm}0.82$ \\
TS-NLL      & $19.95{\pm}0.68$ & $12.77{\pm}0.62$ \\
TvA-TS      & $18.16{\pm}1.10$ & $11.60{\pm}0.84$ \\
SMART+BCE   & $14.02{\pm}1.16$ & $\mathbf{9.08{\pm}0.86}$ \\
SRTS-BCE $K{=}3$ & $\mathbf{13.72{\pm}0.79}$ & $9.44{\pm}0.78$ \\
\midrule
SRTS$-$SMART $\Delta$ (cell-level CI) & $-0.31$ $[-0.37,-0.25]$ & $+0.37$ $[+0.32,+0.41]$ \\
\bottomrule
\end{tabular}
\caption{\textbf{ConvNet CIFAR-100-C deployable rerouting}
($\mathrm{ECE}_{15}$, \%). Values are averaged over 95 corruption/severity
cells per seed and then reported as mean$\pm$std over three seeds; lower is
better. The SRTS$-$SMART interval is a paired cell-level bootstrap
conditional on the three trained seeds, not a seed-hierarchical interval.}
\label{tab:convnet_c100c}
\end{table*}

Deployable rerouting again improves clearly over scalar TvA-TS on both
ConvNets ($-4.44$ pp for ResNet-50 and $-2.16$ pp for RegNetY-1.6GF), but
the SRTS-BCE/SMART+BCE ordering is backbone-dependent, with
bootstrap-supported wins in opposite directions (Table~\ref{tab:convnet_c100c},
last row). We therefore do not claim that SRTS-BCE beats SMART+BCE across
ConvNets; the split refines the regime map.

On ResNet-50, clean-fitted scalar temperatures are harmful under corruption
shift: TS-NLL and TvA-TS are worse than NoTS. Routed methods reduce this
failure mode on average, but no fit-on-clean calibrator is uniformly safe
under all corruptions. This supports the regime-map framing.

The shift mechanism matches the ViT-family result: the high-risk group
share increases from approximately one third on clean data to $69\%$ on
ResNet-50 and $72\%$ on RegNetY at severity 5.

\subsection{Generated Tiny-C and CIFAR-10-C Collapse-Regime Diagnostics}
\label{app:tinyc-c10c}

We additionally evaluate two further corruption suites from cached logits,
reported as diagnostics only. The Tiny-ImageNet suite uses
\emph{generated ImageNet-C-style corruptions}
\citep{hendrycks2019benchmarking} of the clean Tiny validation
images (a per-seed generation cache), \emph{not} the official
Tiny-ImageNet-C release. \textbf{Under these generated corruptions all
methods are in a severe collapse regime, with $\mathrm{ECE}_{15}$ above
$60\%$ for every calibrator; this is a stress test of within-collapse
ordering only, not a main result.} Within that regime the ordering is consistent across all
three backbones (SRTS-BCE $<$ SMART+BCE $<$ TvA-TS $<$ TS-NLL), and the
cell-level SRTS$-$SMART deltas are bootstrap-supported:
ViT $-0.68$ $[-0.75,-0.62]$, DeiT-S $-0.51$ $[-0.57,-0.45]$, and
Swin-T $-0.54$ $[-0.59,-0.49]$.

On CIFAR-10-C (official release, ViT-B/16, three seeds), the outcome is a
tie: SRTS-BCE $3.00$ vs.\ SMART+BCE $3.20$ aggregate $\mathrm{ECE}_{15}$,
with the cell-level CI crossing zero ($[-0.55,\,+0.14]$).

\subsection{ImageNet-C, ImageNet-Sketch, and ImageNet-scale boundaries}
\label{app:imagenetc}

This subsection documents the second corruption-shift dataset: an
evaluation-only ImageNet-C sweep using the pretrained ViT-B/16
checkpoint (ImageNet-21k\,$\to$\,ImageNet-1k head; no fine-tuning). The sweep
covers 75 cells = the 15 main ImageNet-C corruptions
\citep{hendrycks2019benchmarking} $\times$
5 severities, $50{,}000$ images per cell.

\paragraph{Calibration protocol.} All four post-hoc methods are
fit \emph{once} on a deterministic $5{,}000$-sample
stratified-first-5-per-class calibration subset of the clean
ImageNet-1K validation set, and applied unchanged to each of the
75 corruption-test cells and to the remaining $45{,}000$-sample
held-out clean eval subset. No calibrator parameter is updated
using ImageNet-C data.

% Aggregated over 75 cells (15 corruptions × 5 severities). Mean ± std.
\begin{table*}[t]
\centering
\small
\setlength{\tabcolsep}{4pt}
\begin{tabular}{l c c c c c}
\toprule
Method & top-1 (\%) & NLL & ECE$_{15}$ (\%) & AdaECE$_{15}$ (\%) & AURC \\
\midrule
NoTS & 65.69 $\pm$ 12.15 & 1.504 $\pm$ 0.717 & 2.60 $\pm$ 1.79 & 2.59 $\pm$ 1.79 & 0.1356 $\pm$ 0.0858 \\
TS-NLL & 65.69 $\pm$ 12.15 & 1.503 $\pm$ 0.718 & 2.96 $\pm$ 1.89 & 2.94 $\pm$ 1.90 & 0.1355 $\pm$ 0.0858 \\
SRTS-NLL K=1 (ablation) & 65.69 $\pm$ 12.15 & 1.503 $\pm$ 0.718 & 2.96 $\pm$ 1.89 & 2.94 $\pm$ 1.90 & 0.1355 $\pm$ 0.0858 \\
SRTS-NLL K=3 (ablation) & 65.69 $\pm$ 12.15 & 1.503 $\pm$ 0.720 & 3.15 $\pm$ 2.26 & 3.14 $\pm$ 2.26 & 0.1358 $\pm$ 0.0858 \\
SATS-Logit & 65.69 $\pm$ 12.15 & 3.176 $\pm$ 1.644 & 18.47 $\pm$ 6.90 & 18.25 $\pm$ 6.94 & 0.1755 $\pm$ 0.0996 \\
\bottomrule
\end{tabular}
\caption{\textbf{ImageNet-C 75-cell calibration summary (NLL-objective diagnostic only)}
(15 canonical Hendrycks main corruptions $\times$ 5 severities, 
50{,}000 images per cell; single seed; pretrained ViT-B/16). All four 
calibrators are fit ONCE on the 5{,}000-sample stratified-first-5-per-class 
split of clean ImageNet-1K val and applied unchanged to each 
    (corruption, severity) cell (no refit on corruption test data). TS-NLL and SRTS-NLL $K{=}1$ are identical to 4 decimal
places on every cell. Selective-classification cost (AURC) is included to 
expose the SATS val-overfit signature, which can amplify under shift even 
when held-out ECE looks acceptable on clean data.}
\label{tab:imagenetc_summary}
\end{table*}

\paragraph{Three findings (single seed; NLL-objective rows; no
statistical-significance claim).} (i) The $K{=}1 \equiv$ scalar-TS identity
holds across all 75 IN-C cells under NLL. (ii) SATS-Logit val-overfit
amplifies under IN-C shift. (iii) Under the NLL objective, $K{=}3$ does
not win over $K{=}1$ on this base. The strong pretrained ImageNet-1K head
has clean $\mathrm{ECE}_{15} \approx 1.0\%$ and fitted scalar
$T \approx 0.98$; on this well-calibrated base the NLL-objective routing
gain disappears. \emph{The BCE-objective extension is reported in
\S\ref{app:inc-bce}: on the same pretrained base,
TvA-TS ($K{=}1$ BCE) achieves the lowest ECE ($2.47$) and SRTS-BCE /
SMART+BCE offer no improvement, confirming the routing benefit is
specific to fine-tuned small-validation settings.}

\begin{figure}[t]
\centering
\includegraphics[width=0.95\columnwidth]{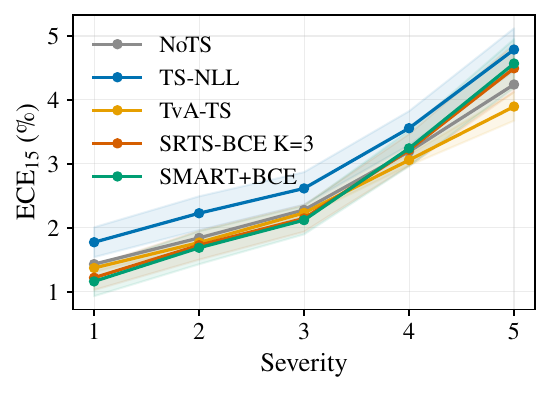}
\caption{\textbf{ImageNet-C $\mathrm{ECE}_{15}$ vs severity (mean over
15 corruptions; deployable BCE comparison).} On this strongly pretrained
base, the BCE-objective variants are tightly clustered and do not reproduce
the CIFAR-100-C routing gain, consistent with the text: TvA-TS remains a very
strong scalar baseline and SRTS-BCE/SMART+BCE offer no aggregate ECE
improvement on ImageNet-C.}
\label{fig:appendix-imagenetc-severity}
\end{figure}

\paragraph{Calibration-subset resampling.}

% Item 5 — ImageNet-C calibration-subset resampling summary (B=10 resamples).
% Eval-only on the pretrained ViT-B/16 ImageNet-1K logits artefacts.
% Each resample draws a stratified first-5-per-class subset of size 5000
% from the 50,000-sample IN-1K val pool; the same subset fits a calibrator
% which is then evaluated unchanged on each of the 75 IN-C cells (15
% corruptions x 5 severities). Numbers are mean ± standard deviation across
% the 10 resamples; per-resample value is the mean over the 75 cells.

\begin{table*}[t]
\centering
\small
\setlength{\tabcolsep}{4pt}
\begin{tabular}{lrrrr}
\toprule
Method & NLL & Brier & $\mathrm{ECE}_{15}$ & AURC \\
\midrule
NoTS                  & $1.504 \pm 0.000$ & $0.453 \pm 0.000$ & $2.599 \pm 0.000$ & $0.1356 \pm 0.0000$ \\
TS-NLL                & $1.504 \pm 0.001$ & $0.453 \pm 0.000$ & $3.169 \pm 0.260$ & $0.1355 \pm 0.0000$ \\
SRTS-NLL $K{=}1$ (ablation) & $1.504 \pm 0.001$ & $0.453 \pm 0.000$ & $3.169 \pm 0.260$ & $0.1355 \pm 0.0000$ \\
SRTS-NLL $K{=}3$ (ablation) & $1.504 \pm 0.001$ & $0.454 \pm 0.000$ & $3.414 \pm 0.320$ & $0.1357 \pm 0.0001$ \\
SATS-Logit            & $3.026 \pm 0.703$ & $0.529 \pm 0.016$ & $17.729 \pm 2.126$ & $0.1695 \pm 0.0079$ \\
\bottomrule
\end{tabular}%
\caption{\textbf{ImageNet-C calibration-subset resampling under the NLL objective on the pretrained ViT-B/16 ImageNet-1K base ($B = 10$ resamples)}. Reported values are the mean $\pm$ standard deviation across resamples of the per-resample 75-cell mean. This is an NLL-objective ablation: TS-NLL and SRTS-NLL $K{=}3$ do not improve $\mathrm{ECE}_{15}$ over NoTS on this well-calibrated base, and $K{=}3$ does not beat $K{=}1$. BCE-objective results are reported in \S\ref{app:inc-bce}.}
\label{tab:imagenetc_resampling}
\end{table*}

Table~\ref{tab:imagenetc_resampling} reports $B = 10$
resamples for the NLL-objective rows. Each resample draws a stratified
first-5-per-class subset ($5{,}000$ samples) from the $50{,}000$-sample
IN-1K validation pool, fits one calibrator, and evaluates it unchanged on
each of the $75$ IN-C cells. The resampling budget is small because each
resample refits the calibrators and re-evaluates the full $75$-cell sweep
at $50{,}000$ images per cell, so the resulting intervals are
correspondingly coarse and this sweep is treated as a diagnostic.
The resampling confirms the
ImageNet-C result: under NLL, $K{=}3$ does not beat $K{=}1$; SATS-Logit
val-overfit persists under resampling.

\paragraph{Large-scale OOD: ImageNet-Sketch.} As a second large-scale
probe we evaluate the pretrained ViT-B/16 on ImageNet-Sketch
($50{,}889$ hand-drawn-sketch images over the $1000$ ImageNet classes; a
severe distribution shift, top-1 $16.3\%$). Because ImageNet-1K images are
not available on our cluster to build a matched clean-validation split, we
use an \emph{in-domain} protocol: a stratified $2{,}500$-sample
Sketch validation subset fits each calibrator, and the remaining
$48{,}389$ images are the test set (three random splits; mean $\pm$ std).
This tests the low-capacity small-budget regime on a new large-scale
distribution not a clean-to-shift transfer. The uncalibrated model
is badly miscalibrated (NoTS $\mathrm{ECE}_{15} = 14.99$); post-hoc
calibration recovers it, and SRTS-BCE attains the lowest
$\mathrm{ECE}_{15}$ ($1.41 \pm 0.06$), narrowly ahead of TvA-TS
($1.44 \pm 0.03$) and well ahead of TS-NLL ($1.88 \pm 0.26$) and SMART+BCE
($2.21 \pm 0.24$). As on clean pretrained ImageNet, the routing gain over
scalar TvA-TS is small ($0.03$ pp): the pretrained head leaves little
structured residual for the router, consistent with the boundary of the
routing regime. On eAURC SMART+BCE is marginally best
($0.0945$ vs.\ SRTS-BCE $0.0983$).

\paragraph{ImageNet-scale \emph{fine-tuned} calibration: ImageNet-100,
three backbones.}
The probes above (ImageNet-C, Sketch) are the \emph{pretrained} regime; to
test the paper's \emph{fine-tuned} regime at ImageNet scale we fine-tune
ViT-B/16, DeiT-S, and Swin-T on ImageNet-100 (a $100$-class ImageNet subset;
$126{,}689$ training images; three seeds each, identical recipe), calibrate on
a ${\approx}2{,}500$-sample clean-validation split, and evaluate on held-out
clean test and on real ImageNet-C restricted to the $100$ classes ($15$
corruptions $\times 5$ severities, deployable rerouting).
Table~\ref{tab:in100} reports the outcome, which refines the boundary of the
regime map. On clean IN-100 the scalar
(TvA-TS), SMART+BCE and SRTS-BCE calibrators are statistically
indistinguishable: all land between $1.55$ and $2.00$ $\mathrm{ECE}_{15}$
and within seed variance of one another on every backbone (TvA-TS /
SMART+BCE / SRTS-BCE: ViT-B/16 $1.58$/$1.55$/$1.64$, DeiT-S
$1.83$/$1.74$/$2.00$, Swin-T $1.70$/$1.67$/$1.72$). SMART+BCE is stable
across seeds and competitive, not unstable. Risk conditioning therefore
yields no routing advantage at this fine-tuned-but-large scale: the
calibration residual is again largely a single scale factor, so
low-capacity routing adds nothing here, and the ordering among these
sub-$2$-pp calibrators is backbone-dependent and within seed variance ---
an ImageNet-scale boundary of the regime map, not a contradiction of it.
We report clean IN-100 here; corruption-shift evidence is drawn from
CIFAR-100-C and pretrained ImageNet-C.

% ImageNet-100 fine-tuned calibration, THREE backbones (ViT-B/16, DeiT-S,
% Swin-T). Clean = held-out test half of the 5000-img IN-100 val (25/class
% cal fit, stratified_half rng0). ALL methods regenerated with the canonical
% post-hoc pipeline (fit_all / fit_srts_stage1), replacing an earlier
% non-canonical evaluator whose SRTS and SMART+BCE values were artifacts
% (diagnosis retained in the repository provenance audit only).
% Seeds: DeiT-S 0-4 (5), ViT-B/16 & Swin-T 0-2 (3).
% Provenance: results/matched_capacity/20260713_122721/in100_canonical_scope.json.
% Clean IN-100 only; IN-100-C corruption shift is not reported in this paper.
\begin{table*}[t]
\centering
\small
\setlength{\tabcolsep}{8pt}
\begin{tabular}{lrrr}
\toprule
Method & ViT-B/16 & DeiT-S & Swin-T \\
\midrule
NoTS      & $3.97{\pm}0.24$ & $4.40{\pm}0.24$ & $3.14{\pm}0.25$ \\
TS-NLL    & $1.63{\pm}0.18$ & $2.58{\pm}0.21$ & $1.91{\pm}0.23$ \\
TvA-TS    & $1.58{\pm}0.24$ & $1.83{\pm}0.16$ & $1.70{\pm}0.28$ \\
SMART+BCE & $1.55{\pm}0.18$ & $1.74{\pm}0.20$ & $1.67{\pm}0.23$ \\
SRTS-BCE  & $1.64{\pm}0.27$ & $2.00{\pm}0.29$ & $1.72{\pm}0.20$ \\
\bottomrule
\end{tabular}
\caption{\textbf{ImageNet-100 fine-tuned calibration, three backbones}
($\mathrm{ECE}_{15}$ \%, mean $\pm$ std; DeiT-S over five seeds,
ViT-B/16 and Swin-T over three; each calibrator fit on a
${\approx}2{,}500$-sample clean-validation split and evaluated on the
held-out IN-100 test half, all with the canonical post-hoc pipeline). An
ImageNet-scale \emph{boundary diagnostic}: on clean IN-100 the scalar,
SMART+BCE and SRTS-BCE calibrators land within ${\approx}0.5$ pp of one
another ($1.5$--$2.0$), so risk conditioning yields no routing advantage
over scalar TvA-TS at this fine-tuned-but-large scale, and SMART+BCE is
stable and competitive across seeds. Ordering among the sub-$2$-pp
calibrators is backbone-dependent and within seed variance. This table
reports clean IN-100; corruption shift is covered on CIFAR-100-C and
ImageNet-C.}
\label{tab:in100}
\end{table*}

\paragraph{Pretrained ImageNet-1K clean calibration (four backbones).}
Complementing the fine-tuned IN-100 result, Table~\ref{tab:in1k_pretrained_clean}
calibrates four off-the-shelf ImageNet-1K backbones (no fine-tuning) on the
$50{,}000$-image validation set with the canonical pipeline; all reported
values use this pipeline, and provenance checks are included with the
released artifacts. On
these well-trained heads the calibration residual is small but not zero:
the low-capacity BCE calibrators give a \emph{small} clean improvement
over scalar TvA-TS on three of four backbones (SMART+BCE lowest on
ResNet-50 and ViT-B/16, SRTS-BCE lowest on Swin-B; TvA-TS best only on
the already well-calibrated EfficientNet-B0). Even a pretrained head
carries a modest structured residual that conditioning can exploit
in-distribution; what does not transfer is the \emph{shift} case: the clean-fitted router
gives no gain under ImageNet-C (\S\ref{app:inc-bce}, TvA-TS best), so
the pretrained boundary in the regime map is specific to distribution
shift, not to clean in-distribution calibration.

% Pretrained ImageNet-1K clean calibration (eval-only, no fine-tuning).
% Off-the-shelf timm backbones on the 50k IN-1K val: 5/class (5000) fits each
% calibrator, remaining 45000 are test. CANONICAL pipeline (fit_t_bce /
% fit_srts_stage1 / train_smart_bce), matching the rest of the paper; the
% earlier calibration_suite.run_all_methods version was a pipeline artifact
% (spurious SMART divergence, inflated SRTS). Provenance:
% results/matched_capacity/20260713_122721/in1k_canonical.json.
\begin{table*}[t]
\centering
\small
\setlength{\tabcolsep}{4pt}
\begin{tabular}{lrrrrrr}
\toprule
Backbone & top-1 & NoTS & TS-NLL & TvA-TS & SMART+BCE & SRTS-BCE \\
\midrule
ResNet-50        & $77.0$ & $14.65$ & $3.83$ & $2.02$ & $\mathbf{1.30}$ & $1.34$ \\
EfficientNet-B0  & $73.4$ & $6.82$ & $2.54$ & $\mathbf{0.93}$ & $0.97$ & $1.00$ \\
Swin-B           & $83.4$ & $49.14$ & $2.45$ & $1.32$ & $1.03$ & $\mathbf{0.91}$ \\
ViT-B/16         & $84.4$ & $1.02$ & $1.17$ & $0.96$ & $\mathbf{0.47}$ & $0.62$ \\
\bottomrule
\end{tabular}
\caption{\textbf{Pretrained ImageNet-1K clean calibration} ($\mathrm{ECE}_{15}$
\%, eval-only off-the-shelf backbones; $5$/class fits each calibrator, remaining
$45{,}000$ val images are test; canonical pipeline). On these
well-trained ImageNet heads the calibration residual is small
(NoTS $\mathrm{ECE}_{15}$ from $1.0$ to $49$; the large Swin-B value
reflects systematic \emph{under}-confidence --- mean confidence $34\%$
against top-1 $83\%$, so the $49$-point gap is the confidence--accuracy
deficit, which a single temperature removes, TS-NLL $2.45$), and the
low-capacity BCE calibrators give a
\emph{small} clean improvement over scalar TvA-TS on three of four
backbones --- consistent with a modest structured residual even on
pretrained heads. This clean gain does \emph{not} transfer to
ImageNet-C shift (\S\ref{app:inc-bce}), where scalar TvA-TS is best: the
pretrained boundary is specific to distribution shift, not to clean
in-distribution calibration. \textbf{Bold}: best deployable calibrator per row.}
\label{tab:in1k_pretrained_clean}
\end{table*}

\subsection{BCE-objective ImageNet-C evaluation}
\label{app:inc-bce}

% BCE-objective ImageNet-C evaluation.
% 10 calibration-subset resamples (5000 samples, 5/class from IN-1K val+test pool).
% Applied unchanged to 75 INC cells (15 corruptions x 5 severities).
% Single pretrained ViT-B/16. Mean +/- std across resamples.
\begin{table*}[t]
\centering
\small
\setlength{\tabcolsep}{4pt}
\begin{tabular}{lrrrr}
\toprule
Method & ECE$_{15}$ & eAURC & NLL & Sev.\ 4--5 ECE \\
\midrule
NoTS             & $2.60{\pm}0.00$ & $0.0565{\pm}0.0000$ & --- & $3.72{\pm}0.00$ \\
TS-NLL           & $2.99{\pm}0.26$ & $0.0565{\pm}0.0013$ & --- & $4.17{\pm}0.36$ \\
TvA-TS ($K{=}1$) & $\mathbf{2.47}{\pm}0.08$ & $0.0566{\pm}0.0005$ & --- & $\mathbf{3.48}{\pm}0.12$ \\
SRTS-BCE $K{=}3$ & $2.56{\pm}0.23$ & $0.0568{\pm}0.0027$ & --- & $3.86{\pm}0.34$ \\
SMART+BCE        & $2.56{\pm}0.26$ & $0.0572{\pm}0.0028$ & --- & $3.91{\pm}0.40$ \\
\bottomrule
\end{tabular}
\\[2pt]
{\small\raggedright\noindent Single pretrained ViT-B/16 (no fine-tuning; off-the-shelf pretrained checkpoint).
Calibration fitted on clean IN-1K only; no refit on IN-C data.
NLL omitted for conciseness (available with the released artefacts).
The BCE routing benefit that drives the CIFAR-100 gains does not transfer
to this strongly pretrained regime.\par}
\caption{\textbf{BCE-objective ImageNet-C evaluation.}
Calibrators fitted on a 5,000-sample clean ImageNet-1K calibration
subset (5 per class; $B{=}10$ resamples), applied unchanged to 75
ImageNet-C cells (15 corruptions $\times$ 5 severities).
Values are mean $\pm$ std across resamples.
On this pretrained backbone setting, TvA-TS ($K{=}1$ BCE) achieves the
lowest ECE$_{15}$; SRTS-BCE and SMART+BCE offer no improvement over
TvA-TS and are comparable to NoTS on ECE.
TS-NLL worsens ECE relative to NoTS, indicating the pretrained model is
already well-calibrated under NLL and a scalar NLL temperature slightly
miscalibrates it further. The BCE routing benefit observed on fine-tuned
CIFAR-100 is absent on ImageNet-C; see the accompanying discussion.}
\label{tab:inc_bce}
\end{table*}

Table~\ref{tab:inc_bce} reports our BCE-objective ImageNet-C
evaluation ($B{=}10$ calibration-subset resamples, 5,000-sample
IN-1K cal subset, 75 cells).

The key finding is qualitatively different from the CIFAR-100 result:
on the pretrained IN-1K ViT-B/16, TvA-TS ($K{=}1$ BCE) achieves the
lowest ECE ($2.47 \pm 0.08$) and lowest severity-4--5 ECE ($3.48 \pm 0.12$);
SRTS-BCE and SMART+BCE are comparable to each other ($2.56 \pm 0.23$ and
$2.56 \pm 0.26$) but offer no improvement over TvA-TS. TS-NLL worsens ECE
relative to NoTS ($2.99$ vs.\ $2.60$), indicating the pretrained model is
already near-calibrated under NLL and a scalar NLL temperature
miscalibrates it.

The contrast with fine-tuned CIFAR-100 (where SRTS-BCE reduces ECE
from $1.65$ to $0.96$ relative to TvA-TS) is that here, \emph{under
shift}, the clean-fitted router transfers no benefit: on the pretrained
IN-1K ViT-B/16 the ImageNet-C ordering is TvA-TS best, SRTS-BCE and
SMART+BCE offering no improvement. This is a shift-transfer boundary,
not a claim that routing is useless on pretrained heads --- on clean
in-distribution IN-1K the same low-capacity BCE calibrators do give a
small improvement over TvA-TS (Table~\ref{tab:in1k_pretrained_clean}).
The regime map's pretrained boundary is therefore specifically about
distribution shift: TvA-TS is the best-performing BCE-objective
calibrator on pretrained ImageNet heads \emph{under corruption}.

\subsection{Natural-shift boundary: ImageNet-R and ImageNet-A}
\label{app:natural-shift}

% Compact distribution-shift scope (main RQ3). Numbers are from the C100-C
% deployable table, IN-R/A (Supp natural-shift), and IN-V2 (Supp).
\begin{table}[t]
\centering
\small
\setlength{\tabcolsep}{3pt}
\begin{tabular}{@{}p{0.31\columnwidth}p{0.17\columnwidth}p{0.44\columnwidth}@{}}
\toprule
Shift & Best & Outcome \\
\midrule
CIFAR-100-C (synthetic, fine-tuned) & SRTS-BCE & $4.63$ vs.\ SMART+BCE $4.75$; matched PWLinear-3 $4.58$ lower, unresolved; hier.\ CI incl.\ 0 \\
\addlinespace
ImageNet-R/-A (natural, pretrained) & NoTS & every clean-fitted calibrator worsens ECE and TopBCE \\
\addlinespace
ImageNet-V2 (natural, fine-tuned) & BCE family & objective transfers (CIs excl.\ 0); routing vs.\ scalar unresolved \\
\bottomrule
\end{tabular}
\caption{\textbf{Distribution-shift scope} ($\mathrm{ECE}_{15}$; best
headline calibrator per regime). The pretrained-vs-fine-tuned split on
natural shift is the boundary: calibration transfers under fine-tuning,
not on an already-calibrated pretrained head.}
\label{tab:shift_scope}
\end{table}

% Natural-shift boundary experiment (ImageNet-R / ImageNet-A), pretrained
% ViT-B/16, subset-restricted deployable protocol; three calibration draws.
% Frozen run: results/natural_shift/20260713_203634 (comment only; not rendered).
\begin{table*}[t]
\centering
\small
\setlength{\tabcolsep}{8pt}
\begin{tabular}{lrrrr}
\toprule
& \multicolumn{2}{c}{ImageNet-R} & \multicolumn{2}{c}{ImageNet-A} \\
\cmidrule(lr){2-3}\cmidrule(lr){4-5}
Method & $\mathrm{ECE}_{15}$ & TopBCE & $\mathrm{ECE}_{15}$ & TopBCE \\
\midrule
NoTS & $\mathbf{5.57{\pm}0.00}$ & $\mathbf{0.395{\pm}0.000}$ & $\mathbf{22.89{\pm}0.00}$ & $\mathbf{0.712{\pm}0.000}$ \\
TS-NLL & $10.23{\pm}0.46$ & $0.430{\pm}0.005$ & $28.09{\pm}1.05$ & $0.818{\pm}0.024$ \\
TvA-TS & $8.55{\pm}0.38$ & $0.415{\pm}0.003$ & $26.32{\pm}0.89$ & $0.779{\pm}0.018$ \\
QaTS-NLL & $9.91{\pm}0.02$ & $0.428{\pm}0.002$ & $28.09{\pm}1.05$ & $0.818{\pm}0.024$ \\
QaTS-BCE & $7.05{\pm}2.03$ & $0.410{\pm}0.006$ & $26.32{\pm}0.89$ & $0.779{\pm}0.018$ \\
SMART+BCE & $9.02{\pm}3.81$ & $0.464{\pm}0.012$ & $27.44{\pm}3.28$ & $0.906{\pm}0.016$ \\
PWLinear-3 & $8.86{\pm}2.49$ & $0.439{\pm}0.008$ & $29.45{\pm}2.38$ & $0.841{\pm}0.045$ \\
SRTS-BCE & $9.21{\pm}1.10$ & $0.433{\pm}0.009$ & $27.71{\pm}3.65$ & $0.874{\pm}0.042$ \\
\bottomrule
\end{tabular}
\caption{\textbf{Natural-shift boundary: clean-ImageNet-fit calibration on
ImageNet-R and ImageNet-A} (pretrained ViT-B/16; label space restricted to
each variant's 200 classes; calibrators fit on a clean five-per-class
validation draw, ${\approx}1{,}000$ samples, applied unchanged to the
shifted test set; mean${\pm}$std over three calibration draws). Every
clean-fitted calibrator \emph{worsens} both metrics relative to the
uncalibrated model. \textbf{Bold}: best per column.}
\label{tab:natural_shift}
\end{table*}

We close the shift evidence with two \emph{natural} distribution shifts
on the same pretrained ViT-B/16: ImageNet-R ($30{,}000$ renditions,
$200$ classes) and ImageNet-A ($7{,}500$ natural adversarial examples,
$200$ classes). For each variant the label space is restricted to its
class subset (logit columns masked before any fitting or evaluation);
calibrators are fit on a clean ImageNet validation draw restricted to
the same subset (five images per class, ${\approx}1{,}000$ samples ---
inside the paper's small-budget regime) and applied unchanged to the
shifted test set. We report three calibration draws (one deterministic
five-per-class split and two stratified random draws) and paired
test-image bootstraps ($B{=}20{,}000$).

Table~\ref{tab:natural_shift} shows the strongest form of the
pretrained-boundary result. (i)~\emph{Every} clean-fitted calibrator
worsens both $\mathrm{ECE}_{15}$ and the proper score relative to the
uncalibrated model, on both variants: the clean draw fits a sharpening
temperature ($T \approx 0.92$--$0.96 < 1$) because the pretrained model
is slightly under-confident on clean data, while under natural shift the
same model is over-confident, so the clean-fitted temperature transfers
with the wrong sign. (ii)~Among the calibrators, scalar TvA-TS has the
lower mean ECE among the BCE calibrators on both variants (paired
$\Delta = $ SRTS-BCE $-$ TvA-TS: $+0.66$ $[+0.65,+0.66]$ on ImageNet-R,
$+1.40$ $[+1.37,+1.42]$ on ImageNet-A). These conditional test-image
bootstraps exclude zero but do not model calibration-draw variability;
Table~\ref{tab:natural_shift} therefore reports the across-draw
dispersion as the primary uncertainty view.
(iii)~The conditioned maps are additionally unstable across
calibration draws at this budget (e.g.\ SMART+BCE std $3.8$ ECE points
on ImageNet-R; on some draws the low-risk group temperatures collapse to
${\approx}0.2$). (iv)~$\mathrm{ECE}_{15}$ and AdaECE$_{15}$ coincide
exactly on these test sets because the shifted model is over-confident
in every confidence bin, so any binning reduces to the mean
confidence--accuracy gap. The scope-map entry for this regime is
therefore that no clean-fitted calibrator in the comparison improves on
the uncalibrated model, and scalar TvA-TS is the least harmful among
them; QaTS-BCE partially escapes on ImageNet-R (its rank-based quantile
map is flatter and sharpens less) but not on ImageNet-A, and remains
worse than not calibrating.

\paragraph{The positive regime transfers: ImageNet-V2 on the fine-tuned
IN-100 models.} The reversal above is a property of the pretrained
model, not of natural shift itself. Evaluating the nine fine-tuned
ImageNet-100 models (ViT-B/16, DeiT-S, Swin-T $\times$ three seeds) on
ImageNet-V2 restricted to the same $100$ classes ($10$ images per
class), with every calibrator fit only on the clean IN-100 calibration
half and applied unchanged, calibration clearly helps: NoTS
$\mathrm{ECE}_{15}$ $8.4$--$10.1$ falls to $2.9$--$3.9$ for the
BCE-objective family. To separate the objective effect from the
capacity effect we hold capacity fixed on each axis. At fixed scalar
capacity, the BCE objective improves over NLL on all three backbones
(TvA-TS $-$ TS-NLL: $-0.71$ $[-1.30,-0.10]$ ViT-B/16, $-1.52$
$[-2.13,-0.85]$ DeiT-S, $-0.74$ $[-1.33,-0.11]$ Swin-T; every CI
excludes zero), and at fixed routed capacity it likewise improves
(SRTS-BCE $-$ SRTS-NLL: $-0.95$ $[-1.53,-0.34]$, $-1.57$ $[-2.24,-0.82]$,
$-0.68$ $[-1.27,-0.06]$; every CI excludes zero). The capacity effect
is separate and unresolved: SRTS-BCE $-$ TvA-TS is $-0.27$
$[-0.76,+0.23]$ (ViT-B/16), $-0.15$ $[-0.82,+0.55]$ (DeiT-S), $+0.20$
$[-0.29,+0.68]$ (Swin-T), all crossing zero at this test size ($1{,}000$
images). ImageNet-V2 therefore supports the objective axis but not an
independent routing advantage: objective alignment transfers to natural
shift, the additional value of routing remains unresolved.

%==========================================================================
% G. Cross-backbone scope
%==========================================================================

\section{Cross-Backbone Scope}
\label{app:scope}

Table~\ref{tab:regime_winloss} summarises the per-regime outcome
across every setting in the paper; the subsections give the
per-dataset evidence.

\begin{table}[!htbp]
\centering
\small
\setlength{\tabcolsep}{3pt}
\begin{tabular}{@{}>{\raggedright\arraybackslash}p{0.36\columnwidth}%
>{\raggedright\arraybackslash}p{0.25\columnwidth}%
>{\raggedright\arraybackslash}p{0.30\columnwidth}@{}}
\toprule
Regime & Best method & Support \\
\midrule
Clean C100, 3 ViT backbones & SRTS-BCE 3/3 & ViT cell bootstrap-tied \\
Clean C100, 2 ConvNets   & split 1/1    & gaps $\leq 0.13$ pp \\
Clean CIFAR-10, 5 backbones & SMART-SoftECE on $\mathrm{ECE}_{15}$ & near-saturated; SRTS-BCE best AdaECE$_{15}$ 5/5 \\
Clean Tiny-IN, 3 backbones  & SMART+BCE 2/3 & point estimates \\
Clean IN-100, 3 fine-tuned backbones & tie (all methods) & no routing advantage; within seed variance ($1.5$--$2.0$) \\
\addlinespace
C100-C, deployable rerouting & SRTS-BCE  & directional; hierarch.\ CI incl.\ 0 \\
C100-C, 2 ConvNets       & split 1/1    & routed $\gg$ scalar; scalar $<$ NoTS (ResNet) \\
C100-C, PWLinear-3 vs.\ SRTS-BCE & neither (hierarchical CI incl.\ 0) & point favours PWLinear-3; cell/corr.-type tiers exclude 0 \\
\addlinespace
ImageNet-C, pretrained ViT & TvA-TS     & $B{=}10$ resamples \\
ImageNet-Sketch, pretrained ViT & SRTS-BCE $\approx$ TvA-TS & in-domain small val; tie \\
IN-R / IN-A (pretrained) & NoTS & clean-fitted calibrators all worsen ECE \\
IN-V2 (fine-tuned IN-100) & BCE family & objective axis transfers; conditioned vs.\ scalar unresolved \\
Val.\ budget $\leq 10\%$ & SRTS-BCE abs.\ / TvA-TS stab. & val-size sweep \\
Unrestricted heads, any regime & never best & ECE $\geq 6$ throughout \\
\bottomrule
\end{tabular}
\caption{\textbf{Explicit per-regime win/loss map} (primary metric
$\mathrm{ECE}_{15}$; the flagship CIFAR-100/ViT-B/16 cell uses five seeds,
other multi-run regimes three unless stated otherwise).
``Support'' qualifies the winning margin (row groups: clean, shift,
boundary). Where the honest outcome is a tie or a loss, the map reports it
as such; per-dataset evidence is in the subsections below.}
\label{tab:regime_winloss}
\end{table}

\subsection{Backbone generalisation: DeiT-S and Swin-T}
\label{app:backbone-generalisation}

This subsection extends the post-hoc calibration evaluation to
DeiT-S/16 and Swin-T under the same modern fine-tuning recipe as the
ID clean evaluation. Three independent seeds are reported for each backbone.

\begin{table*}[t]
\centering
\small
\setlength{\tabcolsep}{3pt}
\begin{tabular}{l l r r r r r r}
\toprule
Backbone & Method & Top-1 & NLL & Brier & $\mathrm{ECE}_{15}$ & AURC & eAURC \\
\midrule
\multirow{11}{*}{DeiT-S}
& NoTS & $88.76 {\pm} 0.63$ & $0.523 {\pm} 0.005$ & $0.178 {\pm} 0.008$ & $8.53 {\pm} 1.11$ & $0.0328 {\pm} 0.0022$ & $0.0262 {\pm} 0.0029$ \\
& TS-NLL & $88.76 {\pm} 0.63$ & $0.458 {\pm} 0.020$ & $0.174 {\pm} 0.006$ & $2.90 {\pm} 0.19$ & $0.0296 {\pm} 0.0026$ & $0.0230 {\pm} 0.0033$ \\
& SRTS-NLL $K{=}1$ (ablation) & $88.76 {\pm} 0.63$ & $0.458 {\pm} 0.020$ & $0.174 {\pm} 0.006$ & $2.90 {\pm} 0.19$ & $0.0296 {\pm} 0.0026$ & $0.0230 {\pm} 0.0033$ \\
& SRTS-NLL $K{=}3$ (ablation) & $88.76 {\pm} 0.63$ & $0.447 {\pm} 0.017$ & $0.170 {\pm} 0.007$ & $2.65 {\pm} 0.12$ & $0.0221 {\pm} 0.0004$ & $0.0156 {\pm} 0.0011$ \\
& SATS-Logit & $88.76 {\pm} 0.63$ & $1.116 {\pm} 0.342$ & $0.191 {\pm} 0.014$ & $7.02 {\pm} 1.05$ & $0.0393 {\pm} 0.0031$ & $0.0327 {\pm} 0.0023$ \\
& LTS-feature & $88.76 {\pm} 0.63$ & $2.048 {\pm} 0.262$ & $0.199 {\pm} 0.012$ & $8.30 {\pm} 0.65$ & $0.0449 {\pm} 0.0034$ & $0.0383 {\pm} 0.0027$ \\
& Feature-SATS & $88.76 {\pm} 0.63$ & $2.058 {\pm} 0.265$ & $0.199 {\pm} 0.012$ & $8.31 {\pm} 0.65$ & $0.0450 {\pm} 0.0035$ & $0.0384 {\pm} 0.0027$ \\
& Frozen-feature UTempH & $88.76 {\pm} 0.63$ & $2.058 {\pm} 0.265$ & $0.199 {\pm} 0.012$ & $8.31 {\pm} 0.65$ & $0.0450 {\pm} 0.0035$ & $0.0384 {\pm} 0.0027$ \\
& TvA-TS $\equiv$ TS-BCE & $88.76 {\pm} 0.63$ & $0.459 {\pm} 0.016$ & $0.173 {\pm} 0.005$ & $3.37 {\pm} 0.40$ & $0.0301 {\pm} 0.0021$ & $0.0235 {\pm} 0.0027$ \\
& \textbf{SRTS-BCE $K{=}3$ risk} & $88.76 {\pm} 0.63$ & $0.451 {\pm} 0.013$ & $0.168 {\pm} 0.005$ & $\mathbf{0.94 {\pm} 0.07}$ & $0.0220 {\pm} 0.0003$ & $0.0154 {\pm} 0.0007$ \\
& SMART+BCE & $88.76 {\pm} 0.63$ & $0.455 {\pm} 0.014$ & $0.169 {\pm} 0.006$ & $1.17 {\pm} 0.01$ & $0.0218 {\pm} 0.0001$ & $0.0152 {\pm} 0.0005$ \\
\midrule
\multirow{11}{*}{Swin-T}
& NoTS & $88.75 {\pm} 0.48$ & $0.514 {\pm} 0.008$ & $0.177 {\pm} 0.004$ & $8.77 {\pm} 0.53$ & $0.0308 {\pm} 0.0006$ & $0.0242 {\pm} 0.0010$ \\
& TS-NLL & $88.75 {\pm} 0.48$ & $0.443 {\pm} 0.007$ & $0.171 {\pm} 0.004$ & $2.98 {\pm} 0.24$ & $0.0270 {\pm} 0.0006$ & $0.0205 {\pm} 0.0012$ \\
& SRTS-NLL $K{=}1$ (ablation) & $88.75 {\pm} 0.48$ & $0.443 {\pm} 0.007$ & $0.171 {\pm} 0.004$ & $2.98 {\pm} 0.24$ & $0.0270 {\pm} 0.0006$ & $0.0205 {\pm} 0.0012$ \\
& SRTS-NLL $K{=}3$ (ablation) & $88.75 {\pm} 0.48$ & $0.435 {\pm} 0.008$ & $0.169 {\pm} 0.004$ & $2.76 {\pm} 0.36$ & $0.0215 {\pm} 0.0003$ & $0.0149 {\pm} 0.0009$ \\
& SATS-Logit & $88.75 {\pm} 0.48$ & $1.319 {\pm} 0.371$ & $0.192 {\pm} 0.007$ & $7.24 {\pm} 0.49$ & $0.0404 {\pm} 0.0036$ & $0.0338 {\pm} 0.0031$ \\
& LTS-feature & $88.75 {\pm} 0.48$ & $3.895 {\pm} 1.527$ & $0.201 {\pm} 0.007$ & $8.44 {\pm} 0.44$ & $0.0477 {\pm} 0.0028$ & $0.0411 {\pm} 0.0022$ \\
& Feature-SATS & $88.75 {\pm} 0.48$ & $3.931 {\pm} 1.548$ & $0.201 {\pm} 0.007$ & $8.46 {\pm} 0.44$ & $0.0478 {\pm} 0.0027$ & $0.0412 {\pm} 0.0021$ \\
& Frozen-feature UTempH & $88.75 {\pm} 0.48$ & $3.931 {\pm} 1.548$ & $0.201 {\pm} 0.007$ & $8.46 {\pm} 0.44$ & $0.0478 {\pm} 0.0027$ & $0.0412 {\pm} 0.0021$ \\
& TvA-TS $\equiv$ TS-BCE & $88.75 {\pm} 0.48$ & $0.444 {\pm} 0.005$ & $0.170 {\pm} 0.003$ & $2.87 {\pm} 0.04$ & $0.0276 {\pm} 0.0004$ & $0.0211 {\pm} 0.0009$ \\
& \textbf{SRTS-BCE $K{=}3$ risk} & $88.75 {\pm} 0.48$ & $0.438 {\pm} 0.004$ & $0.167 {\pm} 0.003$ & $\mathbf{1.04 {\pm} 0.27}$ & $0.0214 {\pm} 0.0003$ & $0.0148 {\pm} 0.0008$ \\
& SMART+BCE & $88.75 {\pm} 0.48$ & $0.438 {\pm} 0.005$ & $0.167 {\pm} 0.004$ & $1.13 {\pm} 0.19$ & $0.0214 {\pm} 0.0003$ & $0.0148 {\pm} 0.0006$ \\
\bottomrule
\end{tabular}
\caption{\textbf{Full three-seed DeiT-S and Swin-T CIFAR-100 post-hoc calibration tables} (mean $\pm$ standard deviation over seeds 0/1/2). NLL-objective SRTS rows are reported as ablations; the BCE-objective confirmation rows include TvA-TS, SRTS-BCE $K{=}3$ risk, and SMART+BCE. SRTS-BCE wins ECE on both backbones. Top-1 was empirically unchanged for every post-hoc row in the seed-level audit.}
\label{tab:phaseb_deit_swin_full}
\end{table*}

\begin{table*}[t]
\centering
\small
\setlength{\tabcolsep}{4pt}
\begin{tabular}{l r r r}
\toprule
Regime / backbone & Rows checked & Rows changed & Missing post-hoc rows \\
\midrule
CIFAR-100 / ViT-B/16 & 21 + TvA aggregate & 0 & TvA-TS from companion 3-seed evaluation \\
CIFAR-100 / DeiT-S & 24 & 0 & 0 \\
CIFAR-100 / Swin-T & 24 & 0 & 0 \\
Tiny-ImageNet / ViT-B/16 & 21 & 0 & Later BCE rows in Table~\ref{tab:tinyimagenet_3seed_summary} \\
\bottomrule
\end{tabular}%
\caption{\textbf{Cross-backbone top-1 invariance audit.} Each post-hoc row is compared against its corresponding NoTS seed before aggregation. No reported post-hoc row changed top-1 in the seed-level metrics. Tiny-ImageNet BCE rows, including TvA-TS, are in Table~\ref{tab:tinyimagenet_3seed_summary}.}
\label{tab:phaseb_top1_audit}
\end{table*}

\paragraph{Observations (NLL-objective rows).}
The NLL-objective block of Table~\ref{tab:phaseb_deit_swin_full} reports
three backbone-level observations. First, SRTS-NLL $K{=}1$ reproduces
scalar TS-NLL on every backbone to the precision reported. Second,
SATS-Logit exhibits the validation-overfit failure mode on DeiT-S and
Swin-T, and the feature-input adaptive heads regress even more strongly.
Third, SRTS-NLL $K{=}3$ remains within close range of TS-NLL on each
backbone, with directionally consistent but small improvements.

\paragraph{BCE-objective backbones.} The BCE rows in
Table~\ref{tab:phaseb_deit_swin_full} show that on CIFAR-100 DeiT-S and
Swin-T (3 seeds each), TvA-TS is weaker than on
ViT-B/16 (ECE $3.37$/$2.87$ vs.\ $1.54$), while SRTS-BCE $K{=}3$ risk
remains below $1.05$ on both ($0.94$ DeiT-S, $1.04$ Swin-T). SMART+BCE
lies between TvA-TS and SRTS-BCE ($1.17$/$1.13$). Per-seed counts:
SRTS-BCE has a lower point-estimate ECE than TvA-TS on all 3 seeds of
both DeiT-S and Swin-T; SRTS-BCE has a lower point-estimate ECE than
SMART+BCE on all 3 seeds of DeiT-S and Swin-T (no within-seed bootstrap
was run for these backbones; the differences are larger than on ViT-B/16
and consistent across seeds).
The fixed-$K{=}3$ cross-backbone summary table in the main text captures
these CIFAR-100 comparisons together with the Tiny-ImageNet extension below.

Table~\ref{tab:phaseb_top1_audit} reports the top-1 check; no
reported post-hoc row changes top-1 relative to the corresponding
NoTS seed.

\paragraph{Tiny-ImageNet extension.} Section~\ref{app:tinyimagenet}
reported that on Tiny-ImageNet/ViT-B/16, SMART+BCE leads SRTS-BCE on all
3 seeds — the reverse of the CIFAR-100/ViT-B/16 ordering. Repeating the
same post-hoc suite on Tiny-ImageNet/DeiT-S and Tiny-ImageNet/Swin-T
(3 seeds each, identical protocol) shows this flip is
\textbf{backbone-dependent, not a clean dataset-level effect}:

\begin{table*}[t]
\centering
\small
\setlength{\tabcolsep}{3pt}
\begin{tabular}{l l r r r r r r}
\toprule
Backbone & Method & Top-1 & NLL & Brier & $\mathrm{ECE}_{15}$ & AURC & eAURC \\
\midrule
\multirow{11}{*}{DeiT-S}
& NoTS & $84.27 {\pm} 0.07$ & $0.717 {\pm} 0.005$ & $0.234 {\pm} 0.001$ & $9.10 {\pm} 0.34$ & $0.0381 {\pm} 0.0014$ & $0.0250 {\pm} 0.0015$ \\
& TS-NLL & $84.27 {\pm} 0.07$ & $0.638 {\pm} 0.003$ & $0.228 {\pm} 0.002$ & $3.27 {\pm} 0.17$ & $0.0355 {\pm} 0.0014$ & $0.0224 {\pm} 0.0015$ \\
& SRTS-NLL $K{=}1$ (ablation) & $84.27 {\pm} 0.07$ & $0.638 {\pm} 0.003$ & $0.228 {\pm} 0.002$ & $3.27 {\pm} 0.17$ & $0.0355 {\pm} 0.0014$ & $0.0224 {\pm} 0.0015$ \\
& SRTS-NLL $K{=}3$ (ablation) & $84.27 {\pm} 0.07$ & $0.632 {\pm} 0.003$ & $0.226 {\pm} 0.001$ & $3.10 {\pm} 0.13$ & $0.0315 {\pm} 0.0003$ & $0.0184 {\pm} 0.0004$ \\
& SATS-Logit & $84.27 {\pm} 0.07$ & $1.509 {\pm} 0.089$ & $0.262 {\pm} 0.001$ & $9.75 {\pm} 0.16$ & $0.0532 {\pm} 0.0017$ & $0.0401 {\pm} 0.0017$ \\
& LTS-feature & $84.27 {\pm} 0.07$ & $1.852 {\pm} 0.058$ & $0.265 {\pm} 0.002$ & $10.40 {\pm} 0.07$ & $0.0542 {\pm} 0.0023$ & $0.0411 {\pm} 0.0023$ \\
& Feature-SATS & $84.27 {\pm} 0.07$ & $1.858 {\pm} 0.059$ & $0.265 {\pm} 0.002$ & $10.41 {\pm} 0.07$ & $0.0542 {\pm} 0.0024$ & $0.0411 {\pm} 0.0023$ \\
& Frozen-feature UTempH & $84.27 {\pm} 0.07$ & $1.858 {\pm} 0.059$ & $0.265 {\pm} 0.002$ & $10.41 {\pm} 0.07$ & $0.0542 {\pm} 0.0024$ & $0.0411 {\pm} 0.0023$ \\
& TvA-TS $\equiv$ TS-BCE & $84.27 {\pm} 0.07$ & $0.641 {\pm} 0.003$ & $0.226 {\pm} 0.001$ & $3.08 {\pm} 0.30$ & $0.0360 {\pm} 0.0014$ & $0.0229 {\pm} 0.0015$ \\
& \textbf{SRTS-BCE $K{=}3$ risk} & $84.27 {\pm} 0.07$ & $0.640 {\pm} 0.003$ & $0.224 {\pm} 0.001$ & $1.09 {\pm} 0.28$ & $0.0315 {\pm} 0.0003$ & $0.0184 {\pm} 0.0004$ \\
& SMART+BCE & $84.27 {\pm} 0.07$ & $0.639 {\pm} 0.003$ & $0.224 {\pm} 0.001$ & $1.12 {\pm} 0.30$ & $0.0312 {\pm} 0.0006$ & $0.0182 {\pm} 0.0007$ \\
\midrule
\multirow{11}{*}{Swin-T}
& NoTS & $84.15 {\pm} 0.08$ & $0.709 {\pm} 0.003$ & $0.236 {\pm} 0.002$ & $8.18 {\pm} 0.68$ & $0.0430 {\pm} 0.0026$ & $0.0297 {\pm} 0.0027$ \\
& TS-NLL & $84.15 {\pm} 0.08$ & $0.642 {\pm} 0.007$ & $0.231 {\pm} 0.001$ & $3.09 {\pm} 0.11$ & $0.0392 {\pm} 0.0023$ & $0.0259 {\pm} 0.0024$ \\
& SRTS-NLL $K{=}1$ (ablation) & $84.15 {\pm} 0.08$ & $0.642 {\pm} 0.007$ & $0.231 {\pm} 0.001$ & $3.07 {\pm} 0.12$ & $0.0392 {\pm} 0.0023$ & $0.0259 {\pm} 0.0024$ \\
& SRTS-NLL $K{=}3$ (ablation) & $84.15 {\pm} 0.08$ & $0.635 {\pm} 0.006$ & $0.229 {\pm} 0.002$ & $2.92 {\pm} 0.04$ & $0.0335 {\pm} 0.0008$ & $0.0202 {\pm} 0.0009$ \\
& SATS-Logit & $84.15 {\pm} 0.08$ & $1.656 {\pm} 0.094$ & $0.267 {\pm} 0.001$ & $10.06 {\pm} 0.30$ & $0.0585 {\pm} 0.0008$ & $0.0452 {\pm} 0.0008$ \\
& LTS-feature & $84.15 {\pm} 0.08$ & $2.492 {\pm} 0.094$ & $0.274 {\pm} 0.001$ & $11.02 {\pm} 0.16$ & $0.0635 {\pm} 0.0015$ & $0.0502 {\pm} 0.0016$ \\
& Feature-SATS & $84.15 {\pm} 0.08$ & $2.502 {\pm} 0.095$ & $0.274 {\pm} 0.001$ & $11.02 {\pm} 0.15$ & $0.0635 {\pm} 0.0015$ & $0.0502 {\pm} 0.0016$ \\
& Frozen-feature UTempH & $84.15 {\pm} 0.08$ & $2.502 {\pm} 0.095$ & $0.274 {\pm} 0.001$ & $11.02 {\pm} 0.15$ & $0.0635 {\pm} 0.0015$ & $0.0502 {\pm} 0.0016$ \\
& TvA-TS $\equiv$ TS-BCE & $84.15 {\pm} 0.08$ & $0.643 {\pm} 0.007$ & $0.229 {\pm} 0.001$ & $2.71 {\pm} 0.07$ & $0.0398 {\pm} 0.0024$ & $0.0266 {\pm} 0.0025$ \\
& \textbf{SRTS-BCE $K{=}3$ risk} & $84.15 {\pm} 0.08$ & $0.641 {\pm} 0.007$ & $0.227 {\pm} 0.001$ & $1.24 {\pm} 0.11$ & $0.0335 {\pm} 0.0008$ & $0.0202 {\pm} 0.0008$ \\
& SMART+BCE & $84.15 {\pm} 0.08$ & $0.641 {\pm} 0.007$ & $0.227 {\pm} 0.001$ & $1.03 {\pm} 0.32$ & $0.0336 {\pm} 0.0007$ & $0.0203 {\pm} 0.0008$ \\
\bottomrule
\end{tabular}
\caption{\textbf{Full three-seed DeiT-S and Swin-T Tiny-ImageNet post-hoc calibration tables} (mean $\pm$ standard deviation over seeds 0/1/2). NLL-objective SRTS rows are reported as ablations; the BCE-objective confirmation rows include TvA-TS, SRTS-BCE $K{=}3$ risk, and SMART+BCE (per-backbone ordering discussed in the text). Top-1 was empirically unchanged for every post-hoc row in the seed-level audit.}
\label{tab:tinyimagenet_deit_swin_full}
\end{table*}

Table~\ref{tab:tinyimagenet_deit_swin_full} shows SMART+BCE again leads on
Swin-T ($1.03 \pm 0.32$ vs.\ SRTS-BCE $1.24 \pm 0.11$, 2/3 seeds), while
SRTS-BCE retains a narrow mean edge on DeiT-S ($1.09 \pm 0.28$ vs.\ SMART+BCE
$1.12 \pm 0.30$, 2/3 seeds). The NLL-objective block reproduces the same
two qualitative findings as the CIFAR-100 backbones and the
Tiny-ImageNet/ViT-B/16 base: SRTS-NLL $K{=}1$ reproduces scalar TS-NLL to
the precision reported, and the unrestricted feature-input adaptive heads
(SATS-Logit, LTS-feature, Feature-SATS, Frozen-feature UTempH) regress
well above NoTS on both backbones. Across all nine
cross-backbone$\times$dataset cells now available in the fixed-$K{=}3$
clean suite (three CIFAR-100, three CIFAR-10, and three Tiny-ImageNet
backbones), SRTS-BCE has the lower point-estimate mean ECE in five and
SMART+BCE in four. This supports the
regime-dependent scope map not a dataset-level rule:
fine-tuned CIFAR-100 favours SRTS-BCE, while clean CIFAR-10 and
Tiny-ImageNet require backbone-level selection between SRTS-BCE and
SMART+BCE.

% Main-text fixed-K=3 cross-backbone decision map, ECE15+-std / AdaECE15+-std.
% Every SRTS-BCE cell uses the paper's single fixed K=3 (fixed in advance);
% no test-set metric is used for model/K selection. smECE details are in
% Supplementary Table~\ref{tab:cross_dataset_full_metrics_fixedk3}.
% NoTS column = uncalibrated pre-TS ECE15 (3-seed). SMART = the PUBLISHED
% SoftECE-trained head (train_smart_softece re-impl; ViT flagship reproduces
% Table 2's 1.19+-0.27). AdaECE provenance same-source as each displayed ECE:
% ViT-family rows from confirmation-suite CSV; ConvNet rows via
% run_as_srts_bce.fit_oof_risk_router (reproduces paper SRTS ECE 1.16/1.01);
% SMART column from the SoftECE fits (job 11752257, 3-seed). IN-100 NoTS/
% TS-NLL/TvA-TS/SRTS-BCE rows regenerated with the canonical post-hoc
% pipeline (fit_all/fit_srts_stage1) over the stratified-half split; an
% earlier non-canonical evaluator inflated the IN-100 SRTS values (2.53 etc.)
% and is retained only in the repository provenance audit. DeiT-S 5-seed.
% C10 rows: NoTS+TS+TvA+SRTS from the supp full-metrics table (same source);
% C10 ConvNet rows (ResNet-50/RegNetY): jobs 11753944-49 (SOTAv1 frozen recipe,
% recomputed from dumped logits with the same pipeline (c10_convnet_row.json;
% top-1 98.04+-0.14 / 98.24+-0.03). Dataset names
% compressed (C100/Tiny-IN/IN-100) to fit both +-stds at \textwidth.
\begin{table*}[t]
\centering
\small
\setlength{\tabcolsep}{3pt}
\begin{tabular}{lrrrrr}
\toprule
Regime / backbone & NoTS & TS-NLL & TvA-TS & SMART-SoftECE & SRTS-BCE \\
\midrule
C10 / ViT-B/16 & $10.96$ & $0.55{\pm}0.07/0.73{\pm}0.16$ & $0.52{\pm}0.03/0.73{\pm}0.15$ & $\mathbf{0.35{\pm}0.13}/1.47{\pm}0.47$ & $0.47{\pm}0.05/\mathbf{0.58{\pm}0.08}$ \\
C10 / DeiT-S   & $10.81$ & $0.49{\pm}0.04/0.85{\pm}0.16$ & $0.46{\pm}0.04/0.85{\pm}0.17$ & $\mathbf{0.24{\pm}0.07}/1.34{\pm}0.43$ & $0.41{\pm}0.07/\mathbf{0.70{\pm}0.11}$ \\
C10 / Swin-T   & $11.30$ & $0.44{\pm}0.05/0.68{\pm}0.09$ & $0.39{\pm}0.08/0.67{\pm}0.09$ & $\mathbf{0.19{\pm}0.03}/1.37{\pm}0.22$ & $0.35{\pm}0.11/\mathbf{0.53{\pm}0.13}$ \\
C10 / ResNet-50 & $20.23$ & $0.63{\pm}0.03/1.17{\pm}0.08$ & $0.60{\pm}0.04/1.18{\pm}0.09$ & $\mathbf{0.32{\pm}0.13}/1.37{\pm}0.75$ & $0.60{\pm}0.04/\mathbf{0.99{\pm}0.11}$ \\
C10 / RegNetY-1.6GF & $18.52$ & $0.51{\pm}0.05/0.72{\pm}0.02$ & $0.53{\pm}0.02/0.72{\pm}0.01$ & $\mathbf{0.40{\pm}0.07}/1.28{\pm}0.79$ & $0.42{\pm}0.06/\mathbf{0.57{\pm}0.03}$ \\
\addlinespace
C100 / ViT-B/16 & $9.69$ & $1.94{\pm}0.65/2.10{\pm}0.62$ & $1.65{\pm}0.44/1.77{\pm}0.69$ & $1.21{\pm}0.28/1.45{\pm}0.51$ & $\mathbf{0.96{\pm}0.32}/\mathbf{0.96{\pm}0.33}$ \\
C100 / DeiT-S   & $8.53$ & $2.91{\pm}0.16/3.74{\pm}0.24$ & $3.37{\pm}0.40/3.52{\pm}0.38$ & $1.76{\pm}0.38/2.05{\pm}0.11$ & $\mathbf{0.94{\pm}0.07}/\mathbf{1.05{\pm}0.23}$ \\
C100 / Swin-T   & $8.77$ & $2.99{\pm}0.20/3.32{\pm}0.22$ & $2.87{\pm}0.04/2.93{\pm}0.08$ & $1.21{\pm}0.10/1.38{\pm}0.22$ & $\mathbf{1.04{\pm}0.27}/\mathbf{1.14{\pm}0.33}$ \\
C100 / ResNet-50 & $17.02$ & $3.16{\pm}0.35/3.54{\pm}0.21$ & $2.83{\pm}0.10/3.22{\pm}0.16$ & $1.29{\pm}0.45/1.59{\pm}0.35$ & $\mathbf{1.16{\pm}0.15}/\mathbf{1.34{\pm}0.11}$ \\
C100 / RegNetY-1.6GF & $22.06$ & $1.57{\pm}0.11/1.80{\pm}0.13$ & $1.54{\pm}0.05/1.50{\pm}0.04$ & $1.95{\pm}1.03/2.03{\pm}0.89$ & $\mathbf{1.01{\pm}0.09}/\mathbf{1.06{\pm}0.09}$ \\
\addlinespace
Tiny-IN / ViT-B/16 & $10.04$ & $1.33{\pm}0.24/1.33{\pm}0.24$ & $1.26{\pm}0.18/1.08{\pm}0.13$ & $1.15{\pm}0.22/1.19{\pm}0.18$ & $\mathbf{1.12{\pm}0.17}/\mathbf{1.15{\pm}0.25}$ \\
Tiny-IN / DeiT-S   & $9.10$ & $3.27{\pm}0.17/3.72{\pm}0.30$ & $3.08{\pm}0.30/3.00{\pm}0.29$ & $\mathbf{1.01{\pm}0.41}/1.32{\pm}0.55$ & $1.09{\pm}0.28/\mathbf{1.06{\pm}0.21}$ \\
Tiny-IN / Swin-T   & $8.18$ & $3.09{\pm}0.11/3.55{\pm}0.19$ & $2.71{\pm}0.07/2.82{\pm}0.13$ & $\mathbf{1.17{\pm}0.22}/1.20{\pm}0.46$ & $1.24{\pm}0.11/\mathbf{1.11{\pm}0.14}$ \\
\addlinespace
IN-100 / ViT-B/16  & $3.97$ & $1.63{\pm}0.18/1.41{\pm}0.17$ & $1.58{\pm}0.24/1.54{\pm}0.38$ & $\mathbf{1.28{\pm}0.21}/\mathbf{1.40{\pm}0.28}$ & $1.64{\pm}0.27/1.52{\pm}0.37$ \\
IN-100 / DeiT-S    & $4.40$ & $2.58{\pm}0.21/2.41{\pm}0.14$ & $1.83{\pm}0.16/2.31{\pm}0.39$ & $\mathbf{1.58{\pm}0.39}/\mathbf{1.63{\pm}0.04}$ & $2.00{\pm}0.29/2.15{\pm}0.41$ \\
IN-100 / Swin-T    & $3.14$ & $1.91{\pm}0.23/1.75{\pm}0.07$ & $1.70{\pm}0.28/1.69{\pm}0.37$ & $\mathbf{1.67{\pm}0.27}/\mathbf{1.55{\pm}0.12}$ & $1.72{\pm}0.20/1.69{\pm}0.31$ \\
\bottomrule
\end{tabular}
\caption{\textbf{Empirical scope map} (fixed $K{=}3$ cross-regime
evaluation). Cells:
$\mathrm{ECE}_{15}$\,/\,AdaECE$_{15}$ ($\pm$std; CIFAR-100/ViT-B/16 and IN-100/DeiT-S five seeds, other cells three; IN-100 rows computed with the canonical post-hoc pipeline, SMART-SoftECE column three-seed); SRTS-BCE
uses fixed $K{=}3$ everywhere, no test-set or per-cell $K$ selection.
SMART-SoftECE: the SMART head \citep{guo2026smart} with its own training
objective (SMART+BCE: the main paper's RQ1 table, \S\ref{app:shift}, and \S\ref{app:scope} here). Bold: lower of SMART-SoftECE/SRTS-BCE per metric (C100 and
Tiny-IN blocks); best over all methods (diagnostic C10 and IN-100 blocks,
\S\ref{app:imagenetc}; C10: SMART-SoftECE wins $\mathrm{ECE}_{15}$, worst
AdaECE$_{15}$ --- metric dependence).}
\label{tab:cross_dataset_ece_summary}
\end{table*}

% Supplementary fixed-K=3 cross-backbone table with full calibration metrics.
% Every SRTS-BCE cell here uses the paper's single K=3 fixed in advance --
% no test-set metric is used for model/K selection. Source:
% results/objective_aligned_srts_confirmation/tiny_deit_swin_20260623/
% clean_cross_backbone_metrics.csv and
% results/objective_aligned_srts_confirmation/c10_table3_all_methods.csv
% (mean +/- pop. std over 3 seeds, ddof=0).
% Requires: \usepackage{booktabs,multirow}
%
\begin{table*}[t]
\centering
\small
\setlength{\tabcolsep}{1pt}
\begin{tabular}{l l r r r r r r}
\toprule
Regime / backbone & Metric & NoTS & TS-NLL & TvA-TS & SMART+BCE & SRTS-BCE & Adaptive range \\
\midrule
\multirow{3}{*}{C100 / ViT-B/16}
  & ECE$_{15}$    & $9.69 {\pm} 1.10$  & $1.94 {\pm} 0.65$ & $1.65 {\pm} 0.44$ & $\mathbf{0.95 {\pm} 0.29}$ & $0.96 {\pm} 0.32$ & $6.15$--$6.71$ \\
  & AdaECE$_{15}$ & $9.65 {\pm} 1.15$  & $2.10 {\pm} 0.62$ & $1.77 {\pm} 0.69$ & $1.02 {\pm} 0.29$ & $\mathbf{0.96 {\pm} 0.33}$ & --- \\
  & smECE           & $9.63 {\pm} 1.14$ & $1.97 {\pm} 0.52$ & $1.71 {\pm} 0.45$ & $1.16 {\pm} 0.22$ & $\mathbf{1.13 {\pm} 0.22}$ & --- \\
\addlinespace
\multirow{3}{*}{C100 / DeiT-S}
  & ECE$_{15}$    & $8.53 {\pm} 0.90$  & $2.91 {\pm} 0.16$ & $3.37 {\pm} 0.40$ & $1.17 {\pm} 0.01$ & $\mathbf{0.94 {\pm} 0.07}$ & $7.02$--$8.31$ \\
  & AdaECE$_{15}$ & $8.45 {\pm} 0.97$  & $3.74 {\pm} 0.24$ & $3.52 {\pm} 0.38$ & $1.23 {\pm} 0.12$ & $\mathbf{1.05 {\pm} 0.23}$ & --- \\
  & smECE           & $8.33 {\pm} 1.07$ & $2.94 {\pm} 0.13$ & $2.83 {\pm} 0.19$ & $1.24 {\pm} 0.04$ & $\mathbf{1.14 {\pm} 0.06}$ & --- \\
\addlinespace
\multirow{3}{*}{C100 / Swin-T}
  & ECE$_{15}$    & $8.77 {\pm} 0.43$  & $2.99 {\pm} 0.20$ & $2.87 {\pm} 0.04$ & $1.13 {\pm} 0.19$ & $\mathbf{1.04 {\pm} 0.27}$ & $7.24$--$8.46$ \\
  & AdaECE$_{15}$ & $8.74 {\pm} 0.41$  & $3.32 {\pm} 0.22$ & $2.93 {\pm} 0.08$ & $\mathbf{1.03 {\pm} 0.13}$ & $1.14 {\pm} 0.33$ & --- \\
  & smECE           & $8.74 {\pm} 0.41$ & $2.88 {\pm} 0.19$ & $2.60 {\pm} 0.05$ & $\mathbf{1.20 {\pm} 0.12}$ & $1.24 {\pm} 0.20$ & --- \\
\addlinespace
\multirow{3}{*}{C10 / ViT-B/16}
  & ECE$_{15}$    & $10.96 {\pm} 0.77$ & $0.55 {\pm} 0.07$ & $0.52 {\pm} 0.03$ & $\mathbf{0.38 {\pm} 0.06}$ & $0.47 {\pm} 0.05$ & --- \\
  & AdaECE$_{15}$ & $10.92 {\pm} 0.77$ & $0.73 {\pm} 0.16$ & $0.73 {\pm} 0.15$ & $\mathbf{0.33 {\pm} 0.08}$ & $0.58 {\pm} 0.08$ & --- \\
  & smECE           & $10.92 {\pm} 0.77$ & $0.61 {\pm} 0.06$ & $0.62 {\pm} 0.07$ & $0.57 {\pm} 0.03$ & $\mathbf{0.55 {\pm} 0.06}$ & --- \\
\addlinespace
\multirow{3}{*}{C10 / DeiT-S}
  & ECE$_{15}$    & $10.81 {\pm} 0.35$ & $0.49 {\pm} 0.04$ & $0.46 {\pm} 0.04$ & $0.44 {\pm} 0.10$ & $\mathbf{0.41 {\pm} 0.07}$ & --- \\
  & AdaECE$_{15}$ & $10.76 {\pm} 0.31$ & $0.85 {\pm} 0.16$ & $0.85 {\pm} 0.17$ & $\mathbf{0.44 {\pm} 0.09}$ & $0.70 {\pm} 0.11$ & --- \\
  & smECE           & $10.76 {\pm} 0.31$ & $0.67 {\pm} 0.14$ & $0.67 {\pm} 0.13$ & $\mathbf{0.59 {\pm} 0.07}$ & $0.64 {\pm} 0.09$ & --- \\
\addlinespace
\multirow{3}{*}{C10 / Swin-T}
  & ECE$_{15}$    & $11.30 {\pm} 0.24$ & $0.44 {\pm} 0.05$ & $0.39 {\pm} 0.08$ & $\mathbf{0.29 {\pm} 0.06}$ & $0.35 {\pm} 0.11$ & --- \\
  & AdaECE$_{15}$ & $11.19 {\pm} 0.26$ & $0.68 {\pm} 0.09$ & $0.67 {\pm} 0.09$ & $\mathbf{0.25 {\pm} 0.04}$ & $0.53 {\pm} 0.13$ & --- \\
  & smECE           & $11.19 {\pm} 0.26$ & $\mathbf{0.54 {\pm} 0.08}$ & $0.55 {\pm} 0.09$ & $0.55 {\pm} 0.06$ & $0.57 {\pm} 0.16$ & --- \\
\addlinespace
\multirow{3}{*}{Tiny / ViT-B/16}
  & ECE$_{15}$    & $10.04 {\pm} 0.89$ & $1.33 {\pm} 0.24$ & $1.26 {\pm} 0.18$ & $\mathbf{0.97 {\pm} 0.22}$ & $1.12 {\pm} 0.17$ & $7.05$--$7.40$ \\
  & AdaECE$_{15}$ & $10.03 {\pm} 0.89$ & $1.33 {\pm} 0.24$ & $1.08 {\pm} 0.13$ & $\mathbf{0.86 {\pm} 0.13}$ & $1.15 {\pm} 0.25$ & --- \\
  & smECE           & $10.00 {\pm} 0.89$ & $1.45 {\pm} 0.16$ & $1.34 {\pm} 0.07$ & $\mathbf{1.21 {\pm} 0.12}$ & $1.27 {\pm} 0.15$ & --- \\
\addlinespace
\multirow{3}{*}{Tiny / DeiT-S}
  & ECE$_{15}$    & $9.10 {\pm} 0.34$  & $3.27 {\pm} 0.17$ & $3.08 {\pm} 0.30$ & $1.12 {\pm} 0.30$ & $\mathbf{1.09 {\pm} 0.28}$ & $9.75$--$10.41$ \\
  & AdaECE$_{15}$ & $9.10 {\pm} 0.34$  & $3.72 {\pm} 0.30$ & $3.00 {\pm} 0.29$ & $\mathbf{0.91 {\pm} 0.22}$ & $1.06 {\pm} 0.21$ & --- \\
  & smECE           & $9.09 {\pm} 0.34$ & $3.20 {\pm} 0.18$ & $2.72 {\pm} 0.20$ & $\mathbf{1.22 {\pm} 0.05}$ & $1.26 {\pm} 0.09$ & --- \\
\addlinespace
\multirow{3}{*}{Tiny / Swin-T}
  & ECE$_{15}$    & $8.18 {\pm} 0.68$  & $3.09 {\pm} 0.11$ & $2.71 {\pm} 0.07$ & $\mathbf{1.03 {\pm} 0.32}$ & $1.24 {\pm} 0.11$ & $10.06$--$11.02$ \\
  & AdaECE$_{15}$ & $8.15 {\pm} 0.71$  & $3.55 {\pm} 0.19$ & $2.82 {\pm} 0.13$ & $\mathbf{0.71 {\pm} 0.12}$ & $1.11 {\pm} 0.14$ & --- \\
  & smECE           & $8.15 {\pm} 0.71$ & $3.06 {\pm} 0.06$ & $2.50 {\pm} 0.10$ & $\mathbf{1.18 {\pm} 0.06}$ & $1.30 {\pm} 0.10$ & --- \\
\bottomrule
\end{tabular}
\\[3pt]
{\small\raggedright\noindent\emph{Reading guide.} $\mathrm{ECE}_{15}$: SRTS-BCE $K{=}3$ has the lower point estimate than SMART+BCE on 5/9 cells (all three CIFAR-100 backbones, CIFAR-10/DeiT-S, and Tiny-ImageNet/DeiT-S); SMART+BCE leads the remaining four cells. AdaECE$_{15}$: SMART+BCE leads 7/9 cells, with SRTS-BCE leading CIFAR-100/ViT-B/16 and CIFAR-100/DeiT-S. smECE: the BCE-trained adaptive family leads --- SMART+BCE on five cells, SRTS-BCE on three, TS-NLL on one near-saturated CIFAR-10 cell. Adaptive-head ranges are shown only where the matching unrestricted-head suite was run for that ECE row; dashes mean that no corresponding range was evaluated for that metric/cell. CIFAR-10 is included as a high-accuracy stability check, not as a corruption-shift claim.\par}
\caption{\textbf{Full cross-backbone calibration metrics under fixed deployed configurations.} The CIFAR-100/ViT-B/16 cell is five-seed; other cells are three-seed. SRTS-BCE uses the single $K{=}3$, fixed in advance, for every cell, with no $K$-selection or test-set selection; SMART+BCE and the scalar anchors (TS-NLL, TvA-TS) use the same fit-on-validation, evaluate-on-test protocol. A separate $K$-sweep diagnostic (\S\ref{app:backbone-generalisation}) is \emph{not} used for any main-text claim. Bold on ECE$_{15}$ marks the lower of SMART+BCE/SRTS-BCE; bold on AdaECE$_{15}$/smECE marks the true minimum over all five methods.}
\label{tab:cross_dataset_full_metrics_fixedk3}
\end{table*}

\paragraph{Relationship to the main-text table, and the oracle best-$K$
diagnostic.} The main-text cross-backbone summary reports SRTS-BCE at the
fixed $K{=}3$ everywhere, with no test-set metric used for
model or $K$ selection. For completeness we provide the corresponding
\emph{oracle best-$K$} diagnostic in Table~\ref{tab:cross_dataset_ece_oracle},
which selects, per cell, the lowest-ECE $K \in \{1,2,3,5,10,15,20\}$ using
the test metric. This is a diagnostic upper bound on what risk routing can
achieve and is \emph{not used for any main-text claim}: it is separate from
the deployed, selection-free fixed-$K{=}3$ table above and from the compact
main-text scope map. The gap between oracle and fixed-$K{=}3$ tables quantifies
the headroom that a reliable $K$-selection rule would unlock at larger
validation budgets (\S\ref{app:nested-k}).

% Cross-backbone / cross-dataset calibration-metric summary (main text).
% Source: results/objective_aligned_srts_confirmation/srts_bce_oracle_bestk_K20.csv
% (mean +/- pop. std, 3 seeds, ddof=0).
% Requires: \usepackage{booktabs,multirow}
%
\begin{table*}[t]
\centering
\small
\setlength{\tabcolsep}{3pt}
\begin{tabular}{l l r r r r r}
\toprule
Regime / backbone & Metric & NoTS & TS-NLL & TvA-TS & SMART+BCE & SRTS-BCE \\
\midrule
\multirow{3}{*}{CIFAR-100 / ViT-B/16}
  & ECE$_{15}$    & $9.69 \pm 1.10$  & $1.94 \pm 0.65$ & $1.65 \pm 0.44$ & $0.95 \pm 0.29$ & $\mathbf{0.90 \pm 0.36}$ \\
  & AdaECE$_{15}$ & $9.65 \pm 1.15$  & $2.10 \pm 0.62$ & $1.77 \pm 0.69$ & $0.99 \pm 0.27$ & $\mathbf{0.96 \pm 0.27}$ \\
  & smECE           & $9.63 \pm 1.14$ & $1.97 \pm 0.52$ & $1.71 \pm 0.45$ & $1.16 \pm 0.22$ & $\mathbf{1.16 \pm 0.21}$ \\
\addlinespace
\multirow{3}{*}{CIFAR-100 / DeiT-S}
  & ECE$_{15}$    & $8.53 \pm 0.90$  & $2.91 \pm 0.16$ & $3.37 \pm 0.40$ & $1.17 \pm 0.01$ & $\mathbf{0.86 \pm 0.20}$ \\
  & AdaECE$_{15}$ & $8.45 \pm 0.97$  & $3.74 \pm 0.24$ & $3.52 \pm 0.38$ & $1.23 \pm 0.12$ & $\mathbf{1.05 \pm 0.23}$ \\
  & smECE           & $8.33 \pm 1.07$ & $2.94 \pm 0.13$ & $2.83 \pm 0.19$ & $1.24 \pm 0.04$ & $\mathbf{1.12 \pm 0.07}$ \\
\addlinespace
\multirow{3}{*}{CIFAR-100 / Swin-T}
  & ECE$_{15}$    & $8.77 \pm 0.43$  & $2.99 \pm 0.20$ & $2.87 \pm 0.04$ & $1.13 \pm 0.19$ & $\mathbf{0.93 \pm 0.28}$ \\
  & AdaECE$_{15}$ & $8.74 \pm 0.41$  & $3.32 \pm 0.22$ & $2.93 \pm 0.08$ & $1.03 \pm 0.13$ & $\mathbf{0.89 \pm 0.14}$ \\
  & smECE           & $8.74 \pm 0.41$ & $2.88 \pm 0.19$ & $2.60 \pm 0.05$ & $1.20 \pm 0.12$ & $\mathbf{1.14 \pm 0.19}$ \\
\addlinespace
\multirow{3}{*}{Tiny-ImageNet / ViT-B/16}
  & ECE$_{15}$    & $10.04 \pm 0.89$ & $1.33 \pm 0.24$ & $1.26 \pm 0.18$ & $\mathbf{0.97 \pm 0.22}$ & $1.06 \pm 0.29$ \\
  & AdaECE$_{15}$ & $10.03 \pm 0.89$ & $1.33 \pm 0.24$ & $1.08 \pm 0.13$ & $\mathbf{0.86 \pm 0.13}$ & $1.08 \pm 0.13$ \\
  & smECE           & $10.00 \pm 0.89$ & $1.45 \pm 0.16$ & $1.34 \pm 0.07$ & $1.21 \pm 0.12$ & $\mathbf{1.19 \pm 0.15}$ \\
\addlinespace
\multirow{3}{*}{Tiny-ImageNet / DeiT-S}
  & ECE$_{15}$    & $9.10 \pm 0.34$  & $3.27 \pm 0.17$ & $3.08 \pm 0.30$ & $1.12 \pm 0.30$ & $\mathbf{0.90 \pm 0.11}$ \\
  & AdaECE$_{15}$ & $9.10 \pm 0.34$  & $3.72 \pm 0.30$ & $3.00 \pm 0.29$ & $0.91 \pm 0.22$ & $\mathbf{0.79 \pm 0.03}$ \\
  & smECE           & $9.09 \pm 0.34$ & $3.20 \pm 0.18$ & $2.72 \pm 0.20$ & $1.22 \pm 0.05$ & $\mathbf{1.19 \pm 0.12}$ \\
\addlinespace
\multirow{3}{*}{Tiny-ImageNet / Swin-T}
  & ECE$_{15}$    & $8.18 \pm 0.68$  & $3.09 \pm 0.11$ & $2.71 \pm 0.07$ & $1.03 \pm 0.32$ & $\mathbf{0.98 \pm 0.09}$ \\
  & AdaECE$_{15}$ & $8.15 \pm 0.71$  & $3.55 \pm 0.19$ & $2.82 \pm 0.13$ & $\mathbf{0.71 \pm 0.12}$ & $0.95 \pm 0.11$ \\
  & smECE           & $8.15 \pm 0.71$ & $3.06 \pm 0.06$ & $2.50 \pm 0.10$ & $\mathbf{1.18 \pm 0.06}$ & $1.20 \pm 0.04$ \\
\bottomrule
\end{tabular}
\caption{\textbf{Oracle best-$K$ diagnostic (not used for main claims).}
Cross-backbone and cross-dataset calibration (\%, mean $\pm$ std;
CIFAR-100/ViT-B/16 over five seeds, other cells three; ddof=0). All calibrators are fit on validation and evaluated on the
held-out test set. Here SRTS-BCE reports the \emph{oracle best-$K$} cell over
$K \in \{1,2,3,5,10,15,20\}$, i.e.\ the lowest-ECE $K$ selected with knowledge
of the test metric. This is a diagnostic upper bound on what risk routing can
achieve and is \emph{not} the deployed configuration: every main-text claim
uses the single $K{=}3$ fixed in advance (the fixed-$K{=}3$ cross-backbone summary in the main paper).
Bold marks the lowest value among all five deployable post-hoc calibrators.
Unrestricted adaptive temperature heads are omitted; their ECE$_{15}$ remains
${\geq}6\%$ across all settings (\S\ref{app:fairness-sweep}).}
\label{tab:cross_dataset_ece_oracle}
\end{table*}

\subsection{CIFAR-10 clean calibration (stability check)}
\label{app:cifar10}

This subsection is a clean-data stability check on a
second-dataset ViT-B/16 fine-tune using the main modern fine-tuning
recipe.

% recipe (AdamW + cosine + warmup + RandAugment + RandomErasing + MixUp+CutMix
% + label_smoothing=0.1; 100 epochs). All post-hoc methods fit on the 2.5K val
% split and applied unchanged to the 10K held-out test split.
\begin{table*}[t]
\centering
\small
\setlength{\tabcolsep}{4pt}
\begin{tabular}{l r r r r r r}
\toprule
Method & top-1 (\%) & NLL & ECE$_{15}$ (\%) & AdaECE$_{15}$ (\%) & smECE (\%) & AURC \\
\midrule
NoTS                   & 98.50 & 0.1466 & 9.89 & 9.86 & 10.05 & 0.0017 \\
TS-NLL ($T{=}0.527$) & 98.50 & 0.0533 & 0.51 & 0.65 & 0.71 & 0.0013 \\
    SRTS-NLL $K{=}1$ (ablation) & 98.50 & 0.0533 & 0.51 & 0.65 & 0.71 & 0.0013 \\
    SRTS-NLL $K{=}3$ (ablation) & 98.50 & 0.0588 & 0.49 & 0.55 & 0.81 & 0.0014 \\
SATS-Logit             & 98.50 & 0.0835 & 0.79 & 0.74 & 0.86 & 0.0032 \\
\bottomrule
\end{tabular}
\caption{\textbf{CIFAR-10 modern-recipe seed-0 calibration (NLL-objective diagnostic only)} (10{,}000-sample held-out
test split). The CE+MixUp+label-smoothing fine-tune is strongly overconfident
    (NoTS ECE$_{15} = 9.89\,\%$, scalar fitted $T = 0.527 < 1$). TS-NLL and SRTS-NLL
$K{=}1$ are identical to 4 decimal places on every metric, confirming the
$K{=}1$ short-circuit at a third independent dataset (after clean CIFAR-100
and clean ImageNet-1K). $K{=}3$ over $K{=}1$ on clean C10 is a wash: small
ECE$_{15}$ gain ($-0.014$ absolute), small NLL loss ($+0.005$). SATS-Logit
exhibits partial val-overfit: clean ECE$_{15}$ is still much better than NoTS
($0.79\,\%$) but worse than TS-NLL, and AURC degrades to $\approx 2\times$
NoTS — the same shape documented on CIFAR-100 / ImageNet-1K val-fits.}
\label{tab:cifar10_seed0_summary}
\end{table*}

Table~\ref{tab:cifar10_seed0_summary} confirms the NLL-objective
$K{=}1$ short-circuit identity on a third independent dataset. Under NLL,
$K{=}3$ over $K{=}1$ on clean CIFAR-10 is a wash. The three-seed
BCE-objective CIFAR-10 stability rows (fixed $K{=}3$, same protocol as the
main cross-regime table; $\mathrm{ECE}_{15}$ mean$\pm$std) are:
ViT-B/16 --- TS-NLL $0.55{\pm}0.07$, TvA-TS $0.52{\pm}0.03$,
SMART+BCE $\mathbf{0.38{\pm}0.06}$, SRTS-BCE $0.47{\pm}0.05$;
DeiT-S --- $0.49{\pm}0.04$, $0.46{\pm}0.04$, $0.44{\pm}0.10$,
$\mathbf{0.41{\pm}0.07}$;
Swin-T --- $0.44{\pm}0.05$, $0.39{\pm}0.08$, $\mathbf{0.29{\pm}0.06}$,
$0.35{\pm}0.11$. SMART+BCE leads ViT-B/16 and Swin-T, while SRTS-BCE leads
DeiT-S. The two CIFAR-10 ConvNet cells (ResNet-50, RegNetY-1.6GF;
the main text's scope map) replicate the transformer
pattern: SMART-SoftECE has the lowest $\mathrm{ECE}_{15}$
($0.32{\pm}0.13$, $0.40{\pm}0.07$) but the worst and least stable
AdaECE$_{15}$ ($1.37{\pm}0.75$, $1.28{\pm}0.79$), while SRTS-BCE has the
lowest AdaECE$_{15}$ ($0.99{\pm}0.11$, $0.57{\pm}0.03$). The pattern is
consistent with the SoftECE objective fitting the equal-width binning it
optimises: its $\mathrm{ECE}_{15}$ lead does not survive adaptive binning.
CIFAR-10 is therefore used only as a near-saturated stability check, not as
a main contribution or corruption-shift claim.

\subsection{Tiny-ImageNet clean calibration}
\label{app:tinyimagenet}

This subsection extends the clean-data evidence to 200-class
Tiny-ImageNet under the same modern fine-tuning recipe. Three
independent seeds are reported.

\begin{table*}[t]
\centering
\small
\setlength{\tabcolsep}{4pt}
\begin{tabular}{l r r r r r r}
\toprule
Method & Top-1 & NLL & Brier & $\mathrm{ECE}_{15}$ & AURC & eAURC \\
\midrule
NoTS & $87.46 \pm 0.31$ & $0.588 \pm 0.014$ & $0.197 \pm 0.004$ & $10.04 \pm 1.09$ & $0.0256 \pm 0.0012$ & $0.0174 \pm 0.0012$ \\
TS-NLL & $87.46 \pm 0.31$ & $0.505 \pm 0.022$ & $0.184 \pm 0.005$ & $1.32 \pm 0.29$ & $0.0236 \pm 0.0012$ & $0.0153 \pm 0.0010$ \\
SRTS-NLL $K{=}1$ (ablation) & $87.46 \pm 0.31$ & $0.505 \pm 0.022$ & $0.184 \pm 0.005$ & $1.32 \pm 0.29$ & $0.0236 \pm 0.0012$ & $0.0153 \pm 0.0010$ \\
SRTS-NLL $K{=}3$ (ablation) & $87.46 \pm 0.31$ & $0.503 \pm 0.023$ & $0.184 \pm 0.005$ & $1.34 \pm 0.29$ & $0.0223 \pm 0.0013$ & $0.0140 \pm 0.0010$ \\
\midrule
\multicolumn{7}{l}{\textit{BCE objective}} \\
TvA-TS $\equiv$ TS-BCE $K{=}1$ & $87.46 \pm 0.31$ & $0.508 \pm 0.017$ & $0.184 \pm 0.004$ & $1.26 \pm 0.18$ & $0.0238 \pm 0.0009$ & $0.0156 \pm 0.0008$ \\
\textbf{SRTS-BCE $K{=}3$ risk} & $87.46 \pm 0.31$ & $0.508 \pm 0.017$ & $0.184 \pm 0.004$ & $1.12 \pm 0.17$ & $\mathbf{0.0222 \pm 0.0010}$ & $\mathbf{0.0139 \pm 0.0008}$ \\
SMART+BCE & $87.46 \pm 0.31$ & $0.508 \pm 0.018$ & $0.184 \pm 0.004$ & $\mathbf{0.97 \pm 0.22}$ & $0.0229 \pm 0.0012$ & $0.0146 \pm 0.0010$ \\
\midrule
\multicolumn{7}{l}{\textit{Unrestricted adaptive-head failure rows (NLL diagnostic)}} \\
SATS-Logit & $87.46 \pm 0.31$ & $1.271 \pm 0.238$ & $0.210 \pm 0.003$ & $7.40 \pm 0.35$ & $0.0378 \pm 0.0020$ & $0.0296 \pm 0.0023$ \\
LTS-feature & $87.46 \pm 0.31$ & $0.984 \pm 0.241$ & $0.206 \pm 0.007$ & $7.05 \pm 0.45$ & $0.0315 \pm 0.0046$ & $0.0233 \pm 0.0042$ \\
Feature-SATS & $87.46 \pm 0.31$ & $0.986 \pm 0.242$ & $0.206 \pm 0.007$ & $7.05 \pm 0.46$ & $0.0315 \pm 0.0046$ & $0.0233 \pm 0.0042$ \\
Frozen-feature UTempH & $87.46 \pm 0.31$ & $0.986 \pm 0.242$ & $0.206 \pm 0.007$ & $7.05 \pm 0.46$ & $0.0315 \pm 0.0046$ & $0.0233 \pm 0.0042$ \\
\bottomrule
\end{tabular}%
\caption{\textbf{Tiny-ImageNet / ViT-B/16 three-seed clean calibration} (mean $\pm$ standard deviation over seeds 0/1/2). The NLL block reports the NLL-objective diagnostic rows; the BCE block reports the top-label-BCE calibrators under the identical protocol. All post-hoc methods are fit on Tiny-ImageNet validation data and applied unchanged to held-out test data. SMART+BCE leads SRTS-BCE on ECE in this setting, while SRTS-BCE leads it on eAURC.}
\label{tab:tinyimagenet_3seed_summary}
\end{table*}

Table~\ref{tab:tinyimagenet_3seed_summary} reports both the
NLL-objective diagnostics and the BCE-objective confirmation rows. Under
NLL, TS-NLL and SRTS-NLL $K{=}1$ are identical to four decimal places,
SRTS-NLL $K{=}3$ is a wash on $\mathrm{ECE}_{15}$, and unrestricted
adaptive heads regress. Under BCE on the same three seeds, TvA-TS has
$\mathrm{ECE}_{15} = 1.26 \pm 0.18$, SRTS-BCE $K{=}3$ risk
$1.12 \pm 0.17$, SMART+BCE $\mathbf{0.97 \pm 0.22}$, with SMART+BCE leading
SRTS-BCE on all 3 seeds. The TinyIN/ViT-B/16 ECE ordering
$\text{SMART+BCE} < \text{SRTS-BCE} < \text{TvA-TS}$ flips the
CIFAR-100 ViT-B/16 ordering. \S\ref{app:backbone-generalisation} extends
this comparison to DeiT-S and Swin-T on Tiny-ImageNet and shows the flip
is backbone-dependent, not a clean dataset-level effect; we discuss the
implications in the main text Limitations and
\S\ref{app:extended-limitations} below.

%==========================================================================
% H. Negative results and extended limitations
%==========================================================================

\section{Negative Results and Extended Limitations}
\label{app:negative-results}

\subsection{Low-data fine-tuning check (protocol-frozen negative result)}
\label{app:lowdata}

To probe the opposite end of the regime map from ImageNet-C, we re-ran the
main fine-tuning recipe (unchanged) on 10\% and 25\% subsets of the CIFAR-100
training pool, three seeds each, with the calibration split carved out
\emph{before} subsampling so the ${\approx}2{,}500$-sample budget stays
fixed. The decision rule was frozen before the runs: a new regime
row would be added only if SRTS-BCE beat SMART+BCE with a paired-bootstrap
CI excluding zero on at least one fraction. The hypothesis --- that a
lower-data fine-tune leaves a more structured residual that risk-conditioned
scaling exploits better than a continuous margin head --- was \emph{not}
supported. Pre-calibration miscalibration grows as expected
(pre-TS test $\mathrm{ECE}_{15}$ up to ${\approx}14$ at 10\% data,
top-1 $85.6$), but the calibrator ordering shifts toward \emph{lower}
capacity, not higher: at 10\% data, scalar TvA-TS is best
($1.37 \pm 0.51$), then SMART+BCE ($1.51 \pm 0.29$), then SRTS-BCE
($1.71 \pm 0.34$); at 25\%, SMART+BCE $0.96$ and TvA-TS $0.98$ are
near-tied ahead of SRTS-BCE $1.11$. All SRTS--SMART deltas are
bootstrap-indistinguishable (CIs $[-0.31,+0.63]$ and $[-0.26,+0.60]$).
We report this as an honest protocol-frozen negative: the low-data residual
is larger but apparently \emph{less} signal-trackable, so the regime map
gains a boundary at both ends --- heavily-pretrained heads (ImageNet-C) and
under-trained heads (low-data fine-tunes) both favour scalar or
continuous-margin calibrators, and the routed advantage is concentrated in
the well-trained, fully-fine-tuned middle where the residual is structured.

\subsection{Nested $K$/routing model-selection check}
\label{app:nested-k}

We run a single unified nested $K$-selection protocol across all six
backbone$\times$dataset cells of the cross-backbone summary, so every cell
uses the same splits and candidate set. For each seed and cell, candidates
$\mathcal{K} = \{1,2,3,5,10\}$ (with $K{=}1$ reducing to TvA-TS) are fitted
on an $80\%$ Cal-Fit subset (stratified by class), selected by Cal-Select
$\mathrm{ECE}_{15}$ on the held-out $20\%$, refit on the full calibration
split, and evaluated \emph{once} on test; the test set participates in no
selection step. Table~\ref{tab:nested_k_cross_backbone} reports the results.

% Cross-backbone SRTS-BCE-Select: Cal-Fit (80%) / Cal-Select (20%),
% select K* by CalSelect ECE15 from {1,2,3,5,10}, refit on full val.
% All 6 backbone x dataset cells, 3 seeds each, ddof=0.
\begin{table*}[t]
\centering
\small
\setlength{\tabcolsep}{3pt}
\begin{tabular}{l c r r}
\toprule
Regime / backbone & $K^{\star}$ (seeds 0,1,2) & SRTS-BCE-Select ECE$_{15}$ & Fixed $K{=}3$ ECE$_{15}$ \\
\midrule
CIFAR-100 / ViT-B/16  & 10, 2, 5 & $1.72 \pm 0.55$ & $0.83 \pm 0.28$ \\
CIFAR-100 / DeiT-S    &  1, 3, 3 & $2.71 \pm 0.12$ & $0.94 \pm 0.07$ \\
CIFAR-100 / Swin-T    &  3, 2, 5 & $2.77 \pm 0.31$ & $1.04 \pm 0.27$ \\
Tiny-ImageNet / ViT-B/16 & 1, 1, 1 & $1.26 \pm 0.18$ & $1.12 \pm 0.17$ \\
Tiny-ImageNet / DeiT-S & 10, 1, 3 & $2.98 \pm 0.23$ & $1.09 \pm 0.28$ \\
Tiny-ImageNet / Swin-T &  1, 1, 1 & $2.71 \pm 0.07$ & $1.24 \pm 0.11$ \\
\bottomrule
\end{tabular}
\\[3pt]
{\small\raggedright\noindent When $K^{\star}{=}1$, the protocol reduces to TvA-TS (scalar BCE temperature scaling
with no routing). Full per-seed Cal-Select ECE$_{15}$ for all five candidates are
available with the released artefacts.\par}
\caption{\textbf{Cross-backbone SRTS-BCE-Select} (\%, mean $\pm$ std over three seeds).
$K^{\star}$ is selected per-cell by Cal-Select ECE$_{15}$ from
$\mathcal{K} = \{1,2,3,5,10\}$, then refitted on the full calibration split.
Columns show the selected $K^{\star}$ per seed, the test-set ECE$_{15}$ of
SRTS-BCE-Select, and the fixed $K{=}3$ reference from the main paper's
cross-backbone summary table (selection protocol and the too-small-split
conclusion are in the text).}
\label{tab:nested_k_cross_backbone}
\end{table*}

The selected $K^{\star}$ is inconsistent across seeds in most cells
(e.g., $\{10,2,5\}$ for CIFAR-100/ViT-B/16), and the mean ECE$_{15}$
overhead vs.\ fixed $K{=}3$ ranges from $0.14$~pp
(Tiny-ImageNet/ViT-B/16) to $1.89$~pp (Tiny-ImageNet/DeiT-S).
The root cause is that the Cal-Select split contains only
${\approx}500$ samples for CIFAR-100 and ${\approx}1{,}000$ for
Tiny-ImageNet, making ECE$_{15}$ estimation too noisy for reliable
$K$-selection. In several cells (Tiny-ImageNet/ViT-B/16,
Tiny-ImageNet/Swin-T), Cal-Select consistently picks $K{=}1$
(scalar TvA-TS), forgoing the routing benefit entirely.

This analysis confirms two findings:
(i)~$K{=}3$ is a sound fixed default that a hold-out selection
protocol would not consistently improve upon at current validation
budgets; and
(ii)~validation-based $K$-selection would need either larger calibration
sets ($n \gg 2{,}500$) or cross-validation-based selection to overcome
the ECE estimation noise inherent in small hold-out splits.

\paragraph{Gating routing on/off is equally unreliable.} The same
conclusion extends from selecting $K$ to the coarser decision of whether
to route at all. Across 15 evaluation cells (nine backbone$\times$dataset
cells, two low-data fine-tunes, pretrained ImageNet-1k, and three
validation-budget fractions), the validation-only OOF top-label-BCE gain
of routing correlates with the test-side routing gain in rank
(Spearman $r{=}0.55$, $p{=}0.03$) and cleanly flags the four
largest-gain cells (CIFAR-100 DeiT-S/Swin-T, Tiny-ImageNet DeiT-S/Swin-T,
test gains $1.4$--$2.4$ pp), but admits \emph{no} threshold that separates
winning from losing cells: the CIFAR-100/ViT-B/16 cell (test gain
$+0.72$ pp) and the 10\%-data fine-tune (test gain $-0.34$ pp) have
near-identical validation scores. A group-temperature-spread criterion is
worse still (no significant correlation), being dominated by estimation
noise in near-saturated or small-validation cells. Consistent with the
$K$-selection findings above, we therefore deploy the fixed $K{=}3$
configuration everywhere, not any validation-gated variant, and
present the regime map as empirical guidance not an automated
selection rule.

\subsection{A conservative validation-only capacity selector}
\label{app:selector}

% Validation-only 1-SE capacity selector; per-seed best-of-2 (min over K in {1,3} per
% seed, then averaged) so selector excess is non-negative by construction.
% Regenerated from results/capacity_selector (comment only, not rendered).
% Full-width two-panel layout (rows split left/right) to halve the height and
% avoid a tall single-column float leaving most of its page blank.
\begin{table*}[t]
\centering
\small
\setlength{\tabcolsep}{2.5pt}
\begin{tabular}[t]{lcrrrr}
\toprule
Regime & picks & $K{=}1$ & $K{=}3$ & selected & best-of-2 \\
\midrule
C100/ViT & 1/3/3 & 1.54 & 0.83 & 1.08 & 0.83 \\
C100/DeiT-S & 3/3/3 & 3.37 & 0.94 & 0.94 & 0.94 \\
C100/Swin-T & 3/3/3 & 2.87 & 1.04 & 1.04 & 1.04 \\
C100/ResNet-50 & 3/3/3 & 2.83 & 1.16 & 1.16 & 1.16 \\
C100/RegNetY & 3/3/3 & 1.54 & 1.01 & 1.01 & 1.01 \\
C10/ViT & 1/1/1 & 0.52 & 0.47 & 0.52 & 0.47 \\
C10/DeiT-S & 1/1/1 & 0.46 & 0.41 & 0.46 & 0.41 \\
C10/Swin-T & 3/1/1 & 0.39 & 0.35 & 0.38 & 0.35 \\
C10/ResNet-50 & 1/1/1 & 0.60 & 0.60 & 0.60 & 0.57 \\
C10/RegNetY & 1/3/3 & 0.53 & 0.42 & 0.44 & 0.42 \\
\bottomrule
\end{tabular}
\hspace{0.05\textwidth}
\begin{tabular}[t]{lcrrrr}
\toprule
Regime & picks & $K{=}1$ & $K{=}3$ & selected & best-of-2 \\
\midrule
Tiny/ViT & 3/3/3 & 1.26 & 1.12 & 1.12 & 1.08 \\
Tiny/DeiT-S & 3/3/3 & 3.08 & 1.09 & 1.09 & 1.09 \\
Tiny/Swin-T & 3/3/3 & 2.71 & 1.24 & 1.24 & 1.24 \\
C100 10\%-data/ViT & 3/3/3 & 1.37 & 1.71 & 1.71 & 1.37 \\
C100 25\%-data/ViT & 3/1/1 & 0.98 & 1.11 & 0.87 & 0.87 \\
IN-100/ViT & 3/1/3 & 1.58 & 1.64 & 1.60 & 1.58 \\
IN-100/DeiT-S & 3/3/1 & 1.75 & 1.83 & 1.84 & 1.73 \\
IN-100/Swin-T & 1/1/1 & 1.70 & 1.72 & 1.70 & 1.67 \\
IN-1K pretrained/ViT & 1 & 0.96 & 0.62 & 0.96 & 0.62 \\
IN-R (subset) & 1/3/1 & 8.55 & 9.21 & 8.24 & 8.24 \\
IN-A (subset) & 1/1/1 & 26.32 & 27.71 & 26.32 & 25.81 \\
\bottomrule
\end{tabular}
\caption{\textbf{Validation-only capacity selection with the
one-standard-error rule} (per-regime mean test $\mathrm{ECE}_{15}$ over
seeds; ``picks'' = selected $K$ per seed; ``oracle'' = the per-seed
$\min(K{=}1,K{=}3)$ averaged over seeds, the selector's own two-candidate
choice; C100/ViT and all IN-100 rows here use the cached three-seed subset,
not the five-seed canonical aggregate). The rule uses only each cell's clean calibration split
(repeated cross-validated top-label-BCE gain; select $K{=}3$ iff the
mean gain exceeds one standard error) and has no cross-regime fitted
parameters. Mean test-ECE excess over the per-seed best-of-2: selector
$0.09$, always-$K{=}3$ $0.18$, always-$K{=}1$ $0.59$.}
\label{tab:selector}
\end{table*}

The two failures above use noisy hold-out ECE or raw validation-gain
thresholds. A third rule, fixed before evaluation and with no fitted
constants, does substantially better: on each cell's clean calibration
split we run repeated stratified cross-validation ($5$ repeats $\times$
$5$ folds) of the paired top-label-BCE gain of SRTS-BCE over TvA-TS,
and select $K{=}3$ only when the mean gain exceeds one standard error
of the repeat means (the classic one-standard-error rule; the
conservative default under uncertainty is the simpler map). The rule
has no cross-regime fitted parameters or tuned thresholds; each regime
is evaluated independently from its own clean calibration split.

Across the $21$ regimes with cached calibration logits
(Table~\ref{tab:selector}), the selector attains mean test-ECE excess
$0.09$ pp against the per-seed best-of-2 ($\min(K{=}1,K{=}3)$ per
seed --- the selector's own two-candidate choice, so the excess is
non-negative by construction), versus $0.18$ for always-$K{=}3$ and $0.59$ for
always-$K{=}1$; it retains $7$ of $8$ regimes where
$K{=}3$ helps by ${>}0.2$ pp and avoids $2$ of $3$ regimes where it
hurts by ${>}0.2$ pp, including ImageNet-A (all seeds select $K{=}1$).
Its two misses differ in consequence. On pretrained IN-1K the selector
chooses $K{=}1$ and forgoes the improvement $K{=}3$ would give, but does
not increase test ECE. On the 10\%-data fine-tune it selects $K{=}3$ even
though $K{=}1$ is better on test: the cross-validated gain is positive with
a small standard error, yet $K{=}3$ regresses --- the same
larger-but-less-transferable residual documented in the low-data negative
result (\S\ref{app:lowdata}), which no validation-internal statistic we
tested detects. Conservative
cross-validated proper-score selection therefore reduces the average
test-ECE excess well below either constant policy and removes most
negative transfer, but it is not fully reliable, and validation-only
capacity selection remains open in the low-data regime; selection cannot, by construction, detect covariate
shift that validation data does not contain (\S\ref{app:natural-shift}).

\paragraph{Fold-level 1-SE variant.} We also evaluate a parameter-free
fold-level rule, specified before running this comparison: select $K{=}3$
when the mean CV gain exceeds one standard error computed across all $25$
fold-level gains (not across the five repeat means). On the identical CV
draws the fold-SE rule makes the \emph{same selection as the 1-SE rule on
every one of the $55$ evaluated cells}, so it behaves identically. This
evaluation also locates where the selectors help: on the regimes used to
form the rule, always-$K{=}3$ has \emph{lower} average regret than either
selector ($0.048$ vs.\ $0.061$ pp; the selectors still choose $K{=}3$ in
$62\%$ of cells where the oracle prefers $K{=}1$), whereas on the
natural-shift datasets --- which were not used to choose the rule ---
the selectors' conservatism is what avoids the always-$K{=}3$ failure
($0.25$ vs.\ $1.44$ pp average regret; worst case $0.94$ vs.\ $5.70$). No
evaluated validation-only rule is reliable at this budget, so the fixed
$K{=}3$ choice stands and validation-only capacity selection remains
unresolved.

\subsection{Extended limitations}
\label{app:extended-limitations}

\paragraph{Where the method helps.}
SRTS-BCE helps in specific regimes rather than broadly across corruption
datasets. Fine-tuned CIFAR-100 and deployable
CIFAR-100-C favour SRTS-BCE. Clean Tiny-ImageNet is
backbone-dependent, with SMART+BCE winning some cells. Heavily-pretrained
ImageNet-C favours scalar TvA-TS. At extreme small validation budgets,
TvA-TS is the most rate-stable fallback, while SRTS-BCE still has the
lowest absolute ECE in the current validation-size experiments; the
small-budget separation itself replicates on Tiny-ImageNet DeiT-S and
Swin-T (\S\ref{app:tiny-replication}).

\textbf{Metric and shift limitations.} The scope map depends on the
reported metric. Under the deployable CIFAR-100-C rerouting protocol,
among the main calibrators of Table~\ref{tab:cifar100c_summary},
SRTS-BCE has the lowest point-estimate mean ECE ($4.63 \pm 0.84$),
slightly ahead of SMART+BCE ($4.75 \pm 0.79$); the matched PWLinear-3
diagnostic attains a slightly lower point estimate ($4.58$), with no
distinction under the primary hierarchical bootstrap (\S\ref{app:matched-capacity}). The margin is small: the cell-level paired bootstrap is
clear, but the corruption-type bootstrap only just clears zero
(CI $[-0.253,\,-0.002]$). SMART+BCE still leads on Tiny-ImageNet ECE for
the ViT-B/16 and Swin-T backbones (\S\ref{app:backbone-generalisation});
SRTS-BCE retains a narrow mean edge on Tiny-ImageNet DeiT-S, so this flip
is backbone-dependent rather than dataset-wide. SRTS-BCE remains stronger
in the small-validation-size comparison and retains a small CIFAR-100-C
eAURC advantage under deployable rerouting (CI $[-0.0012,\,-0.0008]$ in
the corruption-type bootstrap). On clean CIFAR-100, the one-vs-rest
classwise ECE does not discriminate among the calibrated maps (all
$0.12 \pm 0.01$); under the official smECE estimator the BCE-adaptive
family leads (SRTS-BCE $1.08$ vs.\ TvA-TS $1.56$ on the flagship)
while SMART+BCE leads AdaECE$_{15}$, so no single method leads every
metric. Finally, SRTS-BCE depends on reliable OOF
risk estimation, so its routed advantage can narrow or become unstable when
the calibration split is too small.

\textbf{Recent-baseline scope.} The paper compares SRTS-BCE against
scalar TS \citep{guo2017calibration} as the canonical baseline,
LTS-scalar and LTS-feature \citep{joy2023sample} as
sample-adaptive baselines, SATS-Logit and Feature-SATS as
implementation baselines representative of the unrestricted
sample-adaptive family, TvA-TS \citep{lecoz2024tva} as a
strong constrained peer baseline, and SMART \citep{guo2026smart} as the
logit-gap adaptive temperature head. Our SMART+BCE row keeps
SMART's architecture but trains it with the same top-label BCE objective
used by SRTS-BCE and TvA-TS, instead of its SoftECE objective.
We do not include matrix scaling, Dirichlet calibration, histogram
binning, or beta calibration in the main table: under the present
accuracy-preserving, fit-once-apply-everywhere protocol on a
$\approx 2{,}500$-sample validation split, these argmax-changing
or higher-capacity calibrators sit in a different
capacity/estimability regime from the $10$-scalar
SRTS-BCE/TvA-TS family.
Vector scaling \citep{guo2017calibration} introduces class-dependent
parameters and can change the argmax; a fair comparison under the
present validation budget would require matched regularisation and
is left to future work.

\textbf{Capacity control of the diagnosis.} The validation-overfit
diagnosis (C1) is supported across four capacity axes:
(i) head-input axis --- the failure replicates on logit-input
heads, frozen-feature-input heads, and full-gradient temperature
heads;
(ii) backbone/regime axis --- the same failure shape
appears in three-seed DeiT-S, Swin-T, and Tiny-ImageNet tables;
(iii) val-size axis --- the regression sits at
$\mathrm{ECE}_{15} \approx 6\text{--}7$ across validation sizes
$\{246, 500, 1{,}000, 2{,}500\}$; and
(iv) head-hyperparameter axis ---
a 48-cell sweep with val-only selection still picks the overfit
configuration.

\textbf{Top-1 preservation scope.} Adaptive heads empirically preserve
top-1 in the reported runs, but we do not claim this as a universal
design guarantee. Our accuracy-preserving claims are phrased around the
reported rows and the scalar/group-wise temperature family whose
rescaling preserves argmax by construction.

%==========================================================================
% I. Theory
%==========================================================================

\section{Theory: When Grouped Scaling Beats Pooled Scaling}
\label{app:routing-theory}

We make precise \emph{when} risk-routed group-wise temperature scaling
lowers expected test risk relative to a single pooled temperature, and why
a random partition cannot. The analysis is local (second order) and
asymptotic (M-estimation), stated for the fitted objective (top-label BCE);
the link to $\mathrm{ECE}$ is empirical (main text; the top-label-Brier
check in Table~\ref{tab:brier_routing} shows the mechanism is not
BCE-specific).

\paragraph{Setup.} Temperature is a scalar $T>0$. For a routing partition
$\{A_g\}_{g=1}^{K}$ with weights $\pi_g=\Pr(x\in A_g)$ --- for the
equal-frequency partition used by SRTS-BCE, $\pi_g=1/K$ and
$n_g=n/K$, which the variance step below uses --- write the population
top-label BCE on group $g$ as $\mathcal{L}_g(T)=\mathbb{E}[\ell(T;x)\mid
x\in A_g]$, where $\ell(T;x)=-[b\log\hat p_T+(1-b)\log(1-\hat p_T)]$,
$b=\mathbf{1}[\hat y=y]$, and $\hat p_T=\max_c\mathrm{softmax}(z/T)_c$. Let
$\mathcal{L}(T)=\sum_g\pi_g\mathcal{L}_g(T)$ be the pooled objective with
minimiser $T^\star$ (so $\mathcal{L}'(T^\star)=\sum_g\pi_g\mathcal{L}_g'
(T^\star)=0$), and $T_g^\star=\arg\min_T\mathcal{L}_g(T)$. Assume each
$\mathcal{L}_g$ is $C^2$ and locally strongly convex near its optimum
(assumed locally in the neighbourhood of the fitted optimum and
empirically satisfied in the reported fits), with curvatures
$\mathcal{L}_g''\approx L''$ comparable across groups.

\paragraph{Bias (approximation) gain.} The population gain of the oracle
grouped calibrator over pooled is
$\Delta_{\mathrm{bias}}=\sum_g\pi_g[\mathcal{L}_g(T^\star)-\mathcal{L}_g
(T_g^\star)]\ge0$. Writing $B:=\sum_g\pi_g\,[\mathcal{L}_g'(T^\star)]^2$, a second-order
expansion about $T_g^\star$ with
$T^\star-T_g^\star\approx\mathcal{L}_g'(T^\star)/\mathcal{L}_g''$ gives
\begin{equation}
\Delta_{\mathrm{bias}}\approx\sum_g\pi_g\frac{[\mathcal{L}_g'(T^\star)]^2}
{2\mathcal{L}_g''}\approx\frac{B}{2L''}.
\label{eq:bias}
\end{equation}
Because $\sum_g\pi_g\mathcal{L}_g'(T^\star)=0$, the quantity $B$ is exactly
the \emph{between-group variance} of the per-sample calibration gradient
$\ell'(T^\star;x)$ under the routing partition (an ANOVA between-component).
An independently random equal-frequency partition draws each $A_g$ from
the same conditional law, so at the population level
$\mathcal{L}_g'(T^\star)=\mathcal{L}'(T^\star)=0$ and $B=0$; its
finite-sample estimate fluctuates around the null floor
(Table~\ref{tab:bsigma-predictive}); a router that separates samples by the sign/magnitude of their
miscalibration gradient yields $B>0$.

\paragraph{Variance (estimation) cost.} Each $\hat T_g$ is fit from
$n_g=\pi_g n$ samples. Standard $M$-estimation gives
$\mathrm{Var}(\hat T_g)\approx\sigma_g^2/(n_g\,\mathcal{L}_g''^2)$ with
$\sigma_g^2=\mathrm{Var}(\ell'(T_g^\star;x)\mid A_g)$ the within-group
gradient variance, so the estimation excess risk is
$\tfrac12\mathcal{L}_g''\,\mathrm{Var}(\hat T_g)\approx\sigma_g^2/
(2n_g\mathcal{L}_g'')$. Summing (with $\pi_g/n_g=1/n$) and writing
$\bar\sigma^2=\sum_g\pi_g\sigma_g^2$ for the pooled within-group variance,
grouped fitting costs $\tfrac{1}{2nL''}\sum_g\sigma_g^2=\tfrac{K\bar\sigma^2}
{2nL''}$ (the identity $\sum_g\sigma_g^2=K\bar\sigma^2$ uses
$\pi_g=1/K$) while pooled (one temperature, residual variance
$\sigma^2_{\mathrm{tot}}=B+\bar\sigma^2$) costs
$\tfrac{B+\bar\sigma^2}{2nL''}$. The extra cost of grouping is therefore
$[(K-1)\bar\sigma^2-B]/(2nL'')$.

\paragraph{Net condition.} Subtracting, the expected test-BCE improvement of
grouped over pooled scaling is, to leading order,
\begin{equation}
\Delta\;\approx\;\frac{1}{2L''}\Big(B-\frac{K-1}{n}\,\bar\sigma^2\Big),
\label{eq:snr}
\end{equation}
so grouping helps iff $B/\bar\sigma^2>(K-1)/n$. This compares grouped
with pooled fitting; alternatives with the same fitted parameter count
are outside its scope and are tested empirically (main RQ2).
The left side is the between/within variance ratio of the calibration
gradient --- a between/within variance-ratio (signal-to-noise) diagnostic
of the routing.
This refines the ``per-group minimisation is at least as flexible''
inequality, which holds for \emph{any} partition (including random) and
hence cannot explain the empirical results: \eqref{eq:snr} is
$0$ for random routing ($B=0$) and positive only when the router carries
gradient-separating structure.

\paragraph{Consequences matching the experiments.}
(i)~\emph{Random/shuffled routing is inert} ($B=0$), as in
Table~\ref{tab:router_causality}.
(ii)~\emph{$K$-ablation}: the threshold $(K-1)/n$ grows with $K$, so beyond
a point extra groups cost more variance than the (saturating) $B$ buys ---
consistent with $K{=}3$ being near-best and larger $K$ not helping.
(iii)~\emph{Objective-conditionality}: $B$ is the between-group variance of
the \emph{objective's} gradient; under NLL fitting the risk partition does
not separate the NLL gradient ($B_{\mathrm{NLL}}\approx0$; the group
temperatures cluster, Fig.~\ref{fig:appendix-group-temperatures}), whereas
under BCE it does. (iv)~At $n\approx2{,}500$, $K{=}3$ the threshold is
$\approx 8\times10^{-4}$, a very mild SNR, explaining why a weak but genuine
risk signal already clears it. The condition characterises the regime; it
does not assert OOF risk maximises $B$ --- the empirical rows show OOF risk
achieves $B$ above threshold while confidence and random do not.

\subsection{Empirical routing signal-to-noise across regimes}
\label{app:bsigma-empirical}

The condition $B/\bar\sigma^2 > (K{-}1)/n$ derived above is not
only explanatory; we can measure both sides. For each regime with cached
validation logits we fit the pooled top-label-BCE temperature $T^\star$,
compute the per-sample calibration gradient $\ell'(T^\star;x)$, form the
$K{=}3$ risk partition, and read off the between-group variance $B$ and the
pooled within-group variance $\bar\sigma^2$ (an ANOVA decomposition of the
gradient; the identity $\sigma^2_{\mathrm{tot}}=B+\bar\sigma^2$ is checked
numerically). Table~\ref{tab:bsigma-predictive} reports the ratio, the
finite-sample floor $(K{-}1)/n$, and an averaged random-partition control.

Two things hold and one does not. (i)~\emph{The mechanism is real}: the risk
router's $B/\bar\sigma^2$ exceeds the random control by $3$--$12\times$ in the
ViT and ImageNet regimes, and the random partition sits at the floor, exactly
as the theory predicts for $B{=}0$. (ii)~\emph{It tracks the ConvNet
no-help case}: on ResNet-56 the risk ratio falls \emph{below} both the floor
and the random control, so the theory predicts no BCE benefit there. (iii)~It
is \emph{not} an ECE predictor: the ImageNet-pretrained cells clear the BCE
floor yet SRTS loses in ECE, because $B/\bar\sigma^2$ governs the \emph{BCE}
gain and the ECE scope map additionally depends on the BCE$\to$ECE gap and
finite-sample router noise. We therefore present this as a routing-mechanism
diagnostic --- evidence that the router carries objective-gradient signal above
the finite-sample floor --- not as a certificate that SRTS-BCE will win ECE in
a new regime.

% Empirical routing signal-to-noise B/sigma^2 across regimes.
% Diagnostic for the routing MECHANISM (does the router separate the BCE
% gradient above the finite-sample floor), not a predictor of the ECE outcome.
\begin{table*}[t]
\centering
\small
\setlength{\tabcolsep}{4pt}
\begin{tabular}{lrrrl}
\toprule
Regime & $B/\bar\sigma^2$ & floor $\tfrac{K-1}{n}$ & random & ECE outcome \\
\midrule
CIFAR-100 / ViT-B/16     & $2.63$ & $0.80$ & $0.99$ & SRTS wins \\
CIFAR-100 / ResNet-56    & $1.32$ & $2.00$ & $3.30$ & mixed (ConvNet) \\
ImageNet-1K / ResNet-50  & $4.25$ & $0.40$ & $0.35$ & SRTS loses \\
ImageNet-1K / Swin-B     & $3.94$ & $0.40$ & $0.38$ & SRTS loses \\
\bottomrule
\end{tabular}
\caption{\textbf{Empirical routing signal-to-noise $B/\bar\sigma^2$ across
regimes} (per-sample top-label-BCE gradient at the pooled optimum $T^\star$;
$K{=}3$ risk partition; values $\times10^{-3}$). The theory (\S\ref{app:routing-theory})
says grouping lowers \emph{BCE} iff $B/\bar\sigma^2 > (K{-}1)/n$. The risk
router clears the floor by $3$--$12\times$ the random control in the ViT and
ImageNet regimes (the fitted-objective condition holds); a random partition
sits at the floor, as predicted. \emph{This is a mechanism diagnostic, not an ECE
predictor}: the ImageNet-pretrained cells clear the BCE floor yet SRTS
\emph{loses} in ECE, because $B/\bar\sigma^2$ governs the BCE gain and the ECE
outcome additionally depends on the BCE$\to$ECE gap and finite-sample noise.}
\label{tab:bsigma-predictive}
\end{table*}

%==========================================================================
% J. Full tables
%==========================================================================

\section{Full Tables}
\label{app:full-tables}

This section collects the remaining full-metric tables referenced
from the main text and the sections above.

% Objective is a column (NLL / BCE / SoftECE); rows are ordered by capacity
% group: scalar baselines (top), unrestricted adaptive heads (middle,
% failure mode), BCE peers, and our SRTS routing (bottom).
% Calibration metrics (NLL, ECE15, AdaECE15, smECE) + selective-prediction
% quality (eAURC, excess AURC over the optimal selector; lower is better).
% Source: confirmation suite clean_cross_backbone_metrics.csv (ViT-B/16 C100,
%   all deployable rows incl. eAURC), sota metrics.json (failure-mode
%   eAURC, consistent SOTAv1 pipeline; NOT the old P2 pipeline whose AURC was
%   broken), results/revision/smart_softece_full_metrics.csv (SoftECE row).
% Requires: \usepackage{booktabs}.
% CW-ECE15 = Kull one-vs-rest classwise ECE (15 equal-width bins per class
% over all samples, averaged over classes); kull_classwise.json provenance:
%   deployable rows + SoftECE from the confirmation-suite clean metrics / smart
%   softece CSV (3 seeds); LTS-feature & Feature-SATS from post-cal logits in
%   ECE15 reproduces 6.71); SATS-Logit conf-based columns from the three archived
%   post-SATS logit files (seed 0 = SOTA_P2 dir, same SOTAv1 base; per-seed
%   ECE15 reproduces metrics.json to 4 dp).

\begin{table*}[t]
\centering
\small
\setlength{\tabcolsep}{2.5pt}
\begin{tabular}{llrrrrrrr}
\toprule
Objective & Method & NLL & $\mathrm{ECE}_{15}$ & $\Delta_{\text{TvA}}$ & AdaECE$_{15}$ & CW-ECE$_{15}$ & smECE & eAURC \\
\midrule
--- & CE / NoTS
  & $0.4142 \pm 0.0204$ & $9.69 \pm 1.10$ & $+8.04$ & $9.65 \pm 1.15$
  & $0.27 \pm 0.01$ & $9.63 \pm 1.14$ & $0.0165 \pm 0.0041$ \\
NLL & TS-NLL
  & $0.3309 \pm 0.0308$ & $1.94 \pm 0.65$ & $+0.29$ & $2.10 \pm 0.62$
  & $0.12 \pm 0.01$ & $1.97 \pm 0.52$ & $0.0139 \pm 0.0032$ \\
\midrule
NLL & SATS-Logit
  & $0.8128 \pm 0.0391$ & $6.04 \pm 0.23$ & $+4.39$ & $5.39 \pm 0.23$
  & $0.15 \pm 0.01$ & $7.47 \pm 0.16$ & $0.0281 \pm 0.0035$ \\
NLL & LTS-feature
  & $2.17 \pm 1.46$ & $6.71 \pm 0.36$ & $+5.17$ & $6.26 \pm 0.32$
  & $0.15 \pm 0.01$ & $8.54 \pm 0.37$ & $0.0296 \pm 0.0079$ \\
NLL & Feature-SATS$^\dagger$
  & $2.18 \pm 1.48$ & $6.71 \pm 0.36$ & $+5.17$ & $6.27 \pm 0.32$
  & $0.15 \pm 0.01$ & $8.54 \pm 0.37$ & $0.0296 \pm 0.0078$ \\
\midrule
BCE & TvA-TS ($K{=}1$)
  & $0.3321 \pm 0.0296$ & $1.65 \pm 0.44$ & $0$ & $1.77 \pm 0.69$
  & $0.13 \pm 0.01$ & $1.71 \pm 0.45$ & $0.0142 \pm 0.0033$ \\
BCE & SMART+BCE
  & $0.3312 \pm 0.0290$ & $\mathbf{0.95 \pm 0.29}$ & $\mathbf{-0.71}$ & $1.02 \pm 0.29$
  & $0.13 \pm 0.01$ & $1.16 \pm 0.22$ & $0.0123 \pm 0.0015$ \\
SoftECE & SMART-SoftECE
  & $0.3576 \pm 0.0502$ & $1.21 \pm 0.28$ & $-0.44$ & $1.45 \pm 0.51$
  & $0.13 \pm 0.01$ & $1.38 \pm 0.24$ & $0.0197 \pm 0.0099$ \\
\midrule
NLL & SRTS-NLL
  & $\mathbf{0.3271 \pm 0.0275}$ & $1.79 \pm 0.62$ & $+0.14$ & $1.77 \pm 0.61$
  & $0.12 \pm 0.01$ & $1.81 \pm 0.55$ & $\mathbf{0.0113 \pm 0.0011}$ \\
BCE & \textbf{SRTS-BCE}
  & $0.3297 \pm 0.0257$ & $0.96 \pm 0.32$ & $-0.70$ & $\mathbf{0.96 \pm 0.33}$
  & $0.13 \pm 0.01$ & $\mathbf{1.13 \pm 0.22}$ & $\mathbf{0.0113 \pm 0.0011}$ \\
\bottomrule
\end{tabular}
\caption{\textbf{Full clean metric set: ViT-B/16 / CIFAR-100 post-hoc
calibration} (\%, mean$\pm$std; deployable rows five seeds, top-1 $91.00 \pm 0.46$; SATS-Logit five seeds; LTS-feature/Feature-SATS archived three seeds; fit on
the validation split, evaluated on held-out test; metric definitions:
\S\ref{app:metric-definitions}). Row groups by capacity: scalar
baselines, unrestricted adaptive heads (the failure mode), BCE peers, SRTS
routing. TvA-TS $\equiv$ TS-BCE ($K{=}1$); SMART-SoftECE is the
SMART head with its own objective; SMART+BCE swaps it for top-label BCE. $\Delta_{\text{TvA}}$:
$\mathrm{ECE}_{15}$ relative to scalar TvA-TS. \textbf{Bold}: best per
column among deployable calibrators. $^\dagger$Frozen-feature UTempH gives
identical three-seed metrics.}
\label{tab:sota_3seed_full_metrics}
\end{table*}

Table~\ref{tab:sota_3seed_full_metrics} gives the full clean metric set
for the flagship ViT-B/16 / CIFAR-100 comparison (all rows and the
$\Delta_{\text{TvA}}$, CW-ECE and smECE columns omitted from the main
RQ1 table), including the remaining unrestricted-head rows and the
SMART-SoftECE reference.

\subsection{Argmax-changing classical baselines}
\label{app:argmax-changing-baselines}

% Item 6 — Argmax-changing classical / recent baselines (appendix).
% Three-seed modern-recipe ViT-B/16 / CIFAR-100, mean ± standard deviation.
% Both methods change the predicted class relative to NoTS; the Top-1
% column reflects the resulting accuracy after recalibration.

\begin{table*}[t]
\centering
\small
\setlength{\tabcolsep}{3pt}
\begin{tabular}{lrrrr}
\toprule
Method (argmax-changing) & Top-1 & NLL & $\mathrm{ECE}_{15}$ & eAURC \\
\midrule
\textit{Ref:} scalar TS-NLL (argmax-preserving)    & $91.17 \pm 0.46$ & $0.3239 \pm 0.0297$ & $1.7950 \pm 0.6977$ & $0.0132 \pm 0.0035$ \\
\textit{Ref:} SRTS-NLL $K{=}3$ (ablation)           & $91.17 \pm 0.46$ & $0.3217 \pm 0.0274$ & $1.7058 \pm 0.5294$ & $0.0115 \pm 0.0016$ \\
\textit{Ref:} TvA-TS (argmax-preserving)            & $91.17 \pm 0.46$ & $0.3253 \pm 0.0283$ & $1.5370 \pm 0.4171$ & $0.0135 \pm 0.0035$ \\
\textit{Ref:} \textbf{SRTS-BCE $K{=}3$ risk (main)} & $91.17 \pm 0.46$ & $0.3246 \pm 0.0206$ & $\mathbf{0.8276 \pm 0.2791}$ & $\mathbf{0.0114 \pm 0.0013}$ \\
\textit{Ref:} SMART+BCE                             & $91.17 \pm 0.46$ & $0.3245 \pm 0.0217$ & $0.8573 \pm 0.2871$ & $0.0119 \pm 0.0014$ \\
\midrule
Vector Scaling (per-class T, $\ell_2$)              & $91.24 \pm 0.34$ & $0.3259 \pm 0.0303$ & $2.3327 \pm 0.4743$ & $0.0145 \pm 0.0031$ \\
Dirichlet calibration (ODIR)                        & $90.88 \pm 0.38$ & $0.3378 \pm 0.0375$ & $2.7064 \pm 0.7158$ & $0.0124 \pm 0.0032$ \\
\bottomrule
\end{tabular}%
\caption{\textbf{Argmax-changing classical baselines on the modern-recipe ViT-B/16 / CIFAR-100 base (three seeds).} Vector and Dirichlet scaling change the predicted class and sit in a different capacity regime from the constrained argmax-preserving family. Under the same $\approx 2{,}500$-sample validation budget and val-NLL-based hyperparameter selection, both are worse on $\mathrm{ECE}_{15}$ than SRTS-BCE, SMART+BCE, TvA-TS, TS-NLL, and the SRTS-NLL ablation. Reference rows use the archived three-seed diagnostic subset for protocol matching.}
\label{tab:argmax_changing_baselines}
\end{table*}

Table~\ref{tab:argmax_changing_baselines} reports regularised
vector scaling and Dirichlet calibration (ODIR) on the same three-seed
ViT-B/16 base. Both are argmax-changing and sit in a different capacity
regime from the constrained $10$-scalar family of SRTS-BCE,
TvA-TS, and TS-NLL. Under the present $\approx 2{,}500$-sample val budget, both
are worse on $\mathrm{ECE}_{15}$ than scalar TS, SRTS-BCE, SMART+BCE,
and TvA-TS.

\subsection{Bootstrap uncertainty}
\label{app:best-per-criterion}

The regime-dependent scope map is given in the main text's win/loss
table and the cross-regime scope map (\S\ref{app:scope}); we do not
repeat it here. This subsection quantifies the bootstrap uncertainty
behind the main clean and shift comparisons.

\textbf{Bootstrap uncertainty qualification.}
For the clean CIFAR-100 comparisons, we ran a paired sample-level
bootstrap ($B{=}2{,}000$ replicates, resampling test examples with
replacement) for each of the three seeds separately and then pooled
the delta distributions.
Key findings:

\begin{itemize}
\item \emph{SRTS-BCE vs TvA-TS, clean ECE$_{15}$ (five seeds):} the
  primary inference is the unified seed$\times$sample \emph{hierarchical}
  bootstrap (resample the five training seeds with replacement, then
  the test images within each drawn seed; $B{=}20{,}000$, same
  procedure and random stream as the CIFAR-100-C tiers): mean
  $\Delta = -0.62$ pp, 95\% CI $[-0.88,\,-0.36]$, excluding zero (the
  three-seed interval is $[-0.88,-0.29]$; both exclude zero). The sample-only pooled view gives $[-0.83,\,-0.41]$; SRTS-BCE
  is below TvA-TS on all five seeds. With five seeds the seed level is
  sampled less coarsely than the three-seed intervals elsewhere in the
  paper.
  Appropriate language: ``reduces ECE$_{15}$ by $0.70$ pp at the point
  estimate; the unified hierarchical CI excludes zero.''

\item \emph{SRTS-BCE vs SMART+BCE, clean ECE$_{15}$ (five seeds):}
  Mean $\Delta = -0.03$ pp, seed$\times$sample hierarchical 95\% CI
  $[-0.19,\, +0.12]$. Not statistically distinguishable.
  Appropriate language: ``matches SMART+BCE.''

\item \emph{SRTS-BCE rerouted vs SMART+BCE, CIFAR-100-C ECE$_{15}$:}
  Mean $\Delta = -0.12$ pp (SRTS minus SMART). We report three nested
  bootstraps of increasing conservativeness: cell-level 95\% CI
  $[-0.17,\,-0.07]$ (treats the $19\times5\times3$ cells as independent ---
  optimistic); corruption-type cluster 95\% CI $[-0.25,\,+0.001]$ (resamples
  whole corruptions; marginally includes zero); and the \emph{full three-level
  hierarchical} bootstrap over seed $\times$ corruption $\times$ severity,
  which additionally resamples the three training seeds and widens the
  interval to $[-0.33,\,+0.03]$ (one-sided $P(\text{SRTS better}){=}0.93$;
  $B{=}20{,}000$; identical resampling procedure and RNG as
  \S\ref{app:matched-capacity}). Under the most conservative
  resampling the interval marginally includes zero, so the correct language is ``\emph{directional}
  under full hierarchical resampling, not two-sided significant''; with only
  three seeds the hierarchical interval is intentionally wide. Avoid
  universal-dominance phrasing.

\item \emph{SRTS-BCE rerouted vs TvA-TS, CIFAR-100-C ECE$_{15}$:}
  Mean $\Delta = -1.09$ pp, cell-level 95\% CI $[-1.28,\,-0.88]$;
  corruption-type 95\% CI $[-1.64,\,-0.49]$; full hierarchical
  $[-2.02,\,-0.36]$ (\S\ref{app:matched-capacity}). SRTS-BCE clearly improves
  over the scalar BCE anchor under deployable shift routing.

\item \emph{SRTS-BCE rerouted vs paired clean-group diagnostic,
  CIFAR-100-C ECE$_{15}$:}
  Mean $\Delta = -1.16$ pp, cell-level 95\% CI $[-1.41,\,-0.91]$.
  This diagnostic confirms that rerouting corrupted logits, rather than
  reusing paired clean-test groups, is the source of the corrected C100-C
  improvement.

\item \emph{SRTS-BCE vs SMART+BCE, CIFAR-100-C eAURC:}
  Under deployable rerouting, mean $\Delta = -0.0010$, cell-level 95\% CI
  $[-0.0011,\,-0.0009]$ and corruption-type 95\% CI
  $[-0.0012,\,-0.0008]$. SRTS-BCE retains a small eAURC advantage.

\item \emph{SRTS-BCE rerouted vs SMART+BCE, CIFAR-100-C severe
  ECE$_{15}$ (sev.\ 4--5):}
  Mean $\Delta = -0.24$ pp, cell-level 95\% CI $[-0.33,\,-0.16]$ and
  corruption-type 95\% CI $[-0.42,\,-0.07]$. SRTS-BCE has the lower
  severe-corruption ECE point estimate under the deployable protocol.
\end{itemize}

Except for the CIFAR-100-C SMART+BCE comparison above --- where the
three-level hierarchical bootstrap does resample the three training seeds ---
these bootstraps are over test-sample variation within fixed seeds and do not
model between-seed variance. Claims phrased at the seed level
(``on 3/3 seeds'') are empirical observations about three independent
training runs and are separate from the bootstrap evidence.
Both views are reported to support complete uncertainty disclosure.

\end{document}